%% file: neurips_2026.tex
\documentclass{article}

\PassOptionsToPackage{numbers,sort&compress}{natbib}
\PassOptionsToPackage{table}{xcolor}

\usepackage[main,final]{neurips_2026}
\usepackage[utf8]{inputenc}
\usepackage[T1]{fontenc}
\usepackage{microtype}

\usepackage{titletoc}
\usepackage{url}
\usepackage{xcolor}

\usepackage{amsmath}
\usepackage{amssymb}
\usepackage{amsthm}
\usepackage{bbm}
\usepackage{nicefrac}

\usepackage{graphicx}
\usepackage{booktabs}
\usepackage{multirow}
\usepackage{array}
\usepackage{adjustbox}
\usepackage{wrapfig}
\usepackage{placeins}

\usepackage{algorithm}
\usepackage{algpseudocode}

\usepackage{soul}
\usepackage[most]{tcolorbox}

\newtcolorbox{promptbox}{
    colback=gray!5,
    colframe=gray!50,
    arc=2pt,
    boxrule=0.5pt,
    fontupper=\ttfamily,
    left=4pt, right=4pt, top=4pt, bottom=4pt
}

\usepackage{fontawesome5}

\usepackage{hyperref}
\hypersetup{
  colorlinks=true,
  citecolor=blue!70,
  linkcolor=red!70,
  urlcolor=blue
}

\makeatletter
\renewcommand{\@noticestring}{%
  Accepted as a \emph{Spotlight} paper at the \@neuripsordinal\ Conference on Neural Information
  Processing Systems (NeurIPS~\@neuripsyear).\par
  \copyright~\@neuripsyear\ The Authors.%
}
\makeatother

\theoremstyle{plain}

\theoremstyle{definition}
\newtheorem{definition}{Problem}

\theoremstyle{remark}

\newcommand{\method}{\textsc{Chord}}

\title{\method{}: Cross-Model Hallucination Detection via Relational Graph Discrimination}

\author{%
  Yongxin Deng \\
  University of Technology Sydney \\
  Sydney, Australia \\
  \texttt{yongxin.deng@student.uts.edu.au}
  \And
  Zhen Fang\thanks{Corresponding authors: Zhen Fang and Ling Chen.} \\
  University of Technology Sydney \\
  Sydney, Australia \\
  \texttt{zhen.fang@uts.edu.au}
  \AND
  Guansong Pang \\
  Singapore Management \\
  University \\
  Singapore \\
  \texttt{gspang@smu.edu.sg}
  \And
  Sharon Li \\
  University of \\
  Wisconsin--Madison \\
  Madison, WI, USA \\
  \texttt{sharonli@cs.wisc.edu}
  \And
  Ling Chen\footnotemark[1] \\
  University of \\
  Technology Sydney \\
  Sydney, Australia \\
  \texttt{ling.chen@uts.edu.au}
}

\begin{document}

\maketitle

\begin{abstract}

Hallucination detection is critical for deploying large language models (LLMs) in real-world applications. Due to strong empirical performance, internal representation–based methods have emerged as the prevailing direction for detecting hallucinations, yet they remain largely model-specific and often fail to generalize to unseen LLMs. In this paper, we study an important yet underexplored problem, termed \emph{cross-model hallucination detection} (CMHD), which aims to train hallucination detectors on source LLMs while ensuring performance on unseen target LLMs. The core challenge of CMHD lies in cross-model representation heterogeneity: hidden states from different LLMs exhibit severe semantic inconsistency, which limits the transferability of classical representation-based detectors. Through analysis and empirical validation, we show that relational structures over layers, tokens, and features capture transferable detection signals despite semantic inconsistency. Based on this observation, we propose \textit{\underline{c}ross-model \underline{h}allucination detection via heterogeneity-\underline{o}riented \underline{r}elational \underline{d}iscrimination} (\method{}), a relational graph discrimination framework that represents hidden states as joint relational graphs over layers, tokens, and features. It feeds these graphs into relation-aware attention to obtain structure-aware features, and uses meta-learning to shape these features toward transferable detection signals. Extensive experiments show that \method{} outperforms representative detection methods in cross-model generalization. \href{https://github.com/MiracleworkerD/CHORD}{\faGithub}

\end{abstract}

\section{Introduction}
\label{sec:intro}

Large language models (LLMs) may suffer from hallucinations, generating outputs that are fluent and plausible yet factually incorrect or unsupported by verifiable evidence. Such behavior poses a major challenge to LLMs' safe adoption, particularly in safety-critical domains such as education, healthcare, and finance \citep{harvey2025don, roustan2025clinicians, kang2023deficiency}. Hallucination detection has therefore attracted growing attention in recent research. Existing detection methods can be broadly divided into two categories: training-free methods \citep{manakul-etal-2023-selfcheckgpt, farquhar2024detecting} and training-based methods \citep{kossen2024semantic, wei2024toward}. Among them, training-based methods, which mainly leverage internal representations through lightweight classifiers, have emerged as an important and promising research direction, largely driven by their strong in-domain performance~\citep{huang2025survey, sahoo2024comprehensive, zhang2025siren}.

\begin{figure*}[t]
    \centering
        \hspace*{-0.03\textwidth}
    \includegraphics[width=\textwidth]{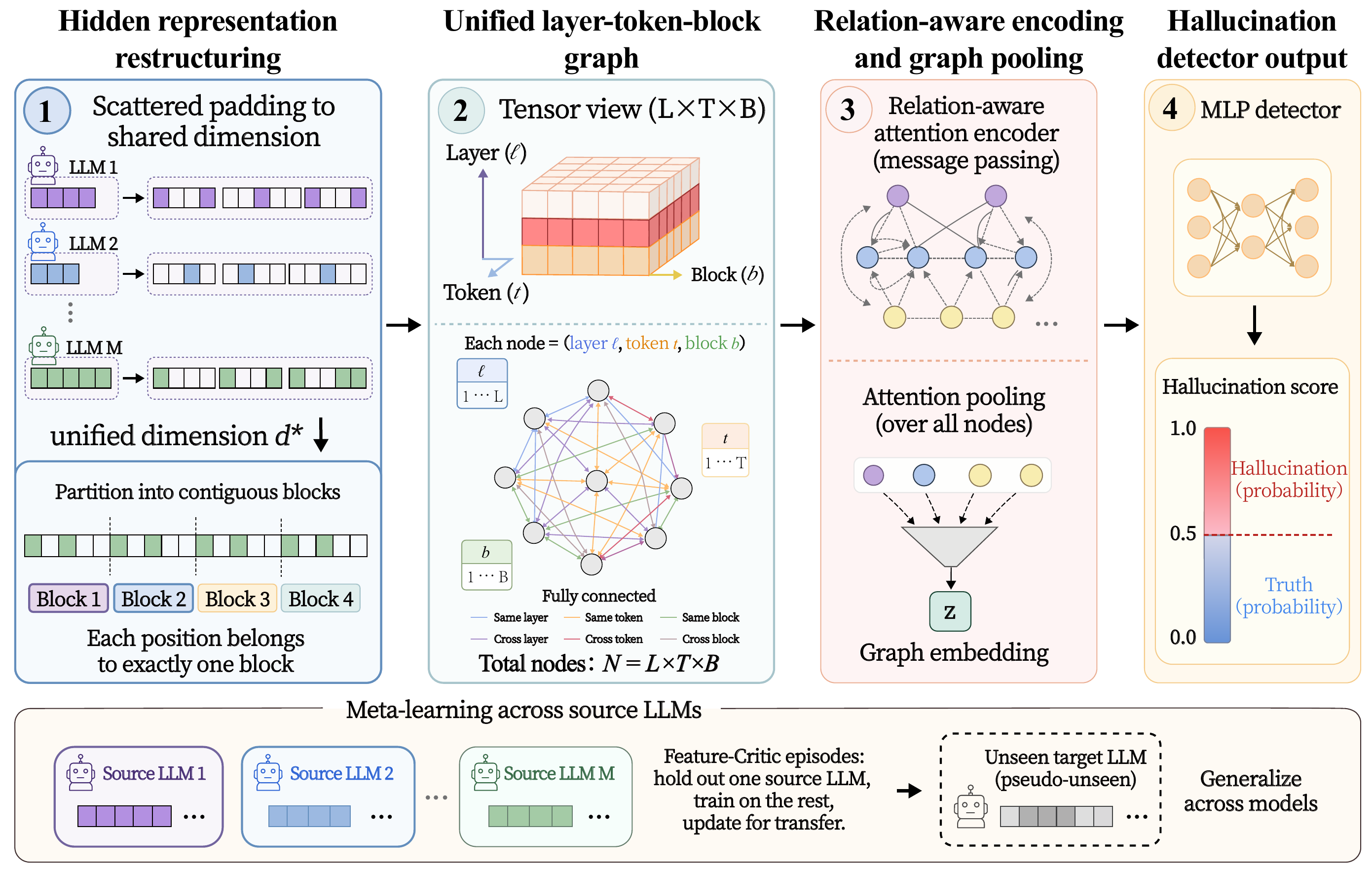}
    \caption{
Overview of \method{}.
\method{} maps hidden states into a shared space, constructs a \textit{layer--token--block graph}, and uses relation-aware attention with graph pooling for hallucination prediction.
Feature-Critic meta-learning encourages the detector to learn transferable signals.
    }
    \vspace{-15pt}
    \label{fig:framework}
\end{figure*}

Despite their empirical success, training-based methods still suffer from a critical limitation: the learned detectors are highly \textit{model-specific}. In particular, detectors trained on one LLM typically fail to generalize to an unseen model, even when the model performs similar tasks (see Appendix~\ref{app:pilot}). Such limited cross-model generalization substantially reduces the practical utility of training-based hallucination detection, as detectors learned on one model often fail on another, requiring retraining or redesign for each new model. Beyond hampering scalability, this also incurs substantial additional costs in supervision, computation, and deployment. These limitations naturally lead to a central question: \textit{how can we build a detector that generalizes well across models, even to unseen ones?}

To tackle the cross-model generalization issue in hallucination detection, we study a more realistic yet challenging setting, termed \emph{cross-model hallucination detection} (CMHD), which seeks to learn a hallucination detector from multiple source models that generalizes well to unseen target models. The core difficulty of CMHD lies in the pronounced \textit{heterogeneity across models}. Different LLMs may vary in several respects, including architecture family, parameter scale, and training procedure, and these differences can further induce different generation behaviors. At the representation level, such heterogeneity typically manifests as severe \textit{cross-model semantic inconsistency}, which fundamentally limits the transferability and generalization of detectors learned from source models to target models.

\begin{figure}[t]
    \vspace{-4mm}
    \centering
    
    \includegraphics[width=\linewidth]{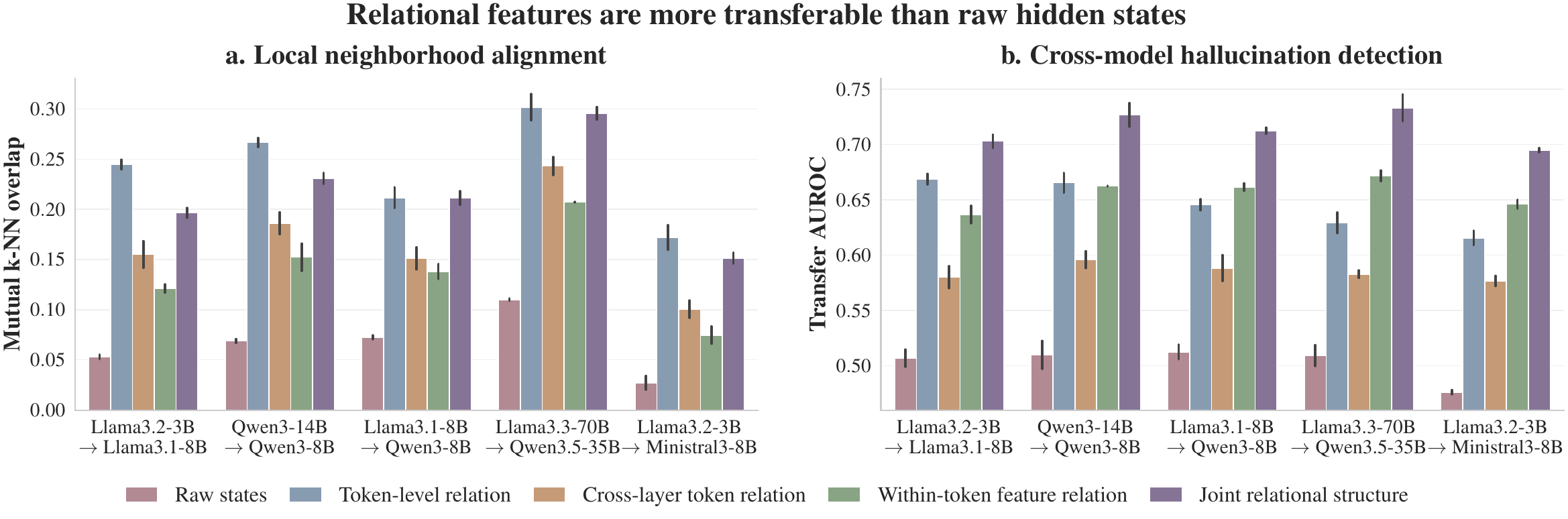}
        \vspace{-18pt}
    \caption{
    Cross-model alignment and hallucination detection transfer.
    Raw hidden states show weak local neighborhood agreement and near-chance transfer, while relational constructions improve both mutual $k$NN overlap and source-to-target AUROC. Experimental details are offered in Appendix~\ref{app:obs_alignment_transfer_details}.
    }
        \vspace{-16pt}
    \label{fig:obs_alignment_detection}
\end{figure}

To enable cross-model generalization, we seek stable, transferable detection signals across heterogeneous semantic representations, reducing reliance on model-specific semantics.
Inspired by \citet{binkowski-etal-2025-hallucination} demonstrating that token-level relational structures aid hallucination detection without relying on individual semantic representations, we construct relation-weighted token graphs from hidden representations, evaluating them through cross-model structural alignment and predictive transfer.
In Figures~\ref{fig:obs_alignment_detection} and~\ref{fig:obs_kde_scores}, token-level relational features exhibit non-trivial cross-model structural alignment, and their source-to-target AUROC is comparable to representative training-free methods across models.
These results suggest relational structures provide a promising transferable signal for CMHD.
Building on this observation and evidence that LLMs encode hierarchical semantics across layers and rich interactions within feature dimensions \citep{tenney2019bert, elhage2021mathematical}, we broaden the relational view to \textit{cross-layer token relations}, \textit{within-token feature relations}, and \textit{joint relations} over layers, tokens, and features.
We summarize each graph with permutation-invariant pooling to reduce sensitivity to node ordering.
Analysis in Section~\ref{sec:observations} shows that the resulting joint relational structures achieve favorable cross-model stability and the highest cross-model AUROC among relational features.

Empirical observations suggest that joint relations over layers, tokens, and feature blocks carry information that is transferable across LLMs and discriminative for hallucination detection. 
Notably, even a construction with fixed handcrafted edges yields useful cross-model signals, suggesting the promise of the relational perspective. 
This motivates us to move beyond fixed relations and learn more effective relational structures for cross-model hallucination detection.
Based on this motivation, we propose 
\textit{\underline{c}ross-model \underline{h}allucination detection via heterogeneity-\underline{o}riented \underline{r}elational \underline{d}iscrimination} (\method{}), 
a relational graph discrimination framework for cross-model hallucination detection.  
\method{} represents hidden states as fully connected joint relational graphs over layers, tokens, and feature blocks, where nodes encode layer-token-feature information and learnable edges capture their relational dependencies. 
It feeds these graphs into relation-aware attention to obtain structure-aware features for distinguishing truthful and hallucinated responses. 
To improve transfer to unseen LLMs, \method{} is trained with a meta-learning strategy \citep{li2019feature} that holds out one source model as a pseudo target model, encouraging the detector to learn transferable detection signals across models.

We conduct extensive experiments in a broad, challenging cross-family transfer setting on eight LLMs from three model families and multiple scales, i.e., Llama-3.2-3B/3.1-8B/3.3-70B \citep{grattafiori2024llama}, Ministral-3-8B/14B \citep{liu2026ministral}, Qwen3-8B/14B \citep{yang2025qwen3} and Qwen3.5-35B-A3B \citep{qwen35blog}, and seven benchmarks, i.e., TruthfulQA \citep{lin2022truthfulqa}, HaluEval \citep{li-etal-2023-halueval}, TriviaQA~\citep{joshi-etal-2017-triviaqa}, CoQA~\citep{reddy-etal-2019-coqa}, GSM8K \citep{cobbe2021training}, MBPP \citep{austin2021program}, and xLAM \citep{liu2024apigen}, spanning factual and conversational question answering, mathematical reasoning, coding, and agent tasks. Section~\ref{sec:exp} shows \method{} consistently improves cross-model generalization and outperforms representative baselines, verifying the effectiveness of relational graph discrimination for hallucination detection across unseen LLMs. \emph{Our main contributions are summarized below:}

$\bullet$ We formulate and study \emph{cross-model hallucination detection} (CMHD), a practical setting where hallucination detectors are trained on source LLMs and evaluated on unseen target LLMs.

$\bullet$ Analysis suggests relational structures over layers, tokens, and feature blocks serve as promising hallucination signals for CMHD, exhibiting favorable cross-model alignment and detection transfer.

$\bullet$ We propose \method{}, a relational graph discrimination framework with relation-aware attention and meta-learning, and demonstrate its effectiveness across model families, scales, and task types.

 \section{Learning Setups}
\label{sec:setups}

\textbf{LLMs and Token Sequences.}
Following \citep{NEURIPS2024_ba927059, oh2025understanding,deng2026beyond}, we model an LLM as a conditional distribution $\mathbb{P}_{\boldsymbol{\theta}}(\cdot)$ over token sequences, where $\boldsymbol{\theta}$ denotes the model parameters. Given an input token sequence $\mathbf{Q} = [x_1, \dots, x_k]$, representing a question optionally concatenated with context, the LLM generates an answer token sequence $\mathbf{A} = [x_{k+1}, \dots, x_{k+l}]$ in an autoregressive manner by predicting each token conditioned on the preceding tokens:
$
    \mathbb{P}_{\boldsymbol{\theta}}(x_j \mid x_1, \dots, x_{j-1}), \text{for } j = k{+}1, \dots, k{+}l.
$

This work studies multiple models. Let $\mathbb{P}_{\boldsymbol{\theta}_1}, \ldots, \mathbb{P}_{\boldsymbol{\theta}_M}$ denote the $M$ LLMs available during training. These training models may vary along several axes, including architecture family, parameter scale, and training procedure, and such variation may induce systematically different generation behaviors.

\textbf{Internal Representations.}
For the $i$-th LLM $\mathbb{P}_{\boldsymbol{\theta}_i}$, let $L_i$ denote its number of layers and $d_i$ its hidden dimension. Given an input question-answer pair $(\mathbf{Q},\mathbf{A})$ with $k+l$ tokens, the hidden representation at layer $\ell$ is defined as
$
\mathbf{H}_{\boldsymbol{\theta}_i}^{\ell}(\mathbf{Q},\mathbf{A})
\in \mathbb{R}^{(k+l)\times d_i}$. 
For each token position $t$, we denote by
$\mathbf{h}_{\boldsymbol{\theta}_i,t}^{\ell}(\mathbf{Q},\mathbf{A})\in\mathbb{R}^{d_i}$
the hidden representation of the $t$-th token at layer $\ell$.

\textbf{Domain and Datasets.} We define the \emph{truthful-answer domain} as a joint distribution $\mathbb{P}_{Q,T}$, where $Q$ is a random variable representing questions and $T$ is a random variable representing factual answers.
Given a question $\mathbf{Q}\sim \mathbb{P}_Q$, the $i$-th model $\mathbb{P}_{\boldsymbol{\theta}_i}(\cdot)$ generates an answer
$\mathbf{A}_i \sim \mathbb{P}_{\boldsymbol{\theta}_i}(\cdot \mid \mathbf{Q})$.
$\mathbf{A}_i$ is then assigned a binary label $y_i$ based on its semantic alignment with a reference answer
$\mathbf{A}_{\mathrm{ref}} \sim \mathbb{P}_{T\mid Q}(\cdot \mid \mathbf{Q})$.
Specifically, if $\mathbf{A}_i$ is semantically aligned with $\mathbf{A}_{\mathrm{ref}}$ or explicitly abstains (e.g., states that it does not know), it is labeled as truthful ($y_i = 1$); if it commits to an answer that is not aligned with $\mathbf{A}_{\mathrm{ref}}$, it is labeled as hallucinated ($y_i = 0$).
Combining $\mathbb{P}_{Q,T}$ with $\mathbb{P}_{\boldsymbol{\theta}_i}$ yields a dataset
\begin{equation}
\label{dataset_format-i}
\mathcal{D}_{i}
=
\big\{
(\mathbf{Q}^1, \mathbf{A}^{1}_i, y_{i}^1), \ldots, (\mathbf{Q}^{n}, \mathbf{A}^{n}_{i}, y^{n}_{i})
\big\},~~\forall i\in\{1,...,M\}.
\end{equation}

\textbf{Cross-Model Hallucination Detection.}
The task considered in this work is defined as follows.

\begin{definition}[Cross-Model Hallucination Detection (CMHD)]\label{def:cmhd}
  Given $M$ training datasets $\mathcal{D}_1,...,\mathcal{D}_M$ constructed from the truthful-answer domain $\mathbb{P}_{Q,T}$ together with $M$ \textit{source models} $\mathbb{P}_{\boldsymbol{\theta}_1}, \ldots, \mathbb{P}_{\boldsymbol{\theta}_M}$
as introduced in Eq.~(\ref{dataset_format-i}), the goal of cross-model hallucination detection (CMHD) is to learn a detector $D$ from $\mathbb{P}_{\boldsymbol{\theta}_1}, \ldots, \mathbb{P}_{\boldsymbol{\theta}_M}$ and $\mathcal{D}_1, \ldots, \mathcal{D}_M$. During testing, given an unseen \textit{target model} $\mathbb{P}_{\boldsymbol{\theta}}$, the detector $D$ is expected to generalize to $\mathbb{P}_{\boldsymbol{\theta}}$. Concretely, for any question $\mathbf{Q} \sim \mathbb{P}_{Q}$ and answer $\mathbf{A} \sim \mathbb{P}_{\boldsymbol{\theta}}(\cdot \mid \mathbf{Q})$, 
\begin{equation}
D(\mathbf{Q},\mathbf{A}, \mathbb{P}_{\boldsymbol{\theta}}) = 1,~\text{if } \mathbf{A} \text{ does not contradict } \mathbf{T} \sim \mathbb{P}_{T\mid Q}(\cdot \mid \mathbf{Q});~\text{otherwise}, ~0.
\end{equation}
\end{definition}

\section{Analysis of Hallucination Signals: What Transfers Across Models?}
\label{sec:observations}

In Figure~\ref{fig_intro_observation}, the representation-based detectors often generalize poorly across models. This failure stems from {cross-model semantic inconsistency}, limiting transfer from source models to target models \citep{kornblith2019similarity, klabunde2025similarity, huang2025cross} (see Appendix~\ref{app:pilot}). We therefore \textit{seek detection signals that remain stable across heterogeneous representations, thereby reducing the detector's reliance on model-specific  representations}.

\subsection{Stable Cross-Model Signals from Token-Level Relations}
\label{sec:token_observations}

Motivated by the recent work of \citet{binkowski-etal-2025-hallucination}, which shows that \textit{token-level relational structures} provide highly reliable signals for downstream hallucination detection, we carefully examine whether these token-level relations can effectively serve as universal cross-model signals.

\textbf{Token-Level Relation.} Given a question-answer pair $(\mathbf{Q}, \mathbf{A})$ and the $i$-th LLM 
$\mathbb{P}_{\boldsymbol{\theta}_i}$, we obtain the hidden representation 
$\mathbf{h}_{\boldsymbol{\theta}_i,t}^{\ell}(\mathbf{Q},\mathbf{A})$ 
of the $t$-th token at layer $\ell$. 
Motivated by \citet{wang2019dynamic} and the state-delta analysis \citep{tang2025augmenting,zhu2025llm,xu2025amplify}, for two token positions $t$ and $\tilde{t}$, we design their relation as follows:
\begin{equation}\label{Token-Relation-1}
     R_{i}^{\mathrm{tok},\ell}(t,\tilde{t};\mathbf{Q},\mathbf{A})
    =
    s^{\mathrm{tok}}(t,\tilde{t})
 \big \|
        \mathbf{h}_{\boldsymbol{\theta}_i,t}^{\ell}(\mathbf{Q},\mathbf{A})
        -
        \mathbf{h}_{\boldsymbol{\theta}_i,\tilde{t}}^{\ell}(\mathbf{Q},\mathbf{A})
\big \|_{2},
\end{equation}
where $\|\cdot\|_{2}$ denotes the $\ell_2$ norm, and $s^{\mathrm{tok}}(t,\tilde{t})= \exp\left(
        -{|t-\tilde{t}|}/{\tau_{\mathrm{tok}}}
    \right)$ is the position-based relation weight between the $t$-th and $\tilde{t}$-th tokens. Here $\tau_{\mathrm{tok}}>0$ controls decay with token-position distance.

\textbf{Remark.} \textit{We use this simple, non-learnable relation as a diagnostic test of whether basic relational information supports cross-model detection. 
Improved transfer under this lightweight construction would imply that relational structure provides meaningful transferable signals for detection.}

\begin{figure}[t]
    \centering
    \vspace{-4mm}
    \includegraphics[width=\linewidth]{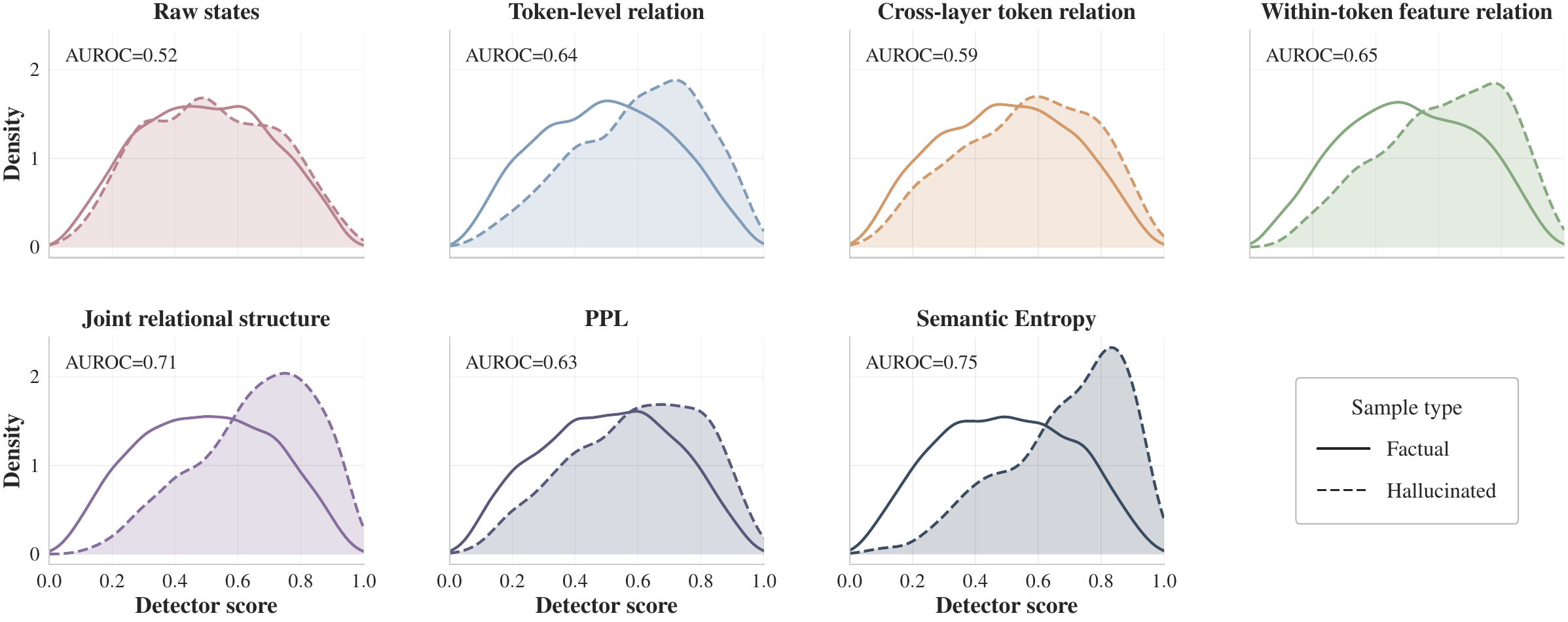}
            \vspace{-10pt}
    \caption{
    Detector score distributions for Llama 8B to Qwen 8B.
    Token-level relations already separate factual and hallucinated examples better than PPL, while stronger relational constructions approach the Semantic Entropy baseline. Experimental details are provided in Appendix~\ref{app:obs_alignment_transfer_details}.
    }
            \vspace{-14pt}
    \label{fig:obs_kde_scores}
\end{figure}

\textbf{Token-Level Relation Graph and Pooling.} While Eq.~(\ref{Token-Relation-1}) captures relations between token pairs, hallucination signals often emerge from the collective structure among multiple tokens. We therefore aggregate pairwise relations over selected tokens into a compact vector representation that summarizes the relational structure of the question-answer pair $(\mathbf{Q},\mathbf{A})$. Let $\mathbf{T}^{\mathrm{sel}}=\{t_1,\ldots,t_r\}$ denote the selected token positions. 
We construct a fully connected token-level graph 
$\mathcal{G}_{i}^{\mathrm{tok},\ell}(\mathbf{Q},\mathbf{A})$ without self-loops, where each node corresponds to a selected token and each edge $(t,\tilde{t})$ is weighted by the token-level relation 
$R_{i}^{\mathrm{tok},\ell}(t,\tilde{t};\mathbf{Q},\mathbf{A})$ defined in Eq.~(\ref{Token-Relation-1}). 
The node feature at position $t$ is the hidden representation 
$\mathbf{h}_{\boldsymbol{\theta}_i,t}^{\ell}(\mathbf{Q},\mathbf{A})$ extracted from $\mathbb{P}_{\boldsymbol{\theta}_i}$. Following the standard graph message-passing and readout paradigm \citep{gilmer2017neural,kipf2016semi}, we perform one relation-weighted aggregation step and mean readout:
\begin{equation}\label{Token-Aggregation}
    \mathbf{z}_{i}^{\mathrm{tok},\ell}(\mathbf{Q},\mathbf{A})
    =
    \frac{1}{r}
    \sum_{t\in\mathbf{T}^{\mathrm{sel}}}
    \sum_{\tilde{t}\in\mathbf{T}^{\mathrm{sel}},\,\tilde{t}\neq t}
   \frac{
R_{i}^{\mathrm{tok},\ell}(t,\tilde{t};\mathbf{Q},\mathbf{A})
}{
\sum_{t'\in\mathbf{T}^{\mathrm{sel}},\,t'\neq t}
R_{i}^{\mathrm{tok},\ell}(t,t';\mathbf{Q},\mathbf{A})
}\mathbf{h}_{\boldsymbol{\theta}_i,\tilde{t}}^{\ell}(\mathbf{Q},\mathbf{A}).
\end{equation} 

The resulting vector $\mathbf{z}_{i}^{\mathrm{tok},\ell}(\mathbf{Q},\mathbf{A})$ provides a compact relation-weighted summary of the selected token representations. 
Because the final mean readout is implemented as graph-level average pooling over all selected token nodes, the summary is permutation-invariant, providing theoretical support for reducing sensitivity to different node orderings in graphs derived from different LLMs.

\textbf{Remark.} \textit{To ensure dimension consistency across LLMs, we map hidden representations into a common space before the aggregation in Eq.~(\ref{Token-Aggregation}). Details are provided in Eq. (\ref{eq:index_map}) and Section~\ref{sec:method}.}

\textbf{Evaluation for Cross-Model Stability and Transferability.}
We assess the token-level features 
$\mathbf{z}_{i}^{\mathrm{tok},\ell}(\mathbf{Q},\mathbf{A})$ 
using two evaluation criteria: \textit{cross-model structural stability} and \textit{predictive transfer}. 
1) Structural stability is measured by mutual $k$-nearest-neighbor alignment 
\citep{huh2024position, oron2017best, klabunde2025similarity, park2023quantifying}, 
which assesses whether the induced relational structure is consistent across LLMs. 
2) Predictive transfer is measured by cross-model AUROC, which evaluates whether features from one LLM can predict hallucinations on another. 
Both evaluation criteria involve two LLMs
$\mathbb{P}_{\boldsymbol{\theta}_i}$ and 
$\mathbb{P}_{\boldsymbol{\theta}_j}$.

Following \citet{huh2024position}, we use mutual $k$-nearest-neighbor ($k$NN) alignment to assess cross-model structural stability. 
For each pair $(\mathbf{Q}^s, \mathbf{A}^{s}_i)$ from dataset $\mathcal{D}_i$ (see Eq. (\ref{dataset_format-i})), mutual $k$NN alignment is
\begin{equation}\label{Token-mNN}
m_{\mathrm{NN}}^{\mathrm{tok},\ell}(i,j)
    =
    \frac{1}{nk}\sum_{s=1}^{n}
    \big|\mathcal{N}_{k}^{i}(s)\cap\mathcal{N}_{k}^{j}(s)\big|,~\text{where} ~\mathcal{N}_{k}^{i}(s)
=
\operatorname{Top}k_{s' \in \{1,\ldots,n\}\setminus\{s\}}
\,
\mathrm{sim}_{i}^{\mathrm{tok},\ell}(s,s'),
\end{equation}
here $|\cdot|$ counts the number of elements in the intersection $\mathcal{N}_{k}^{i}(s)\cap\mathcal{N}_{k}^{j}(s)$, and $\operatorname{Top}k$ returns the indices of the $k$ largest similarity scores $\mathrm{sim}_{i}^{\mathrm{tok},\ell}(s,s')$. Following \citet{huh2024position}, we use the inner product to compute the  similarity score, i.e., $\mathrm{sim}_{i}^{\mathrm{tok},\ell}(s,s')=
\langle
\mathbf{z}_{i}^{\mathrm{tok},\ell}(\mathbf{Q}^s,\mathbf{A}_{i}^s),
\mathbf{z}_{i}^{\mathrm{tok},\ell}(\mathbf{Q}^{s'},\mathbf{A}_{i}^{s'})
\rangle$. 

Cross-model AUROC is used to assess predictive transfer of hallucination detection. 
Specifically, an MLP detector is trained on the source features 
$\mathbf{z}_{i}^{\mathrm{tok},\ell}(\mathbf{Q}^s,\mathbf{A}_{i}^s)$ from $\mathcal{D}_i$ with labels $y_i^s$. 
The trained detector is then evaluated directly on the target features 
$\mathbf{z}_{j}^{\mathrm{tok},\ell}(\mathbf{Q}^s,\mathbf{A}_{j}^s)$ from $\mathcal{D}_j$ using labels $y_j^s$. 
A higher AUROC indicates stronger cross-model transfer of hallucination-discriminative signals.

\textbf{Empirical Observations and Implications.}
We evaluate 
$\mathbf{z}_{i}^{\mathrm{tok},\ell}(\mathbf{Q},\mathbf{A})$ 
on five source-target LLM pairs and highlight two key observations with their implications. The observations are shown below.

\noindent (1) \emph{Cross-model stability.} In Figure~\ref{fig:obs_alignment_detection}\textcolor{red}{a}, raw hidden representations yield mutual $k$NN alignments of $0.03$--$0.11$, whereas the token-level relational features increase the alignment to $0.17$--$0.30$ across the five transfer pairs. 
This indicates that the relational features preserve a more consistent neighborhood structure across LLMs than raw representations. 
To contextualize these values, if the two $k$NN sets were randomly and independently selected from the $n-1$ remaining samples, their expected normalized overlap would be $k/(n-1)$. 
In our setup, where $k=10$ and $n=5000$, this random-overlap baseline is near $0.002$. 
The observed token-level alignments are therefore roughly $85$ to $150$ times random neighborhood agreement, indicating non-trivial cross-model relational stability.

\noindent (2) \emph{Cross-model transferability.} In Figure~\ref{fig:obs_alignment_detection}\textcolor{red}{b}, an MLP detector trained on raw hidden representations from a source LLM and evaluated on a target LLM stays close to chance ($\mathrm{AUROC}\approx 0.47$–$0.52$), while the same detector trained on token-level relational features reaches $0.62$–$0.67$ across all transfer pairs. We further compare with two training-free baselines (see Figure~\ref{fig:obs_kde_scores}) that are inherently cross-model: Perplexity (PPL)~\citep{ren2023outofdistribution} as a general-purpose baseline at about $0.63$, and Semantic Entropy~\citep{farquhar2024detecting} as a strong specialized baseline at about $0.75$. The token-level cross-model AUROC thus falls short of Semantic Entropy, yet matches PPL despite the simplicity of its construction.

\textit{Implication.} Taken together, these observations suggest that token-level relational features provide a stable and transferable structural signal for cross-model hallucination detection, beyond what is captured by raw hidden representations. 
However, the remaining gap to Semantic Entropy indicates that token-level relations capture only part of the transferable structure needed for robust cross-model generalization. 
This motivates the broader relational analysis in Section~\ref{sec:joint_relations}, where we further study cross-layer relations, within-token feature relations, and their joint relational structures.

\definecolor{SkyBlueRow}{RGB}{224,244,250}
\begin{table*}[thbp]
\caption{
Cross-model performance (AUROC). 
To ensure fair comparison, we use a leave-one-family-out protocol: training-based methods are trained on models from two source families and evaluated on models from the held-out target family, while training-free methods are directly applied to the same targets (\emph{Appendix~\ref{app:additional_exp_analysis}}). 
Best results are \textbf{bold}, and second-best results are \underline{underlined}.
}
\vspace{1mm}
\centering
\label{tab:main_cross_family_results}
\resizebox{\linewidth}{!}{%
\begin{tabular}{lcccccccccccccccc}
\toprule
\multirow{4}{*}{\textbf{Method}} & \multicolumn{16}{c}{\textbf{Target Model}} \\
& \multicolumn{8}{c}{Llama3.2-3B} & \multicolumn{8}{c}{Llama3.1-8B} \\
\cmidrule(lr){2-9}\cmidrule(lr){10-17}
& \multicolumn{8}{c}{Mean AUROC$\uparrow$} & \multicolumn{8}{c}{Mean AUROC$\uparrow$} \\
& TruthfulQA & HaluEval & TriviaQA & CoQA & GSM8K & MBPP & xLAM & \textbf{Average} & TruthfulQA & HaluEval & TriviaQA & CoQA & GSM8K & MBPP & xLAM & \textbf{Average} \\
\midrule
\rowcolor{gray!18} \multicolumn{17}{l}{\textbf{Training-free methods}} \\
Perplexity & 0.6329 & 0.6464 & 0.6448 & 0.6296 & 0.6536 & 0.6188 & 0.6621 & 0.6412 & 0.6546 & 0.6452 & 0.6545 & 0.6395 & 0.6523 & 0.6024 & 0.6857 & 0.6477 \\
Semantic Entropy & \underline{0.6862} & 0.7095 & 0.7006 & \underline{0.6934} & 0.7112 & 0.6571 & 0.7223 & 0.6972 & 0.6991 & 0.7113 & 0.7065 & \underline{0.6919} & 0.7258 & 0.6636 & 0.7438 & 0.7060 \\
EigenScore & 0.6636 & 0.6872 & 0.6767 & 0.6719 & 0.6772 & 0.6508 & 0.7045 & 0.6760 & 0.6985 & 0.6963 & 0.6803 & 0.6747 & 0.7068 & 0.6614 & 0.6957 & 0.6877 \\
Lexical Similarity & 0.6542 & 0.6747 & 0.6878 & 0.6786 & 0.6454 & 0.6225 & 0.6603 & 0.6605 & 0.6760 & 0.6681 & 0.6928 & 0.6628 & 0.6495 & 0.6236 & 0.6871 & 0.6657 \\
Verbalize & 0.5199 & 0.5306 & 0.5220 & 0.5290 & 0.5207 & 0.5069 & 0.5470 & 0.5252 & 0.5238 & 0.5398 & 0.5110 & 0.5050 & 0.5474 & 0.4884 & 0.5595 & 0.5250 \\
InterrogateLLM & 0.6845 & \underline{0.7353} & \underline{0.7012} & 0.6915 & \underline{0.7177} & \underline{0.6915} & \underline{0.7253} & \underline{0.7067} & \underline{0.7185} & \underline{0.7211} & \underline{0.7201} & 0.6881 & \underline{0.7293} & \underline{0.6865} & \underline{0.7469} & \underline{0.7158} \\
\rowcolor{gray!18} \multicolumn{17}{l}{\textbf{Training-based methods}} \\
MM & 0.4857 & 0.5170 & 0.5129 & 0.4874 & 0.5061 & 0.5035 & 0.5200 & 0.5047 & 0.5127 & 0.5029 & 0.5247 & 0.5000 & 0.5141 & 0.4808 & 0.5052 & 0.5058 \\
SEP & 0.5079 & 0.5193 & 0.5373 & 0.5189 & 0.5300 & 0.5032 & 0.5290 & 0.5208 & 0.5191 & 0.5346 & 0.5142 & 0.5089 & 0.5382 & 0.4876 & 0.5395 & 0.5203 \\
SAPLMA & 0.4941 & 0.5191 & 0.5239 & 0.5390 & 0.5158 & 0.4901 & 0.5397 & 0.5174 & 0.5326 & 0.5311 & 0.5250 & 0.5013 & 0.5382 & 0.4818 & 0.5386 & 0.5212 \\
LapEigvals & 0.4817 & 0.5021 & 0.4740 & 0.4965 & 0.5264 & 0.4765 & 0.5295 & 0.4981 & 0.4981 & 0.4916 & 0.4712 & 0.4789 & 0.5083 & 0.4725 & 0.5225 & 0.4919 \\
\rowcolor{gray!18} \multicolumn{17}{l}{\textbf{Proposed method}} \\
\rowcolor{SkyBlueRow} \textbf{CHORD} & \textbf{0.7766} & \textbf{0.7926} & \textbf{0.7991} & \textbf{0.7920} & \textbf{0.8193} & \textbf{0.7351} & \textbf{0.8437} & \textbf{0.7941} & \textbf{0.7989} & \textbf{0.8102} & \textbf{0.8156} & \textbf{0.7823} & \textbf{0.8362} & \textbf{0.7307} & \textbf{0.8650} & \textbf{0.8056} \\
\midrule[1.5pt]
\multicolumn{17}{c}{} \\[-2mm]
& \multicolumn{8}{c}{Llama3.3-70B} & \multicolumn{8}{c}{Qwen3-8B} \\
\cmidrule(lr){2-9}\cmidrule(lr){10-17}
\midrule
\rowcolor{gray!18} \multicolumn{17}{l}{\textbf{Training-free methods}} \\
Perplexity & 0.6575 & 0.6747 & 0.6717 & 0.6748 & 0.6765 & 0.6353 & 0.6931 & 0.6691 & 0.6345 & 0.6398 & 0.6781 & 0.6542 & 0.6653 & 0.6179 & 0.6726 & 0.6518 \\
Semantic Entropy & 0.7096 & 0.7366 & 0.7231 & \underline{0.7272} & 0.7410 & 0.6697 & 0.7406 & 0.7211 & 0.6840 & 0.6800 & \underline{0.7127} & 0.6994 & \underline{0.7143} & 0.6502 & 0.7228 & 0.6948 \\
EigenScore & 0.7076 & 0.7087 & 0.6961 & 0.7095 & 0.7094 & 0.6703 & 0.7093 & 0.7016 & 0.6785 & 0.6817 & 0.6869 & 0.6873 & 0.6813 & 0.6683 & 0.7085 & 0.6846 \\
Lexical Similarity & 0.6844 & 0.6797 & 0.7040 & 0.7088 & 0.6854 & 0.6432 & 0.6858 & 0.6845 & 0.6538 & 0.6538 & 0.7089 & 0.6966 & 0.6441 & 0.6401 & 0.6688 & 0.6666 \\
Verbalize & 0.5472 & 0.5531 & 0.5472 & 0.5385 & 0.5592 & 0.5139 & 0.5584 & 0.5454 & 0.5193 & 0.5090 & 0.5200 & 0.5170 & 0.5339 & 0.4969 & 0.5463 & 0.5203 \\
InterrogateLLM & \underline{0.7343} & \underline{0.7454} & \underline{0.7472} & 0.7219 & \underline{0.7473} & \underline{0.7181} & \underline{0.7477} & \underline{0.7374} & \underline{0.6950} & \underline{0.7077} & 0.7044 & \underline{0.7173} & 0.7082 & \underline{0.6834} & \underline{0.7483} & \underline{0.7092} \\
\rowcolor{gray!18} \multicolumn{17}{l}{\textbf{Training-based methods}} \\
MM & 0.5095 & 0.5244 & 0.5288 & 0.5195 & 0.5259 & 0.4814 & 0.5153 & 0.5150 & 0.4923 & 0.4849 & 0.5044 & 0.5095 & 0.5046 & 0.4837 & 0.5171 & 0.4995 \\
SEP & 0.5117 & 0.5352 & 0.5400 & 0.5243 & 0.5263 & 0.4905 & 0.5297 & 0.5225 & 0.5147 & 0.5030 & 0.4993 & 0.5062 & 0.5159 & 0.4823 & 0.5228 & 0.5063 \\
SAPLMA & 0.5250 & 0.5177 & 0.5231 & 0.5364 & 0.5375 & 0.4840 & 0.5386 & 0.5232 & 0.4822 & 0.5223 & 0.5171 & 0.5093 & 0.5167 & 0.4922 & 0.5384 & 0.5112 \\
LapEigvals & 0.5083 & 0.4992 & 0.5167 & 0.5204 & 0.5132 & 0.4718 & 0.5280 & 0.5082 & 0.4831 & 0.4872 & 0.5141 & 0.4873 & 0.5063 & 0.4716 & 0.5292 & 0.4970 \\
\rowcolor{gray!18} \multicolumn{17}{l}{\textbf{Proposed method}} \\
\rowcolor{SkyBlueRow} \textbf{CHORD} & \textbf{0.8276} & \textbf{0.8374} & \textbf{0.8257} & \textbf{0.8302} & \textbf{0.8568} & \textbf{0.7605} & \textbf{0.8713} & \textbf{0.8299} & \textbf{0.7896} & \textbf{0.7941} & \textbf{0.8192} & \textbf{0.7991} & \textbf{0.8256} & \textbf{0.7481} & \textbf{0.8484} & \textbf{0.8034} \\
\midrule[1.5pt]
\multicolumn{17}{c}{} \\[-2mm]
& \multicolumn{8}{c}{Qwen3-14B} & \multicolumn{8}{c}{Qwen3.5-35B} \\
\cmidrule(lr){2-9}\cmidrule(lr){10-17}
\midrule
\rowcolor{gray!18} \multicolumn{17}{l}{\textbf{Training-free methods}} \\
Perplexity & 0.6408 & 0.6650 & 0.6442 & 0.6519 & 0.6363 & 0.6164 & 0.6851 & 0.6485 & 0.6460 & 0.6488 & 0.6751 & 0.6601 & 0.6614 & 0.6361 & 0.6732 & 0.6572 \\
Semantic Entropy & 0.6810 & 0.6968 & 0.6966 & 0.6958 & 0.7193 & 0.6438 & 0.7195 & 0.6933 & \underline{0.7078} & \underline{0.7110} & 0.7211 & 0.7052 & 0.7392 & 0.6738 & 0.7278 & 0.7123 \\
EigenScore & 0.6931 & 0.6891 & 0.6965 & 0.6748 & 0.7013 & 0.6556 & 0.7090 & 0.6885 & 0.7061 & 0.7018 & 0.7072 & 0.6854 & 0.7093 & 0.6648 & 0.7046 & 0.6970 \\
Lexical Similarity & 0.6498 & 0.6560 & 0.6779 & 0.6795 & 0.6529 & 0.6368 & 0.6834 & 0.6623 & 0.6752 & 0.6756 & 0.7050 & 0.7083 & 0.6583 & 0.6377 & 0.6730 & 0.6762 \\
Verbalize & 0.5349 & 0.5262 & 0.5274 & 0.5261 & 0.5329 & 0.5091 & 0.5578 & 0.5306 & 0.5286 & 0.5352 & 0.5511 & 0.5279 & 0.5460 & 0.5150 & 0.5550 & 0.5370 \\
InterrogateLLM & \underline{0.7009} & \underline{0.7127} & \underline{0.7062} & \underline{0.7014} & \underline{0.7496} & \underline{0.6904} & \underline{0.7435} & \underline{0.7150} & 0.6978 & 0.6997 & \underline{0.7248} & \underline{0.7273} & \underline{0.7438} & \underline{0.6902} & \underline{0.7376} & \underline{0.7173} \\
\rowcolor{gray!18} \multicolumn{17}{l}{\textbf{Training-based methods}} \\
MM & 0.5033 & 0.5228 & 0.5270 & 0.4910 & 0.5120 & 0.4966 & 0.5269 & 0.5114 & 0.4844 & 0.4817 & 0.5243 & 0.5240 & 0.5280 & 0.4848 & 0.5116 & 0.5055 \\
SEP & 0.5277 & 0.5040 & 0.5072 & 0.5170 & 0.5258 & 0.4980 & 0.5317 & 0.5159 & 0.5395 & 0.5139 & 0.5284 & 0.5222 & 0.5301 & 0.4874 & 0.5350 & 0.5224 \\
SAPLMA & 0.5109 & 0.5192 & 0.5226 & 0.5385 & 0.5287 & 0.4916 & 0.5370 & 0.5212 & 0.5198 & 0.5174 & 0.5268 & 0.5303 & 0.5240 & 0.4868 & 0.5392 & 0.5206 \\
LapEigvals & 0.4840 & 0.5049 & 0.5113 & 0.4969 & 0.5062 & 0.4822 & 0.5263 & 0.5017 & 0.4712 & 0.4891 & 0.5108 & 0.4829 & 0.5289 & 0.4727 & 0.5291 & 0.4978 \\
\rowcolor{gray!18} \multicolumn{17}{l}{\textbf{Proposed method}} \\
\rowcolor{SkyBlueRow} \textbf{CHORD} & \textbf{0.7720} & \textbf{0.8237} & \textbf{0.8143} & \textbf{0.8070} & \textbf{0.8387} & \textbf{0.7248} & \textbf{0.8722} & \textbf{0.8075} & \textbf{0.8069} & \textbf{0.8104} & \textbf{0.8333} & \textbf{0.8027} & \textbf{0.8406} & \textbf{0.7207} & \textbf{0.8472} & \textbf{0.8088} \\
\midrule[1.5pt]
\multicolumn{17}{c}{} \\[-2mm]
& \multicolumn{8}{c}{Ministral3-8B} & \multicolumn{8}{c}{Ministral3-14B} \\
\cmidrule(lr){2-9}\cmidrule(lr){10-17}
\midrule
\rowcolor{gray!18} \multicolumn{17}{l}{\textbf{Training-free methods}} \\
Perplexity & 0.6443 & 0.6328 & 0.6576 & 0.6397 & 0.6528 & 0.6308 & 0.6785 & 0.6481 & 0.6335 & 0.6510 & 0.6471 & 0.6357 & 0.6526 & 0.6029 & 0.6635 & 0.6409 \\
Semantic Entropy & 0.6828 & 0.6800 & \underline{0.7148} & 0.6867 & 0.7045 & 0.6424 & 0.7338 & 0.6921 & 0.6882 & 0.6925 & 0.7006 & \underline{0.6930} & 0.7104 & 0.6534 & 0.7283 & 0.6952 \\
EigenScore & 0.6856 & 0.6729 & 0.6918 & 0.6788 & 0.6836 & 0.6519 & 0.7076 & 0.6817 & 0.6668 & 0.7015 & 0.6846 & 0.6746 & 0.6926 & 0.6509 & 0.7090 & 0.6829 \\
Lexical Similarity & 0.6530 & 0.6474 & 0.7075 & 0.6906 & 0.6479 & 0.6265 & 0.6698 & 0.6632 & 0.6632 & 0.6595 & 0.6945 & 0.6788 & 0.6610 & 0.6255 & 0.6682 & 0.6644 \\
Verbalize & 0.5019 & 0.5116 & 0.5440 & 0.5172 & 0.5230 & 0.5112 & 0.5528 & 0.5231 & 0.5143 & 0.5340 & 0.5158 & 0.5075 & 0.5231 & 0.5048 & 0.5506 & 0.5214 \\
InterrogateLLM & \underline{0.6963} & \underline{0.7002} & 0.7030 & \underline{0.6969} & \underline{0.7254} & \underline{0.6961} & \underline{0.7446} & \underline{0.7089} & \underline{0.6989} & \underline{0.7295} & \underline{0.7097} & 0.6862 & \underline{0.7163} & \underline{0.6957} & \underline{0.7432} & \underline{0.7114} \\
\rowcolor{gray!18} \multicolumn{17}{l}{\textbf{Training-based methods}} \\
MM & 0.4807 & 0.4840 & 0.5294 & 0.5123 & 0.5298 & 0.4853 & 0.5285 & 0.5071 & 0.4866 & 0.5095 & 0.5123 & 0.4854 & 0.5284 & 0.4924 & 0.5232 & 0.5054 \\
SEP & 0.5232 & 0.5188 & 0.5162 & 0.4990 & 0.5033 & 0.5030 & 0.5288 & 0.5132 & 0.4860 & 0.5076 & 0.5123 & 0.4999 & 0.5186 & 0.4895 & 0.5391 & 0.5076 \\
SAPLMA & 0.4816 & 0.5275 & 0.5396 & 0.5073 & 0.5335 & 0.4944 & 0.5392 & 0.5176 & 0.4865 & 0.5232 & 0.4952 & 0.5091 & 0.5266 & 0.4839 & 0.5386 & 0.5090 \\
LapEigvals & 0.4708 & 0.4737 & 0.5019 & 0.5031 & 0.4983 & 0.4733 & 0.5295 & 0.4929 & 0.4847 & 0.4766 & 0.5179 & 0.4800 & 0.5058 & 0.4702 & 0.5161 & 0.4930 \\
\rowcolor{gray!18} \multicolumn{17}{l}{\textbf{Proposed method}} \\
\rowcolor{SkyBlueRow} \textbf{CHORD} & \textbf{0.7663} & \textbf{0.8005} & \textbf{0.7914} & \textbf{0.7836} & \textbf{0.8021} & \textbf{0.7263} & \textbf{0.8474} & \textbf{0.7882} & \textbf{0.7760} & \textbf{0.7988} & \textbf{0.8046} & \textbf{0.7933} & \textbf{0.8193} & \textbf{0.7237} & \textbf{0.8557} & \textbf{0.7959} \\
\bottomrule
\end{tabular}%
}
\end{table*}

\subsection{From Token-Level to Joint Relational Structures}
\label{sec:joint_relations}

Building on the token-level relations in Section~\ref{sec:token_observations}, and motivated by findings that LLMs process hierarchical semantics across layers and exhibit rich feature-level interactions \citep{tenney2019bert, elhage2021mathematical}, we study three broader forms of relational structure: \textit{cross-layer token relations}, \textit{within-token feature relations}, and \textit{joint relations} over layers, tokens, and features. All three follow Section~\ref{sec:token_observations}'s pipeline, differing only in graph nodes and edge weights. We reuse $\mathbf{T}^{\mathrm{sel}}$ and select a layer subset $\mathbf{L}^{\mathrm{sel}}$ from the $L_i$ layers of $\mathbb{P}_{\boldsymbol{\theta}_i}$, with $|\mathbf{L}^{\mathrm{sel}}|$ shared across LLMs to keep graph size constant. For \textit{within-token feature} and \textit{joint relations}, the representation dimension is partitioned into $B$ equal-size contiguous blocks, with $\mathcal{I}_b$ indexing the $b$-th block (Eq.~(\ref{eq:block_partition})); each block serves as a node~\citep{xu2018powerful,dosovitskiy2021an}. Details in Appendix~\ref{app:obs_pipeline}.

\textbf{Cross-Layer Token Relations.}
For each layer $\ell\in\mathbf{L}^{\mathrm{sel}}$, we form a layer-level summary by stacking the hidden representations of all selected tokens at this layer into a matrix:
\begin{equation}\label{Layer-Relation-0}
    \mathbf{H}_{\boldsymbol{\theta}_i}^{\ell,\mathrm{sel}}(\mathbf{Q},\mathbf{A})
    =
    \big[
        \mathbf{h}_{\boldsymbol{\theta}_i,t_1}^{\ell}(\mathbf{Q},\mathbf{A}),
        \ldots,
        \mathbf{h}_{\boldsymbol{\theta}_i,t_r}^{\ell}(\mathbf{Q},\mathbf{A})
    \big]^{\top}
    \in\mathbb{R}^{r\times d_i}.
\end{equation}
Then, we define the node for layer $\ell$ as $\mathrm{vec}(\mathbf{H}_{\boldsymbol{\theta}_i}^{\ell,\mathrm{sel}})\in\mathbb{R}^{rd_i}$, with $\mathrm{vec}(\cdot)$ flattening a matrix into a vector by stacking its rows. To define edges between layer nodes, we use the standard cosine distance for comparing Transformer-layer representations~\citep{phang2021fine,voita2019bottom,sajjad2023effect}, modulated by a depth-decay factor:
\begin{equation}\label{Layer-Relation}
    R_{i}^{\mathrm{lyr}}(\ell,\tilde{\ell};\mathbf{Q},\mathbf{A})
    =
    s^{\mathrm{lyr}}(\ell,\tilde{\ell})\,
    \big(
        1-
        \cos
        \big(
            \mathrm{vec}(\mathbf{H}_{\boldsymbol{\theta}_i}^{\ell,\mathrm{sel}}),
            \mathrm{vec}(\mathbf{H}_{\boldsymbol{\theta}_i}^{\tilde{\ell},\mathrm{sel}})
        \big)
    \big),
\end{equation}
where $\cos(\cdot,\cdot)$ denotes cosine similarity, and 
$s^{\mathrm{lyr}}(\ell,\tilde{\ell})=\exp(-|\ell-\tilde{\ell}|/\tau_{\mathrm{lyr}})$ is the depth-decay factor between layers $\ell$ and $\tilde{\ell}$. The bandwidth $\tau_{\mathrm{lyr}}>0$ controls relation decay with layer distance.

\textbf{Within-Token Feature Relations.}
For each feature block $b\in\{1,\ldots,B\}$ and each selected layer $\ell\in\mathbf{L}^{\mathrm{sel}}$, we first collect the $b$-th hidden-state block of all selected tokens at layer $\ell$:
\begin{equation}\label{Feat-Relation-0}
    \mathbf{V}_{\boldsymbol{\theta}_i,b}^{\ell}(\mathbf{Q},\mathbf{A})
    =
    \big[
        \mathbf{h}_{\boldsymbol{\theta}_i,t_1}^{\ell}(\mathbf{Q},\mathbf{A})[\mathcal{I}_b],
        \ldots,
        \mathbf{h}_{\boldsymbol{\theta}_i,t_r}^{\ell}(\mathbf{Q},\mathbf{A})[\mathcal{I}_b]
    \big]^{\top}
    \in \mathbb{R}^{r\times |\mathcal{I}_b|}.
\end{equation}
Then, we stack these token-block matrices vertically over the selected layers $\mathbf{L}^{\mathrm{sel}}=\{\ell_1,\ldots,\ell_{|\mathbf{L}^{\mathrm{sel}}|}\}$:
\begin{equation}\label{Feat-Relation-1}
    \mathbf{V}_{\boldsymbol{\theta}_i,b}(\mathbf{Q},\mathbf{A})
    =
    \big[
        (\mathbf{V}_{\boldsymbol{\theta}_i,b}^{\ell_1}(\mathbf{Q},\mathbf{A}))^{\top},
        \ldots,
        (\mathbf{V}_{\boldsymbol{\theta}_i,b}^{\ell_{|\mathbf{L}^{\mathrm{sel}}|}}(\mathbf{Q},\mathbf{A}))^{\top}
    \big]^{\top}
    \in \mathbb{R}^{|\mathbf{L}^{\mathrm{sel}}|r\times|\mathcal{I}_b|}.
\end{equation}
The block node for $b$ is then represented by $\mathrm{vec}(\mathbf{V}_{\boldsymbol{\theta}_i,b})\in\mathbb{R}^{|\mathbf{L}^{\mathrm{sel}}|r|\mathcal{I}_b|}$. To compute edges between block nodes, we use the $\ell_2$ distance:
$
    R_{i}^{\mathrm{feat}}(b,\tilde{b};\mathbf{Q},\mathbf{A})
    =
    \big\|
        \mathrm{vec}(\mathbf{V}_{\boldsymbol{\theta}_i,b})
        -
        \mathrm{vec}(\mathbf{V}_{\boldsymbol{\theta}_i,\tilde{b}})
    \big\|_2.
$

\textbf{Joint Relations.}
Finally, we define a fine-grained node for each $(\ell,t,b)$ triple, whose feature is the corresponding hidden-state block $\mathbf{h}_{\boldsymbol{\theta}_i,t}^{\ell}(\mathbf{Q},\mathbf{A})[\mathcal{I}_b]\in\mathbb{R}^{|\mathcal{I}_b|}$. This yields a fully connected graph over $|\mathbf{L}^{\mathrm{sel}}|\,r\,B$ nodes, with the relation between two such nodes given by
\begin{equation}\label{Joint-Relation}
    R_{i}^{\mathrm{joint}}\big((\ell,t,b),(\tilde{\ell},\tilde{t},\tilde{b});\mathbf{Q},\mathbf{A}\big)
    =
    s^{\mathrm{lyr}}(\ell,\tilde{\ell})\,s^{\mathrm{tok}}(t,\tilde{t})\,
    \big\|
        \mathbf{h}_{\boldsymbol{\theta}_i,t}^{\ell}(\mathbf{Q},\mathbf{A})[\mathcal{I}_b]
        -\mathbf{h}_{\boldsymbol{\theta}_i,\tilde{t}}^{\tilde{\ell}}(\mathbf{Q},\mathbf{A})[\mathcal{I}_{\tilde{b}}]
    \big\|_2,
\end{equation}
where $s^{\mathrm{lyr}}$ and $s^{\mathrm{tok}}$ are the layer- and token-distance decay factors defined in Eq.~\eqref{Layer-Relation} and Section~\ref{sec:token_observations}.

For each relation type, following the same procedure as Section~\ref{sec:token_observations}, we apply the same relation-weighted propagation and mean pooling as in Eq.~(\ref{Token-Aggregation}) to obtain the per-sample feature vectors $\mathbf{z}_{i}^{\mathrm{lyr}}(\mathbf{Q},\mathbf{A})$, $\mathbf{z}_{i}^{\mathrm{feat}}(\mathbf{Q},\mathbf{A})$, and $\mathbf{z}_{i}^{\mathrm{joint}}(\mathbf{Q},\mathbf{A})$, with full procedure in Appendix~\ref{app:obs_pipeline}.

\textbf{Empirical Observations and Implications.} We now have four relational features: $\mathbf{z}_{i}^{\mathrm{tok},\ell}$, $\mathbf{z}_{i}^{\mathrm{lyr}}$, $\mathbf{z}_{i}^{\mathrm{feat}}$, and $\mathbf{z}_{i}^{\mathrm{joint}}$. We follow the same setup of Section~\ref{sec:token_observations}, and report the observations as follows.

\noindent (1) \emph{Cross-model stability.} In Figure~\ref{fig:obs_alignment_detection}\textcolor{red}{a}, raw hidden representations yield mutual $k$NN alignments of $0.03$--$0.11$ across the five transfer pairs, whereas the four relational features yield consistently higher alignments: token-level relations reach $0.17$--$0.30$, joint relational structures $0.15$--$0.30$, within-token feature relations $0.07$--$0.21$, and cross-layer token relations $0.10$--$0.24$. Following the same experimental setup in Section~\ref{sec:token_observations}, the random-overlap baseline with $k=10$ and $n=5000$ is near $0.002$, so the relational alignments are roughly $35$ to $150$ times random neighborhood agreement, indicating non-trivial cross-model relational similarity across all four features.

\noindent (2) \emph{Cross-model transferability.} In Figure~\ref{fig:obs_alignment_detection}\textcolor{red}{b} and Figure~\ref{fig:obs_kde_scores}, the four relational features improve cross-model AUROC over the raw hidden representations, with a different ordering. Joint relational structures achieve the highest AUROC, around $0.69$–$0.73$, followed by within-token feature relations at $0.64$–$0.67$, token-level relations at $0.62$–$0.67$, and cross-layer token relations at $0.56$–$0.60$. Comparing against the two training-free baselines used in Section~\ref{sec:token_observations}, namely PPL ($0.63$) and Semantic Entropy ($0.75$), three of the four relational features already match or exceed the general-purpose PPL baseline, and the joint relational structures further close the gap to Semantic Entropy.

\textit{Implication.} Among the four relational features, joint relational structures over layers, tokens, and feature blocks achieve near-highest cross-model stability and the highest cross-model AUROC. This indicates that the relations across all three indices, when combined, carry information that is both transferable across LLMs and discriminative for hallucination detection. Notably, this is achieved with a deliberately coarse design where the edge weights are fixed handcrafted decay functions. Such a simple construction already delivers non-trivial cross-model transfer. This leaves substantial room for a learnable, input-dependent edge mechanism, which we develop in Section~\ref{sec:method}.

\section{\method{}: Relational Attention for Cross-Model Hallucination Detection}
\label{sec:method}

Building on the implication in Section~\ref{sec:observations}, we develop \method{}, which builds a unified layer-token-feature graph with position- and content-aware edges, and learns transferable detection signals through relation-aware attention and meta-learning. See Appendices~\ref{app:graph_construction}, \ref{app:relation_aware_attention}, \ref{app:meta_learning_details} for more details.

\textbf{Hidden Representation Restructuring.}
Different LLMs have different hidden dimensions $d_i$, so we first map each hidden representation $\mathbf{h}_{\boldsymbol{\theta}_i,t}^{\ell}\in\mathbb{R}^{d_i}$ into a unified space $\mathbb{R}^{d^{\star}}$ via scattered padding, where $d^{\star}\geq\max_i d_i$. Specifically, for each LLM $\mathbb{P}_{\boldsymbol{\theta}_i}$, we define an index map $p_i:\{1,\ldots,d_i\}\to\{1,\ldots,d^{\star}\}$ that distributes the $d_i$ original coordinates evenly across the padded dimension \citep{oppenheim1999discrete},
\begin{equation}\label{eq:index_map}
    p_i(\kappa) = \left\lfloor (\kappa-1)\,d^{\star}/d_i \right\rfloor + 1, ~\text{for}~ \kappa\in\{1,\ldots,d_i\},
\end{equation}
and obtain the padded representation $\bar{\mathbf{h}}_{t}^{\ell}\in\mathbb{R}^{d^{\star}}$ by
$\bar{\mathbf{h}}_{t}^{\ell}[p_i(\kappa)] = \mathbf{h}_{\boldsymbol{\theta}_i,t}^{\ell}[\kappa]$ {and}~$\bar{\mathbf{h}}_{t}^{\ell}[j]=0$, for $j\notin\mathrm{Im}(p_i)$,
where $\mathrm{Im}(p_i)$ denotes the set of positions reached by $p_i$. To prepare the unified vector for graph construction below, we partition the padded dimension into $B$ equal-size contiguous blocks:
\begin{equation}\label{eq:block_partition}
    \mathcal{I}_b = \big\{(b-1)\,d^{\star}/B+1,\,\ldots,\,b\,d^{\star}/B\big\}, ~\text{for}~b\in\{1,\ldots,B\}.
\end{equation}
We then use $\bar{\mathbf{h}}_{\boldsymbol{\theta}_i,t}^{\ell}[\mathcal{I}_b]\in\mathbb{R}^{d^{\star}/B}$ to represent the feature of the $b$-th block. 

\textbf{Graph Construction and Relation-Aware Encoding.} For a given sample generated by a specific LLM $\mathbb{P}_{\boldsymbol{\theta}_i}$, we construct an individual relation graph. Within this graph, we treat each $(\ell, t, b)$ triple with $\ell\in\mathbf{L}^{\mathrm{sel}}$, $t\in\mathbf{T}^{\mathrm{sel}}$, and $b\in\{1,\dots,B\}$ as a node, indexed by $u\in\{1,\dots,N\}$ with $N=|\mathbf{L}^{\mathrm{sel}}|\cdot r\cdot B$. We write $(\ell_u,t_u,b_u)$ for the layer, token, and block indices of $u$; the resulting graph is fully connected over the $N$ nodes. Each node $u$ takes the corresponding block feature $\bar{\mathbf{h}}_{t_u}^{\ell_u}[\mathcal{I}_{b_u}]$ and combines it with these three \textit{learnable position embeddings} of its layer, token, and block indices to form the initial node representation $\mathbf{x}_u = \mathbf{W}_{\mathrm{p}}\,\bar{\mathbf{h}}_{t_u}^{\ell_u}[\mathcal{I}_{b_u}] + \mathbf{e}_{\ell}(\ell_u) + \mathbf{e}_{t}(t_u) + \mathbf{e}_{b}(b_u) \in\mathbb{R}^{d_{\mathrm{m}}}$, where $\mathbf{W}_{\mathrm{p}}\in\mathbb{R}^{d_{\mathrm{m}}\times d^{\star}/B}$ is a learnable matrix that linearly projects the block feature to the detector's internal dimension $d_{\mathrm{m}}$, and $\mathbf{e}_{\ell}, \mathbf{e}_{t}, \mathbf{e}_{b}$ are learnable lookup tables~\citep{devlin2019bert} that map indices in $\{1,\ldots,|\mathbf{L}^{\mathrm{sel}}|\}, \{1,\ldots,r\}, \{1,\ldots,B\}$, respectively, to dense continuous vectors in $\mathbb{R}^{d_{\mathrm{m}}}$.

For each pair of nodes $(u, v)$, we let $\boldsymbol{\phi}_{\ell}, \boldsymbol{\phi}_{t}, \boldsymbol{\phi}_{b}$ be learnable lookup tables that map the layer, token, and block index differences $\ell_u-\ell_v, t_u-t_v, b_u-b_v$ to vectors, and define an edge feature that conveys to the attention both how the two nodes differ in position and how similar they are in content,
\begin{equation}\label{eq:edge_feature}
    \mathbf{e}_{uv} = \mathrm{MLP}_1\!\big(\big[\boldsymbol{\phi}_{\ell}(\ell_u-\ell_v),\,\boldsymbol{\phi}_{t}(t_u-t_v),\,\boldsymbol{\phi}_{b}(b_u-b_v),\,c_{uv},\,d_{uv}\big]\big) \in\mathbb{R}^{d_{\mathrm{m}}},
\end{equation}
where $c_{uv}=\cos(\mathbf{x}_u,\mathbf{x}_v)$, $d_{uv}=\|\mathbf{x}_u-\mathbf{x}_v\|_2$, and $\mathrm{MLP}_1$ is a learnable MLP that fuses these signals. 

At this point, we have a fully connected graph in which each node carries layer-token-block-aware features and each edge carries learned positional and content relations. To turn this graph into a single feature vector for hallucination detection, we proceed in two steps that mirror the relational findings in Section~\ref{sec:joint_relations}. \textit{First}, we propagate messages along the edges so that each node aggregates relational information from every other node, absorbing the edge features into updated node representations~\citep{gilmer2017neural, velivckovic2017graph}. This is exactly where the edge feature $\mathbf{e}_{uv}$ from Eq.~\eqref{eq:edge_feature} earns its purpose: by participating in the propagation, the positional and content relations encoded on each edge are written into the receiving node, so subsequent stages no longer need to access edges directly. We instantiate this propagation with a learnable relation-aware attention encoder~\citep{shaw2018self}:
$
    \{\mathbf{x}'_u\}_{u=1}^{N} = \mathrm{RA}\big(\{\mathbf{x}_u\}_{u=1}^{N},\,\{\mathbf{e}_{uv}\}_{u,v=1}^{N}\big).
$ \textit{Second}, with relations now stored inside the nodes, we collapse the $N$ updated nodes into a single graph-level vector via attention pooling~\citep{lin2017structured}:
$
    \mathbf{z} = \mathrm{AttPool}\big(\{\mathbf{x}'_u\}_{u=1}^{N}\big)\in\mathbb{R}^{d_{\mathrm{m}}}.
$ \textit{Last}, the truthfulness probability is given by $\hat{y}=\text{Sigmoid}\big(\mathrm{MLP}_{2}(\mathbf{z})\big)\in[0,1]$, where $\mathrm{MLP}_{2}$ is a small learnable MLP.

\textbf{Meta-Learning Training.}
We view the preceding relational pipeline as a parameterized detector $\hat{y}=D_{\Theta}(\mathbf{Q},\mathbf{A};\mathbb{P}_{\boldsymbol{\theta}})\in[0,1]$, where $\Theta$ collects all learnable parameters. Let $\ell^{\mathrm{bce}}$ denote the binary cross-entropy loss between $\hat{y}$ and the truthfulness label $y$, and recall that $\mathcal{D}_i$ (defined in Eq.~\eqref{dataset_format-i}) collects the labeled $(\mathbf{Q},\mathbf{A},y)$ triples produced by the $i$-th training LLM. A straightforward training choice would solve $\arg\min_{\Theta}\sum_{i=1}^{M}\sum_{(\mathbf{Q},\mathbf{A},y)\in\mathcal{D}_i}\ell^{\mathrm{bce}}(\hat{y};y)$ on the training datasets, but our actual goal is to make $\Theta$ generalize to LLMs unseen at training time, and such empirical risk minimization may exploit shortcuts specific to individual training LLMs. We therefore train $\Theta$ by following Feature-Critic~\citep{li2019feature}, a representative meta-learning strategy, abstracted as a procedure that takes the parameterized detector and the training datasets as input and returns updated parameters, $\Theta'\leftarrow\mathrm{FeatCrit}(D_{\Theta},\{\mathcal{D}_i\}_{i=1}^{M})$.
The intuition is to manufacture the missing test condition from the training data itself: at each step, $\mathrm{FeatCrit}$ holds out one training LLM as a pseudo-unseen model, runs candidate updates on the remaining LLMs, and rewards updates whose effect transfers to the held-out LLM. Over training, this drives $\Theta'$ toward signals shared across LLMs rather than those idiosyncratic to any one of them. The trained detector $D_{\Theta'}$ is then used at test time on unseen LLMs.

\begin{table*}[htbp]

\caption{
Ablation study of \method{} (AUROC). 
The comparison protocol follows Table~\ref{tab:main_cross_family_results}. 
In the \textit{Best training-free} and \textit{Best training-based} rows, each benchmark column reports the strongest baseline of that group on that benchmark, so different columns may come from different baselines; the QA Avg. and All Avg. columns report the averages of the single strongest baseline of the group, and are therefore not the row-wise means of the columns beside them.
Full results are provided in Appendix~\ref{app:additional_exp_analysis}.
Best results are \textbf{bold}, and second-best results are \underline{underlined}.
}
\vspace{2mm}
\centering
\label{tab:ablation_main}
\resizebox{\linewidth}{!}{%
\begin{tabular}{lccccccccccccccc}
\toprule
\multirow{4}{*}{\textbf{Method}} & \multicolumn{15}{c}{\textbf{Target Model}} \\
& \multicolumn{5}{c}{Llama3.2-3B} & \multicolumn{5}{c}{Llama3.1-8B} & \multicolumn{5}{c}{Llama3.3-70B} \\
\cmidrule(lr){2-6}\cmidrule(lr){7-11}\cmidrule(lr){12-16}
& \multicolumn{5}{c}{Mean AUROC$\uparrow$} & \multicolumn{5}{c}{Mean AUROC$\uparrow$} & \multicolumn{5}{c}{Mean AUROC$\uparrow$} \\
& QA Avg. & GSM8K & MBPP & xLAM & \textbf{All Avg.} & QA Avg. & GSM8K & MBPP & xLAM & \textbf{All Avg.} & QA Avg. & GSM8K & MBPP & xLAM & \textbf{All Avg.} \\
\midrule
Best training-free & 0.7031 & 0.7177 & \underline{0.6915} & 0.7253 & 0.7067 & 0.7119 & 0.7293 & \underline{0.6865} & 0.7469 & 0.7158 & 0.7372 & 0.7473 & \underline{0.7181} & 0.7477 & 0.7374 \\
Best training-based & 0.5209 & 0.5300 & 0.5035 & 0.5397 & 0.5208 & 0.5225 & 0.5382 & 0.4876 & 0.5395 & 0.5212 & 0.5278 & 0.5375 & 0.4905 & 0.5386 & 0.5232 \\
CHORD w/o RelAttn & 0.6901 & 0.7241 & 0.6458 & 0.7259 & 0.6937 & 0.7023 & 0.7403 & 0.6678 & 0.7916 & 0.7156 & 0.7373 & 0.7646 & 0.7017 & 0.7821 & 0.7425 \\
CHORD w/o Meta & \underline{0.7102} & \underline{0.7394} & 0.6563 & \underline{0.7594} & \underline{0.7137} & \underline{0.7278} & \underline{0.7538} & 0.6693 & \underline{0.7975} & \underline{0.7331} & \underline{0.7571} & \underline{0.8101} & 0.7151 & \underline{0.7886} & \underline{0.7632} \\
\rowcolor{SkyBlueRow} \textbf{CHORD} & \textbf{0.7901} & \textbf{0.8193} & \textbf{0.7351} & \textbf{0.8437} & \textbf{0.7941} & \textbf{0.8017} & \textbf{0.8362} & \textbf{0.7307} & \textbf{0.8650} & \textbf{0.8056} & \textbf{0.8302} & \textbf{0.8568} & \textbf{0.7605} & \textbf{0.8713} & \textbf{0.8299} \\
\midrule[1.5pt]
\multicolumn{16}{c}{} \\[-2mm]
& \multicolumn{5}{c}{Qwen3-8B} & \multicolumn{5}{c}{Qwen3-14B} & \multicolumn{5}{c}{Qwen3.5-35B} \\
\cmidrule(lr){2-6}\cmidrule(lr){7-11}\cmidrule(lr){12-16}
\midrule
Best training-free & 0.7061 & 0.7143 & \underline{0.6834} & 0.7483 & 0.7092 & 0.7053 & 0.7496 & 0.6904 & 0.7435 & 0.7150 & 0.7124 & 0.7438 & \underline{0.6902} & 0.7376 & 0.7173 \\
Best training-based & 0.5077 & 0.5167 & 0.4922 & 0.5384 & 0.5112 & 0.5228 & 0.5287 & 0.4980 & 0.5370 & 0.5212 & 0.5260 & 0.5301 & 0.4874 & 0.5392 & 0.5224 \\
CHORD w/o RelAttn & 0.6965 & 0.7233 & 0.6416 & 0.7602 & 0.7016 & 0.6831 & 0.7645 & 0.6652 & 0.7594 & 0.7031 & 0.7096 & 0.7609 & 0.6550 & 0.7640 & 0.7169 \\
CHORD w/o Meta & \underline{0.7108} & \underline{0.7444} & 0.6577 & \underline{0.7706} & \underline{0.7166} & \underline{0.7146} & \underline{0.7679} & \underline{0.6913} & \underline{0.7788} & \underline{0.7281} & \underline{0.7253} & \underline{0.7661} & 0.6515 & \underline{0.7641} & \underline{0.7262} \\
\rowcolor{SkyBlueRow} \textbf{CHORD} & \textbf{0.8005} & \textbf{0.8256} & \textbf{0.7481} & \textbf{0.8484} & \textbf{0.8034} & \textbf{0.8043} & \textbf{0.8387} & \textbf{0.7248} & \textbf{0.8722} & \textbf{0.8075} & \textbf{0.8133} & \textbf{0.8406} & \textbf{0.7207} & \textbf{0.8472} & \textbf{0.8088} \\
\midrule[1.5pt]
\multicolumn{16}{c}{} \\[-2mm]
& \multicolumn{5}{c}{Ministral3-8B} & \multicolumn{5}{c}{Ministral3-14B} & \multicolumn{5}{c}{\textbf{Target Avg.}} \\
\cmidrule(lr){2-6}\cmidrule(lr){7-11}\cmidrule(lr){12-16}
\midrule
Best training-free & 0.6991 & 0.7254 & \underline{0.6961} & 0.7446 & 0.7089 & 0.7061 & 0.7163 & \underline{0.6957} & 0.7432 & 0.7114 & 0.7102 & 0.7297 & \underline{0.6940} & 0.7421 & 0.7152 \\
Best training-based & 0.5143 & 0.5335 & 0.5030 & 0.5392 & 0.5176 & 0.5035 & 0.5284 & 0.4924 & 0.5391 & 0.5090 & 0.5173 & 0.5276 & 0.4927 & 0.5387 & 0.5177 \\
CHORD w/o RelAttn & 0.6855 & 0.7148 & 0.6591 & 0.7514 & 0.6953 & 0.6988 & 0.6829 & 0.6463 & 0.7328 & 0.6939 & 0.7004 & 0.7344 & 0.6603 & 0.7584 & 0.7078 \\
CHORD w/o Meta & \underline{0.7024} & \underline{0.7574} & 0.6821 & \underline{0.7738} & \underline{0.7176} & \underline{0.7222} & \underline{0.7186} & 0.6456 & \underline{0.7594} & \underline{0.7160} & \underline{0.7213} & \underline{0.7572} & 0.6711 & \underline{0.7740} & \underline{0.7268} \\
\rowcolor{SkyBlueRow} \textbf{CHORD} & \textbf{0.7854} & \textbf{0.8021} & \textbf{0.7263} & \textbf{0.8474} & \textbf{0.7882} & \textbf{0.7932} & \textbf{0.8193} & \textbf{0.7237} & \textbf{0.8557} & \textbf{0.7959} & \textbf{0.8023} & \textbf{0.8298} & \textbf{0.7337} & \textbf{0.8564} & \textbf{0.8042} \\
\bottomrule
\end{tabular}%
}
\vspace{-4mm}
\end{table*}

\section{Experiments}
\label{sec:exp}

\label{sec:exp_setups}

\textbf{Experimental Setups.} We evaluate \method{} on eight open-source LLMs from the Llama, Qwen, and Ministral families \citep{grattafiori2024llama, yang2025qwen3, liu2026ministral}, spanning different scales. 
We use seven benchmarks: TruthfulQA, HaluEval, TriviaQA, CoQA, GSM8K, MBPP, and xLAM \citep{lin2022truthfulqa, li-etal-2023-halueval, joshi-etal-2017-triviaqa, reddy-etal-2019-coqa, cobbe2021training, austin2021program, liu2024apigen}, covering factual QA, hallucination detection, conversational QA, reasoning, coding, and agent tasks. 
Baselines include Perplexity, Semantic Entropy, EigenScore, Lexical Similarity, Verbalize, InterrogateLLM, MM, SEP, SAPLMA, and LapEigvals \citep{ren2023outofdistribution, farquhar2024detecting, chen2024inside, lin2024generating, lin2022teaching, yehuda-etal-2024-interrogatellm, marks2023geometry, kossen2024semantic, azaria2023internal, binkowski-etal-2025-hallucination}. 
For training-based methods requiring fixed dimensions, we use the restructuring procedure in Section~\ref{sec:method} to map hidden states into a common space. 
Sampling strategies, model details, baseline configurations, extraction details, and hyperparameters are in Appendix~\ref{app:exp_details}.\looseness=-1

\label{sec:exp_results}

\textbf{Overall Cross-model Hallucination Detection Performance.}
Table~\ref{tab:main_cross_family_results} evaluates transfer to a held-out model family, with no target-family examples used for detector training.
Averaged equally over the eight target models and seven benchmarks, \method{} reaches an AUROC of $0.8042$, compared with $0.7152$ for InterrogateLLM, the strongest baseline by overall average.
It also ranks first in all 56 model--benchmark evaluations, so the aggregate gain is not driven by a few favorable targets.
In contrast, MM, SEP, SAPLMA, and LapEigvals obtain overall averages of $0.4976$--$0.5177$.
Their weak transfer despite dimension-compatible inputs shows that matching feature dimensions alone does not resolve cross-model generalization.
Training-free methods avoid fitting a source-specific detector and remain competitive, but \method{} demonstrates that source supervision can provide additional value when combined with relational features and transfer-oriented training.

\textbf{Dataset-level Differences.}
The gains are not uniform across tasks.
On xLAM and GSM8K, \method{} achieves target-averaged AUROCs of $0.8564$ and $0.8298$, compared with $0.7421$ and $0.7297$ for InterrogateLLM.
The advantage also extends to standard QA: its average over TruthfulQA, HaluEval, TriviaQA, and CoQA is $0.8023$, versus $0.7102$ for the same baseline.
Thus, the benefit is not confined to reasoning or tool-use outputs.
MBPP presents a clearer boundary: \method{} reaches $0.7337$, but its margin over InterrogateLLM ($0.6940$) is much smaller.
Limited supervision and execution-dependent code correctness may contribute to this gap, although these results do not isolate their effects.
The task differences suggest that relational features provide broadly useful evidence, but should not be assumed to capture all forms of answer correctness equally well.

\textbf{Variation across Target Models.}
The advantage extends beyond the largest targets.
Across eight models in Table~\ref{tab:main_cross_family_results}, the benchmark-averaged margin over InterrogateLLM ranges from $0.0793$ to $0.0942$.
Within Llama, \method{} improves from $0.7941$ on 3B to $0.8299$ on 70B, while the Qwen targets occupy a narrower range of $0.8034$--$0.8088$.
Thus, larger targets often provide stronger detection performance, but the magnitude of this change varies across families.
Since model versions and architectures change, these results describe the evaluated targets rather than establish a general scaling law.\looseness=-1

\textbf{Ablation Study.}
Table~\ref{tab:ablation_main} separates the contributions of relation-aware attention and meta-learning.
Removing relation-aware attention reduces overall AUROC from $0.8042$ to $0.7078$, while replacing meta-learning with standard ERM reduces it to $0.7268$.
These are drops of $9.64$ and $7.74$ AUROC percentage points, respectively.
The first variant retains the remaining framework, indicating that the gain cannot be attributed solely to hidden-state restructuring or graph-node construction.
The second retains the relational architecture, showing that source-family supervision alone does not recover the full transfer benefit.
Both variants remain above ordinary training-based baselines, but their advantage over training-free detection is task-dependent.
On MBPP, their averages of $0.6603$ and $0.6711$ fall below InterrogateLLM's $0.6940$, whereas full \method{} reaches $0.7337$.
This contrast supports using both components rather than relying on either modification alone.

\textbf{Contributions of Individual Relation Channels.}
The finer-grained ablations in Appendix~\ref{app:exp_edge_ablation} remove one edge channel while preserving the other channels and training procedure.
Removing token, layer, or intra-block feature edges reduces overall AUROC to $0.7592$, $0.7629$, or $0.7607$, respectively.
Their relative importance varies by task: token-edge removal produces the largest loss among the three variants on xLAM, layer-edge removal does so on GSM8K, and intra-block feature-edge removal does so on MBPP.
Thus, no single channel accounts for the best performance across all tasks.
Each removal is also less damaging than removing relation-aware attention entirely, consistent with the remaining channels retaining useful detection evidence.

\textbf{Transfer across Model Scales.}
Appendix~\ref{app:exp_scale_group_transfer} evaluates six directed transfers between small, medium, and large model groups.
\method{} ranks first in each direction, with average AUROC ranging from $0.7664$ to $0.7854$.
Large-to-small transfer reaches $0.7854$, compared with $0.7664$ in the reverse direction, showing that transfer difficulty can be asymmetric.
However, these groups differ in model-family composition and source-model count.
The results therefore support transfer across the evaluated scale groups, rather than a causal claim that larger source models always provide better supervision.\looseness=-1

\textbf{Distinguishing Model Shift from Benchmark Shift.}
Appendix~\ref{app:exp_cross_dataset_transfer} holds the LLM fixed, trains on six benchmarks, and evaluates on the seventh.
\method{} reaches an overall AUROC of $0.7800$, compared with $0.7152$ for InterrogateLLM.
More importantly, ordinary training-based baselines recover to $0.6366$--$0.6797$, rather than remaining near chance as in cross-family transfer.
This contrast indicates that their weakness in Table~\ref{tab:main_cross_family_results} is not a general inability to learn hallucination signals: they retain useful discrimination when the model's representation space is shared across training and testing.
Because the training pools also differ, the comparison supports rather than proves the representation-mismatch explanation.
These experiments test two distinct shifts and should not be read as evidence for simultaneous transfer to an unseen model and an unseen benchmark.

\section{Conclusion}
\label{sec:conclusion}

In this paper, we study \emph{cross-model hallucination detection} (CMHD), which aims to train hallucination detectors on source LLMs and generalize them to unseen target LLMs. 
Through cross-model analysis, we identify layer-token-feature relational structures as transferable hallucination signals. 
We propose \method{}, a relational graph discrimination framework that learns structure-aware features through relation-aware attention and meta-learning. 
Experiments across model families, scales, and benchmarks show \method{} outperforms representative baselines, while ablations verify core modules.

\clearpage

\begin{ack}
This work was supported in part by the 2026 KTP Seed Funding
from the Faculty of Engineering and Information Technology,
University of Technology Sydney, and by the Australian Government
through the Australian Research Council (ARC) under a Discovery
Early Career Researcher Award (DE250100363) and a Future Fellowship
(FT230100121).

\textit{The views and conclusions in this paper reflect the authors'
work and should not be taken as official positions of the Australian
Government, the ARC, or any other funding body.}
\end{ack}

The authors declare no competing interests.

\bibliographystyle{unsrtnat}
\bibliography{references}

\newpage
\appendix

\startcontents[appendix]

\begingroup
\hypersetup{linkcolor=black}
\section*{Appendix Contents}
\printcontents[appendix]{}{1}{\setcounter{tocdepth}{2}}
\endgroup
\clearpage

\section{Experimental Setup Details and Hyperparameter Selection}

\label{app:exp_details}

\subsection{Pilot Transfer-failure Analysis}
\label{app:pilot}

\begin{figure}[htbp]
  \centering
  \includegraphics[width=\linewidth]{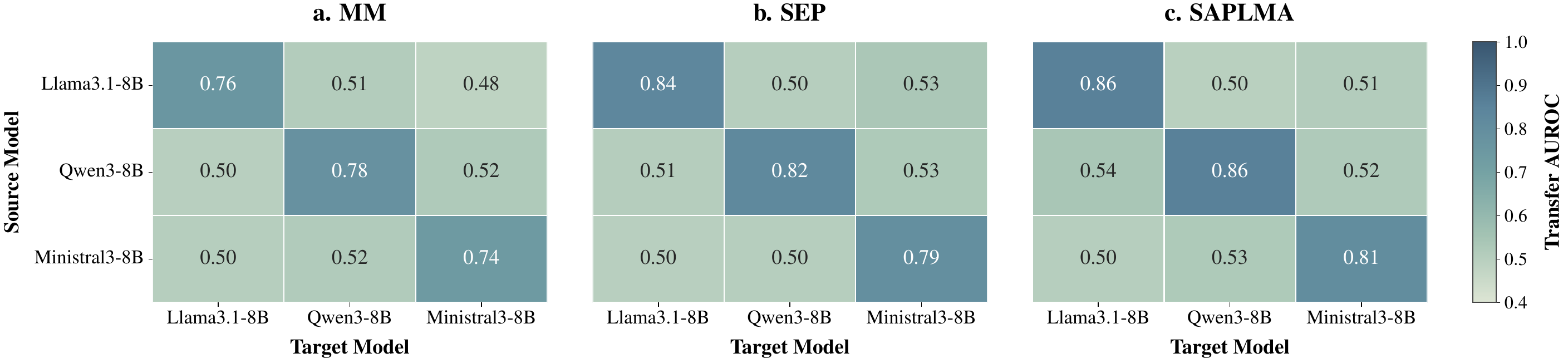}

    \caption{
    Pilot analysis of cross-model transfer failure in hallucination detection. 
    \textbf{(a-c)} Transfer AUROC matrices for representative training-based methods (MM \citep{marks2023geometry}, SEP \citep{kossen2024semantic}, SAPLMA \citep{azaria2023internal}). While all methods achieve strong in-model performance (diagonal), they suffer severe degradation when across models (off-diagonal), dropping to near-random accuracy. 
    See Appendix~\ref{app:pilot} for details.
    }

  \label{fig_intro_observation}
\end{figure}

Figure~\ref{fig_intro_observation} reports a small pilot study used to verify whether standard training-based hallucination detectors transfer across LLMs. 
We consider three 8B-scale models, Llama3.1-8B, Qwen3-8B, and Ministral3-8B, whose hidden-state dimensions are the same in this pilot; therefore, no cross-model dimension mapping is applied. 
For each model, we construct a mixed benchmark pool from TruthfulQA~\citep{lin2022truthfulqa}, HaluEval~\citep{li-etal-2023-halueval}, TriviaQA~\citep{joshi-etal-2017-triviaqa}, CoQA~\citep{reddy-etal-2019-coqa}, GSM8K~\citep{cobbe2021training}, MBPP~\citep{austin2021program}, and xLAM~\citep{liu2024apigen}. 
We sample 1,000 examples from each benchmark whenever possible, and use all available examples for MBPP because it contains fewer than 1,000 usable samples in our setting. 
The mixed pool is split into 20\% training and 80\% evaluation data. 
For each source model, we train MM, SEP, and SAPLMA on its training split and evaluate the trained detector on the evaluation splits of all target models. 
The diagonal entries therefore measure in-model performance, while the off-diagonal entries measure direct cross-model transfer. 
All three methods achieve strong in-model AUROC on the diagonal, but their off-diagonal AUROC drops close to random guessing. 
This pilot result motivates the main problem studied in this paper: supervised detectors can learn useful model-specific decision rules, but these rules do not transfer reliably across heterogeneous LLM representation spaces.

\subsection{Token--Layer Window Size and Position}
\label{app:obs_token_layer_window}

The relational features studied in Section~\ref{sec:observations} depend on which generated tokens and Transformer layers are selected. 
The same choice also affects \method{}, since its graph is constructed over selected layer--token--block nodes. 
We therefore study two aspects of this design: the number of selected tokens/layers and the position of the selected window. 
Throughout this section, token windows are selected from generated answer tokens, and layer windows are selected from Transformer layers.

\paragraph{Evaluation settings.}
We consider two settings. 
The first is the fixed \emph{observational diagnostic} setting used in Appendix~\ref{app:obs_alignment_transfer_decoupling}: joint relational features are constructed with hand-designed edges, an MLP detector is trained on a source model, and transfer AUROC is evaluated on a target model. 
The second is the \emph{method} setting: the full \method{} detector is trained and evaluated under the leave-one-family-out protocol used in the main experiments. 
These two settings use the same kind of token--layer selection, but they serve different purposes: the diagnostic setting exposes transferable relational structure, while the method setting measures the end-to-end detector.

\paragraph{Count sweeps.}
We first vary the number of selected tokens and layers while keeping their positions fixed. 
Following common practice that late generated tokens contain more answer-level evidence and middle layers provide more transferable semantic structure, we use late token windows and middle layer windows for the count sweeps. 
For the observational diagnostic, the token-count sweep varies $T\in\{1,\ldots,10\}$ with $L=5$ fixed, and the layer-count sweep varies $L\in\{1,\ldots,16\}$ with $T=7$ fixed. 
For \method{}, we use the same sweep ranges, while fixing the other axis at the method default.

\begin{figure}[t]
    \centering
    \includegraphics[width=\linewidth]{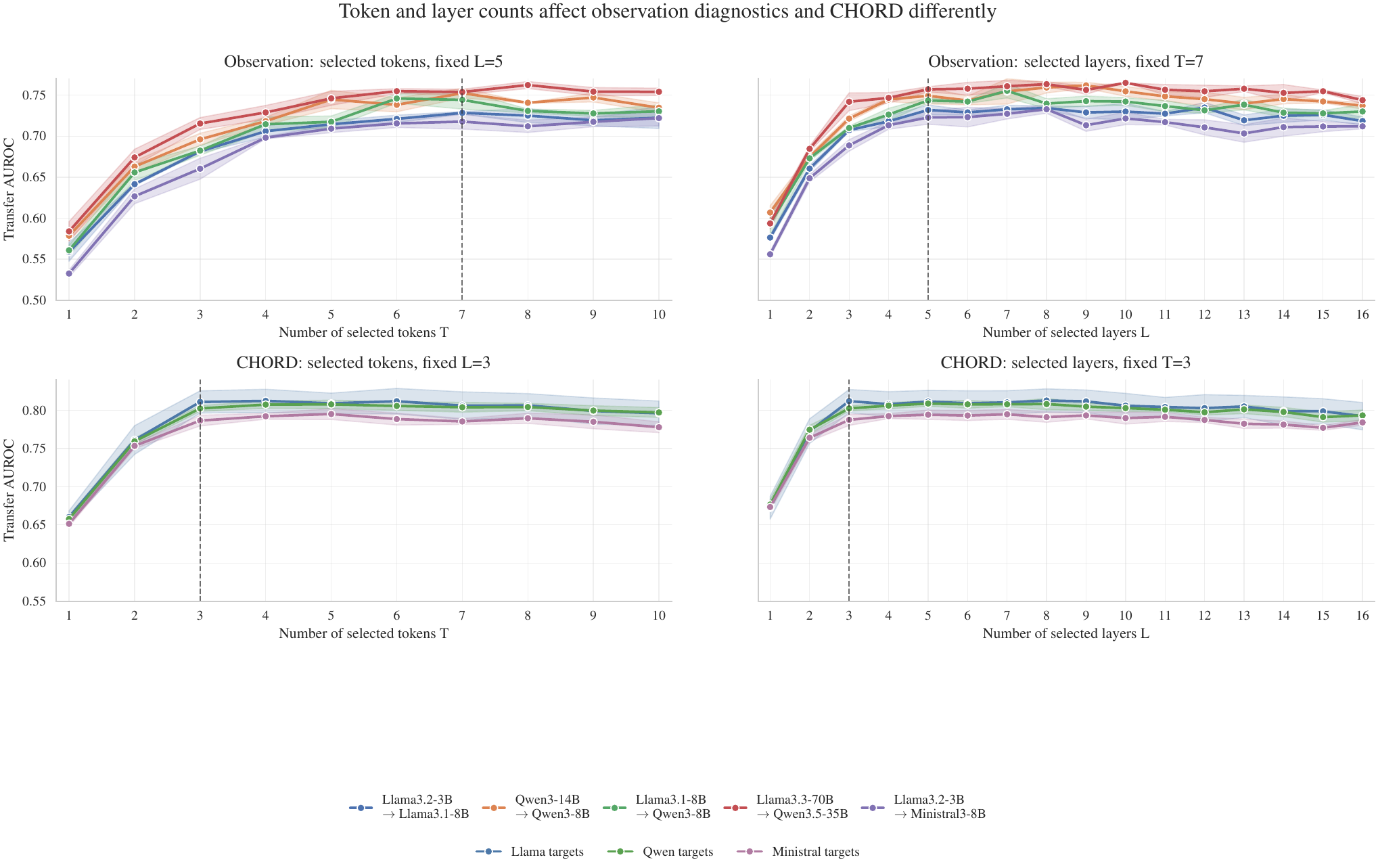}
    \caption{
    Token and layer count sweeps for the observational diagnostic and the full \method{} method. 
    The diagnostic setting benefits from a moderately larger window, reaching its strongest region around $T=7$ and $L=5$. 
    \method{} recovers most of its performance with a compact $T=3,L=3$ setting, after which larger windows provide only limited additional gain.
    }
    \label{fig:app_window_count_sweeps}
\end{figure}

Figure~\ref{fig:app_window_count_sweeps} shows a clear difference between the fixed diagnostic features and the full method. 
For the observational diagnostic, using only one token or one layer leaves little relational structure, and transfer AUROC is low. 
Performance improves as the window grows and is strongest around $T=7$ and $L=5$, after which the curve largely saturates. 
For \method{}, the curve rises more quickly and reaches a stable region around $T=3,L=3$. 
This suggests that fixed handcrafted relational features benefit from a broader window, whereas the learnable relation-aware encoder can extract most of the useful signal from a smaller graph.

\paragraph{Position sweeps.}
We next fix the window size and vary its position. 
For the observational diagnostic, we use the stronger count-sweep setting $T=7,L=5$. 
For \method{}, we use the default method setting $T=3,L=3$. 
When comparing token positions, we keep the layer window in the middle layers and place the token window at the early, middle, or late part of the generated answer. 
When comparing layer positions, we keep the token window at the late generated tokens and place the layer window at early, middle, or late layers.

\begin{figure}[t]
    \centering
    \includegraphics[width=\linewidth]{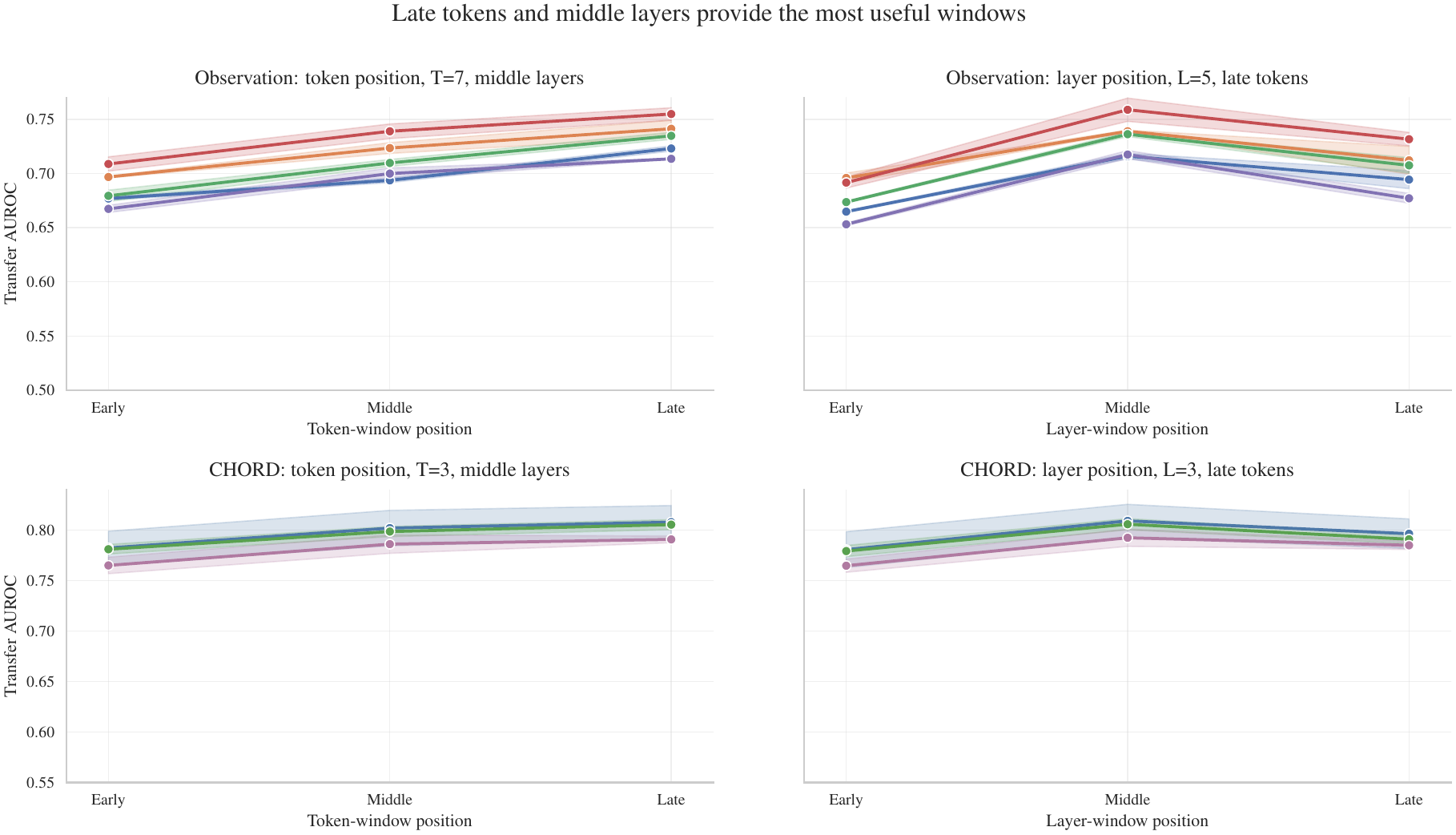}
    \caption{
    Token and layer position sweeps. 
    Late generated tokens perform best across both the observational diagnostic and \method{}. 
    Middle layers are also consistently strong, supporting the use of late-token and middle-layer windows in the main experiments.
    }
    \label{fig:app_window_position_sweeps}
\end{figure}

Figure~\ref{fig:app_window_position_sweeps} shows that late generated tokens are more useful than early or middle generated tokens. 
This is consistent with the fact that hallucination evidence often becomes clearer after more of the answer has been generated. 
For layers, middle-layer windows perform best or near-best. 
Early layers are closer to lexical encoding, while very late layers can be more tied to model-specific output behavior. 
Middle layers provide a more stable region for relational transfer.

\paragraph{Resulting choices.}
Based on these sweeps, the observational diagnostics use $T=7$ late generated tokens and $L=5$ middle layers. 
The full \method{} method uses the more compact $T=3,L=3$ setting with the same late-token and middle-layer positions. 
This gives the diagnostics enough context to reveal transferable relational structure, while keeping the final method efficient and avoiding unnecessary graph expansion.

\subsection{Feature-Block Granularity}
\label{app:feature_block_granularity}

Besides the token and layer windows, the joint relational graph also depends on how the hidden dimension is partitioned into feature blocks. 
In the main experiments, hidden states from all LLMs are first mapped to a common padded dimension $d_{\max}=8192$, and this dimension is divided into $B=64$ contiguous feature blocks. 
Each block then serves as the feature component of a joint node. 
This appendix studies the effect of varying the block count $B$.

\paragraph{Block construction.}
For a given value of $B$, we partition the padded feature indices into $B$ disjoint contiguous blocks:
\begin{equation}
    \{1,\ldots,d_{\max}\}
    =
    \mathcal{I}^{(B)}_1
    \cup
    \cdots
    \cup
    \mathcal{I}^{(B)}_B,
    \qquad
    \mathcal{I}^{(B)}_b\cap \mathcal{I}^{(B)}_{\tilde b}=\emptyset
    \ \text{for } b\neq \tilde b,
\end{equation}
where each block has dimension
\begin{equation}
    d_B = \frac{d_{\max}}{B}.
\end{equation}
For the joint relational graph, each node corresponds to a triple $(\ell,t,b)$, where $\ell$ is a selected layer, $t$ is a selected generated-token position, and $b$ is a feature block. 
The node feature is the corresponding block of the hidden state,
\[
    \mathbf{h}_{\boldsymbol{\theta}_i,t}^{\ell}(\mathbf{Q},\mathbf{A})[\mathcal{I}^{(B)}_b].
\]
Changing $B$ therefore changes both the granularity of each node and the graph size. 
When $B$ is small, each node contains a high-dimensional block and may mix unrelated coordinates. 
When $B$ is very large, each node contains very few coordinates; at $B=1024$, each block has only eight dimensions, so the graph becomes close to coordinate-level and each node carries little standalone information.

\paragraph{Evaluation settings.}
We evaluate
\[
    B\in\{2,4,8,16,32,64,128,256,512,1024\}.
\]
The sweep is conducted in two settings. 
For the observational diagnostic, we use the fixed joint relational structure from Section~\ref{sec:observations}, with $T=7$ late generated tokens and $L=5$ middle layers. 
For the full method, we use \method{} with its default compact window, $T=3$ late generated tokens and $L=3$ middle layers. 
All other components are kept unchanged, so the only varied factor is the feature-block count $B$.

The number of graph nodes grows linearly with $B$. 
In the observational setting, the joint graph contains $7\times 5\times B=35B$ nodes. 
In the method setting, it contains $3\times 3\times B=9B$ nodes. 
Thus, increasing $B$ not only changes the feature granularity but also increases the cost of relation aggregation. 
This makes the sweep a direct test of the signal--compute trade-off induced by feature-block partitioning.

\begin{figure}[t]
    \centering
    \includegraphics[width=\linewidth]{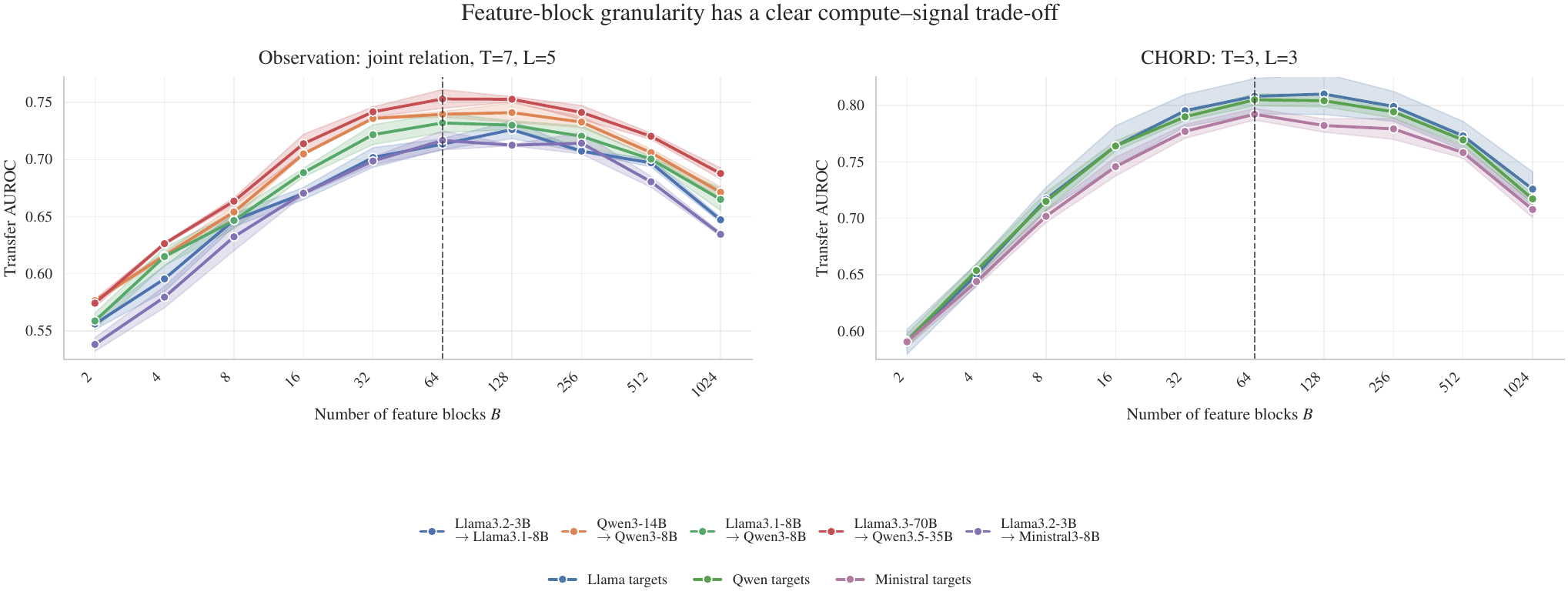}
    \caption{
    Effect of feature-block count $B$ on the observational joint-relation diagnostic and the full \method{} method. 
    Hidden states are padded to $d_{\max}=8192$ and partitioned into $B$ contiguous blocks. 
    Very small $B$ produces overly coarse blocks, while very large $B$ makes each node too low-dimensional and greatly increases graph size. 
    The default $B=64$ provides a strong trade-off between transfer performance and graph complexity.
    }
    \label{fig:app_block_count_sweeps}
\end{figure}

\paragraph{Results.}
Figure~\ref{fig:app_block_count_sweeps} shows a clear non-monotonic trend. 
When $B$ is very small, performance is low in both settings. 
This is expected because each feature block is too coarse. 
For example, $B=2$ or $B=4$ means that each node contains thousands of hidden dimensions, so relations between feature blocks cannot isolate more localized activation patterns. 
As $B$ increases from $2$ to $32$, transfer AUROC improves rapidly, indicating that finer feature partitioning exposes more useful internal relational structure.

The best region appears around $B=64$, with $B=128$ remaining close in performance. 
This suggests that the method does not rely on a single fragile block count, but benefits from a moderate block granularity. 
At $B=64$, each block has dimension $8192/64=128$, which is small enough to provide localized feature structure while still large enough for each node to contain meaningful multi-dimensional information. 
Larger values such as $B=128$ and $B=256$ can preserve similar accuracy, but they increase the graph size and therefore provide a less favorable compute--performance trade-off.

For very large $B$, the performance starts to decline. 
At the largest setting we evaluate, $B=1024$, each block has only eight dimensions and the graph is an order of magnitude larger than at the default. 
In this regime, each node carries too little information, and the relational graph becomes dominated by fine coordinate-level fragments rather than stable feature-block relations. 
The corresponding transfer AUROC falls well below the peak, confirming that simply increasing the number of nodes is not beneficial.

\paragraph{Implication.}
The block-count sweep supports the choice of $B=64$ in the main experiments. 
A very small $B$ loses fine-grained feature relations, while a very large $B$ produces low-information nodes and excessive graph expansion. 
The default $B=64$ lies near the strongest region of the curve and keeps the method computationally manageable. 
Thus, the gain of \method{} is not obtained by making the graph arbitrarily large; it comes from using a block granularity that preserves enough information per node while still exposing relational structure across tokens, layers, and feature blocks.

\subsection{Why We Use Mutual $k$-Nearest-Neighbor Alignment}
\label{app:knn_alignment}

We use mutual $k$-nearest-neighbor ($k$NN) alignment to quantify whether two LLMs induce similar sample-level structures after applying a given feature construction. 
Our choice follows the representation-alignment view in \citet{huh2024position}: a representation can be characterized by the similarity structure, or kernel, that it induces over data samples, and representational alignment measures the agreement between such induced similarity structures. 
Under this view, centered kernel alignment (CKA), SVCCA, and nearest-neighbor-based metrics can all be regarded as different ways of comparing representation-induced kernels.

A natural choice would be CKA, which compares two centered kernel matrices globally. 
However, this global comparison can be unnecessarily strict in our setting. 
Cross-model hallucination detection does not require two heterogeneous LLMs to have globally identical representation geometry. 
Different LLMs may differ in architecture, hidden dimension, scale, coordinate gauge, and model-specific semantic organization. 
What matters for detector transfer is instead whether the local neighborhood around each example is preserved: examples that are similar under the source-model feature should remain similar under the target-model feature. 
Such local preservation indicates that the feature carries a transferable sample structure even if the global geometry is not fully aligned.

This locality consideration is also consistent with the analysis of \citet{huh2024position}. 
They start from CKA as a global kernel-alignment measure, but then consider a nearest-neighbor-restricted variant, CKNNA, which computes kernel alignment only over samples that are local neighbors. 
When the neighborhood size approaches the full batch, this variant recovers standard CKA; when the neighborhood is small, it closely resembles mutual nearest-neighbor alignment. 
Their empirical analysis further shows that emphasizing local neighborhoods can reveal clearer alignment trends, and their appendix compares mutual $k$NN with CKA, unbiased CKA, SVCCA, CKNNA, and other nearest-neighbor variants. 
Thus, mutual $k$NN is not an ad hoc proxy, but a lightweight local-alignment measure supported by the broader kernel-alignment framework.

Formally, given two feature maps induced by LLMs $i$ and $j$ on the same $n$ examples, let $\mathcal{N}_{k}^{i}(s)$ and $\mathcal{N}_{k}^{j}(s)$ denote the $k$ nearest neighbors of sample $s$ under the two feature spaces, respectively. 
We compute neighbors using the inner-product similarity of the corresponding feature vectors. 
The per-sample neighborhood agreement is
\begin{equation}
    a_{ij}(s)
    =
    \frac{1}{k}
    \left|
    \mathcal{N}_{k}^{i}(s)
    \cap
    \mathcal{N}_{k}^{j}(s)
    \right|,
\end{equation}
and the reported alignment score is the average
\begin{equation}
    A_{ij}
    =
    \frac{1}{n}
    \sum_{s=1}^{n}
    a_{ij}(s).
\end{equation}
The score lies in $[0,1]$, where larger values indicate stronger agreement between the local neighborhood structures induced by the two LLMs.

We set $k=10$ in all experiments. 
This value is small relative to our evaluation set size ($n=5000$), so the metric remains strongly local: if two $k$NN sets were selected independently at random from the remaining $n-1$ examples, the expected normalized overlap would be approximately $k/(n-1)\approx 0.002$. 
At the same time, $k=10$ is large enough to avoid the instability of extremely small neighborhoods, such as $k=1$, which can be sensitive to noise, ties, or minor representation perturbations. 
This setting also matches the scale commonly used in nearest-neighbor alignment analyses such as \citet{huh2024position}. 
We use the same $k$ for all model pairs and feature types, and do not tune it using target-model performance.

\subsection{Graph Construction Details}
\label{app:graph_construction}

This subsection provides the implementation details of the graph construction used in Section~\ref{sec:method}. 
The purpose of this step is to convert hidden states from heterogeneous LLMs into a unified layer-token-feature graph on which relation-aware attention can be applied.

\paragraph{Scattered padding for dimension compatibility.}
Different LLMs may have different hidden dimensions $d_i$. 
To make their hidden states compatible with a shared graph encoder, we map every hidden state into a common ambient space $\mathbb{R}^{d^\star}$, where $d^\star\geq \max_i d_i$. 
For model $\mathbb{P}_{\boldsymbol{\theta}_i}$, the deterministic index map
\begin{equation}
    p_i(\kappa)
    =
    \left\lfloor
    (\kappa-1)d^\star/d_i
    \right\rfloor+1,
    \qquad
    \kappa\in\{1,\ldots,d_i\},
\end{equation}
assigns each original coordinate to a unique position in the padded dimension. 
The padded hidden state is defined as
\begin{equation}
    \bar{\mathbf{h}}_{\boldsymbol{\theta}_i,t}^{\ell}[p_i(\kappa)]
    =
    \mathbf{h}_{\boldsymbol{\theta}_i,t}^{\ell}[\kappa],
    \qquad
    \bar{\mathbf{h}}_{\boldsymbol{\theta}_i,t}^{\ell}[j]
    =
    0
    \quad
    \text{for }
    j\notin \mathrm{Im}(p_i),
\end{equation}
where $\mathrm{Im}(p_i)$ is the set of padded coordinates reached by $p_i$. 
This mapping is deterministic and parameter-free. 
It is not intended to semantically align individual hidden coordinates across LLMs; rather, it provides a common input dimension so that all models can be processed by the same downstream graph encoder. 
The scattered placement avoids concentrating all zero padding at the end of the vector and gives every feature block a more balanced mixture of original and padded coordinates.

\paragraph{Feature-block partition.}
After padding, the common dimension is partitioned into $B$ equal-size contiguous blocks:
\begin{equation}
    \mathcal{I}_b
    =
    \left\{
    (b-1)d^\star/B+1,\ldots,bd^\star/B
    \right\},
    \qquad
    b\in\{1,\ldots,B\}.
\end{equation}
We choose $d^\star$ to be divisible by $B$. 
In our implementation, we set $d^\star$ to the smallest multiple of $B$ that is no smaller than the largest hidden dimension among the evaluated LLMs. 
The block representation of token $t$ at layer $\ell$ is then
\begin{equation}
    \bar{\mathbf{h}}_{\boldsymbol{\theta}_i,t}^{\ell}[\mathcal{I}_b]
    \in
    \mathbb{R}^{d^\star/B}.
\end{equation}
The number of feature blocks $B$ controls the granularity of the feature-axis decomposition; its selection is discussed in Appendix~\ref{app:feature_block_granularity}.

\paragraph{Selected layer-token window.}
For each input $(\mathbf{Q},\mathbf{A})$, we select a token subset
$\mathbf{T}^{\mathrm{sel}}=\{t_1,\ldots,t_r\}$ and a layer subset
$\mathbf{L}^{\mathrm{sel}}_i=\{\ell_1,\ldots,\ell_m\}$ from model $\mathbb{P}_{\boldsymbol{\theta}_i}$. 
The same numbers $r=|\mathbf{T}^{\mathrm{sel}}|$ and $m=|\mathbf{L}^{\mathrm{sel}}_i|$ are used for all LLMs, while the actual layer indices are chosen by relative depth so that models with different numbers of layers can be compared. 
The token and layer selection strategy is described in Appendix~\ref{app:obs_token_layer_window}. 
For the final detector, we use the compact setting $r=3$ and $m=3$ unless otherwise stated.

\paragraph{Node set.}
The graph contains one node for every selected layer-token-block triple:
\begin{equation}
    u
    \leftrightarrow
    (\ell_u,t_u,b_u),
    \qquad
    \ell_u\in\mathbf{L}^{\mathrm{sel}}_i,\quad
    t_u\in\mathbf{T}^{\mathrm{sel}},\quad
    b_u\in\{1,\ldots,B\}.
\end{equation}
Thus, the number of nodes is
\begin{equation}
    N
    =
    |\mathbf{L}^{\mathrm{sel}}_i|
    \cdot
    |\mathbf{T}^{\mathrm{sel}}|
    \cdot
    B
    =
    m r B.
\end{equation}
For the default setting $m=3$, $r=3$, and $B=64$, the graph has $N=576$ nodes.

\paragraph{Initial node representation.}
For node $u=(\ell_u,t_u,b_u)$, the raw block feature is
\begin{equation}
    \bar{\mathbf{h}}_{\boldsymbol{\theta}_i,t_u}^{\ell_u}[\mathcal{I}_{b_u}]
    \in
    \mathbb{R}^{d^\star/B}.
\end{equation}
We project this block feature into the detector dimension and add learnable index embeddings for the layer, token, and block identities:
\begin{equation}
    \mathbf{x}_u
    =
    \mathbf{W}_{\mathrm{p}}
    \bar{\mathbf{h}}_{\boldsymbol{\theta}_i,t_u}^{\ell_u}[\mathcal{I}_{b_u}]
    +
    \mathbf{e}_{\ell}(\ell_u)
    +
    \mathbf{e}_{t}(t_u)
    +
    \mathbf{e}_{b}(b_u)
    \in
    \mathbb{R}^{d_{\mathrm{m}}}.
\end{equation}
Here $\mathbf{W}_{\mathrm{p}}\in\mathbb{R}^{d_{\mathrm{m}}\times d^\star/B}$ is a learnable projection matrix, and
$\mathbf{e}_{\ell}$, $\mathbf{e}_{t}$, and $\mathbf{e}_{b}$ are learnable lookup tables. 
The projection encodes the hidden-state block content, while the three embeddings tell the graph encoder where the node lies in the layer-token-feature grid.

\paragraph{Directed fully connected edge set.}
We construct a directed fully connected graph over the $N$ nodes. 
That is, every ordered pair $(u,v)$ is allowed to participate in relation-aware attention. 
Following standard self-attention, we include self-loops; when $u=v$, the relative offsets are zero and the content similarity terms take their self-comparison values. 
The graph connectivity is therefore not learned or sparsified. 
What is learned is the representation of each edge and how the attention encoder uses it.

\paragraph{Edge feature.}
For each directed pair $(u,v)$, we compute relative offsets along the three graph axes:
\begin{equation}
    \Delta^\ell_{uv}=\ell_u-\ell_v,\qquad
    \Delta^t_{uv}=t_u-t_v,\qquad
    \Delta^b_{uv}=b_u-b_v.
\end{equation}
These offsets are embedded by learnable lookup tables
$\boldsymbol{\phi}_{\ell}$, $\boldsymbol{\phi}_{t}$, and $\boldsymbol{\phi}_{b}$. 
We also compute two content-dependent terms from the initial node representations:
\begin{equation}
    c_{uv}
    =
    \cos(\mathbf{x}_u,\mathbf{x}_v),
    \qquad
    d_{uv}
    =
    \|\mathbf{x}_u-\mathbf{x}_v\|_2.
\end{equation}
The edge feature is then
\begin{equation}
    \mathbf{e}_{uv}
    =
    \mathrm{MLP}_{1}
    \left(
    [
    \boldsymbol{\phi}_{\ell}(\Delta^\ell_{uv}),
    \boldsymbol{\phi}_{t}(\Delta^t_{uv}),
    \boldsymbol{\phi}_{b}(\Delta^b_{uv}),
    c_{uv},
    d_{uv}
    ]
    \right)
    \in
    \mathbb{R}^{d_{\mathrm{m}}}.
\end{equation}
Thus, each edge is both position-aware and content-aware: the relative offsets encode where two nodes are located in the layer-token-feature grid, while the similarity and distance terms encode how their hidden-state blocks relate for the current input.

\paragraph{Output of graph construction.}
The graph construction step produces
\begin{equation}
    \mathcal{G}_{i}(\mathbf{Q},\mathbf{A})
    =
    \left(
    \{\mathbf{x}_u\}_{u=1}^{N},
    \{\mathbf{e}_{uv}\}_{u,v=1}^{N}
    \right).
\end{equation}
This graph is then passed to the relation-aware attention encoder described in Appendix~\ref{app:relation_aware_attention}. 
The construction above is shared across all source and target LLMs; only the extracted hidden states and the deterministic padding map $p_i$ depend on the specific LLM.

\subsection{Details of Relation-Aware Attention}
\label{app:relation_aware_attention}

This subsection expands the relation-aware attention module used in Section~\ref{sec:method}. 
The module follows the general formulation of relation-aware self-attention in \citet{shaw2018self}, which augments self-attention with pairwise edge representations. 
While the original formulation uses relative positions in a sequence as edge labels, \method{} applies the same principle to the fully connected layer-token-feature graph constructed in Section~\ref{sec:method}. 
That is, every node corresponds to a triple $(\ell_u,t_u,b_u)$, and every directed edge $(u,v)$ carries an input-dependent relation feature $\mathbf{e}_{uv}$ defined from relative layer, token, and block offsets, together with content similarity and distance.

\paragraph{Fully connected relation graph.}
Although we use relative offsets to parameterize edge types, the graph itself is not sparsified. 
For every ordered node pair $(u,v)$ with $u,v\in\{1,\ldots,N\}$, we construct an edge feature $\mathbf{e}_{uv}$ and allow node $u$ to attend to node $v$. 
Therefore, the relation-aware encoder operates on a directed fully connected graph. 
The relative offsets determine how each edge is represented, not whether the edge exists. 
This is consistent with the relation-aware view of \citet{shaw2018self}, where self-attention can be interpreted as operating over a labeled, directed, fully connected graph.

\paragraph{Edge feature construction.}
Recall that each node $u$ has an initial representation $\mathbf{x}_u\in\mathbb{R}^{d_{\mathrm{m}}}$, and each node is associated with a layer index $\ell_u$, token index $t_u$, and feature-block index $b_u$. 
For a directed pair $(u,v)$, we define clipped relative offsets
\begin{equation}
    \Delta^{\ell}_{uv}
    =
    \mathrm{clip}(\ell_u-\ell_v,c_{\ell}),\quad
    \Delta^{t}_{uv}
    =
    \mathrm{clip}(t_u-t_v,c_{t}),\quad
    \Delta^{b}_{uv}
    =
    \mathrm{clip}(b_u-b_v,c_{b}),
\end{equation}
where $\mathrm{clip}(a,c)=\max(-c,\min(c,a))$. 
Following the motivation of relative-position attention, clipping ties together sufficiently distant offsets so that they share the same relation label. 
Importantly, clipping does not remove any edge: all directed pairs $(u,v)$ still participate in the attention softmax. 
In our implementation, we set
\begin{equation}
    c_{\ell}=|\mathbf{L}^{\mathrm{sel}}|-1,\qquad
    c_t=|\mathbf{T}^{\mathrm{sel}}|-1,\qquad
    c_b=B-1.
\end{equation}
Thus, for the compact graphs used by the final detector, all relative offsets inside the selected layer-token-block grid are represented exactly. 
This choice avoids introducing an additional sparsification or truncation rule, while still keeping the number of relation labels bounded by the selected graph size.

We embed the three relative offsets with learnable lookup tables,
\begin{equation}
    \boldsymbol{\phi}_{\ell}(\Delta^{\ell}_{uv}),\quad
    \boldsymbol{\phi}_{t}(\Delta^{t}_{uv}),\quad
    \boldsymbol{\phi}_{b}(\Delta^{b}_{uv})
    \in \mathbb{R}^{d_{\mathrm{r}}},
\end{equation}
and combine them with two content-dependent scalars,
\begin{equation}
    c_{uv} = \cos(\mathbf{x}_u,\mathbf{x}_v),\qquad
    d_{uv} = \|\mathbf{x}_u-\mathbf{x}_v\|_2 .
\end{equation}
The final edge feature is
\begin{equation}
    \mathbf{e}_{uv}
    =
    \mathrm{MLP}_{1}
    \Big(
    [
    \boldsymbol{\phi}_{\ell}(\Delta^{\ell}_{uv}),
    \boldsymbol{\phi}_{t}(\Delta^{t}_{uv}),
    \boldsymbol{\phi}_{b}(\Delta^{b}_{uv}),
    c_{uv},
    d_{uv}
    ]
    \Big)
    \in \mathbb{R}^{d_{\mathrm{m}}}.
\end{equation}
Thus, unlike a purely relative-position Transformer, our edge feature is both \emph{position-aware} and \emph{content-aware}. 
This is important because hallucination-related evidence may depend not only on where two nodes occur in the layer-token-feature grid, but also on how their hidden-state blocks differ for a particular input.

\paragraph{Relation-aware multi-head attention.}
Let $\mathbf{x}^{(q)}_u$ be the representation of node $u$ before the $q$-th relation-aware attention layer, with $\mathbf{x}^{(0)}_u=\mathbf{x}_u$. 
For each attention head $h\in\{1,\ldots,H\}$, we compute
\begin{equation}
    \mathbf{q}^{(q,h)}_u
    =
    \mathbf{W}^{(q,h)}_{Q}\mathbf{x}^{(q)}_u,\quad
    \mathbf{k}^{(q,h)}_v
    =
    \mathbf{W}^{(q,h)}_{K}\mathbf{x}^{(q)}_v,\quad
    \mathbf{v}^{(q,h)}_v
    =
    \mathbf{W}^{(q,h)}_{V}\mathbf{x}^{(q)}_v,
\end{equation}
where $\mathbf{q}^{(q,h)}_u,\mathbf{k}^{(q,h)}_v,\mathbf{v}^{(q,h)}_v\in\mathbb{R}^{d_h}$ and $d_h=d_{\mathrm{m}}/H$. 
Following \citet{shaw2018self}, we use two edge projections: one for attention compatibility and one for value propagation,
\begin{equation}
    \mathbf{a}^{K,(q,h)}_{uv}
    =
    \mathbf{W}^{K,(q,h)}_{E}\mathbf{e}_{uv},
    \qquad
    \mathbf{a}^{V,(q,h)}_{uv}
    =
    \mathbf{W}^{V,(q,h)}_{E}\mathbf{e}_{uv},
\end{equation}
where $\mathbf{a}^{K,(q,h)}_{uv},\mathbf{a}^{V,(q,h)}_{uv}\in\mathbb{R}^{d_h}$. 
The compatibility score from node $u$ to node $v$ is then
\begin{equation}
    s^{(q,h)}_{uv}
    =
    \frac{
    \big(\mathbf{q}^{(q,h)}_u\big)^{\top}
    \big(
    \mathbf{k}^{(q,h)}_v
    +
    \mathbf{a}^{K,(q,h)}_{uv}
    \big)
    }{\sqrt{d_h}},
\end{equation}
and the attention weight is
\begin{equation}
    \alpha^{(q,h)}_{uv}
    =
    \frac{\exp(s^{(q,h)}_{uv})}
    {\sum_{v'=1}^{N}\exp(s^{(q,h)}_{uv'})}.
\end{equation}
Since the graph is fully connected, the denominator ranges over all nodes $v'\in\{1,\ldots,N\}$. 
The message received by node $u$ in head $h$ is
\begin{equation}
    \mathbf{m}^{(q,h)}_u
    =
    \sum_{v=1}^{N}
    \alpha^{(q,h)}_{uv}
    \Big(
    \mathbf{v}^{(q,h)}_v
    +
    \mathbf{a}^{V,(q,h)}_{uv}
    \Big).
\end{equation}
The head outputs are concatenated and projected:
\begin{equation}
    \mathbf{m}^{(q)}_u
    =
    \mathbf{W}^{(q)}_{O}
    \big[
    \mathbf{m}^{(q,1)}_u;
    \cdots;
    \mathbf{m}^{(q,H)}_u
    \big].
\end{equation}
Finally, we use the standard Transformer-style residual, layer normalization, and feed-forward update:
\begin{align}
    \tilde{\mathbf{x}}^{(q+1)}_u
    &=
    \mathrm{LN}
    \Big(
    \mathbf{x}^{(q)}_u
    +
    \mathrm{Dropout}(\mathbf{m}^{(q)}_u)
    \Big),\\
    \mathbf{x}^{(q+1)}_u
    &=
    \mathrm{LN}
    \Big(
    \tilde{\mathbf{x}}^{(q+1)}_u
    +
    \mathrm{Dropout}
    \big(
    \mathrm{FFN}(\tilde{\mathbf{x}}^{(q+1)}_u)
    \big)
    \Big).
\end{align}
After $Q$ relation-aware attention layers, we set $\mathbf{x}'_u=\mathbf{x}^{(Q)}_u$ and apply the attention pooling operator in Section~\ref{sec:method} to obtain the graph-level feature $\mathbf{z}$.

\paragraph{Connection to the original relation-aware attention.}
The above equations are a direct graph-level adaptation of the relation-aware self-attention mechanism of \citet{shaw2018self}. 
In their sequence setting, the edge representation is determined by the relative position $j-i$ between two tokens. 
In our setting, the edge representation is determined by the relative layer-token-block offsets and the input-dependent content relation between two graph nodes. 
Thus, our module replaces a one-dimensional sequential relation with a three-dimensional layer-token-feature relation, while preserving the same key idea: edge information modifies both the attention score and the propagated value.

\paragraph{Implementation details.}
Unless otherwise stated, we use $Q=2$ relation-aware attention layers, $H=4$ attention heads, $d_{\mathrm{m}}=256$, and $d_h=64$. 
The relative-offset embedding size is $d_{\mathrm{r}}=64$. 
The edge MLP $\mathrm{MLP}_{1}$ is a two-layer network with hidden dimension $256$, GELU activation, dropout $0.1$, and output dimension $d_{\mathrm{m}}$. 
The feed-forward network inside each relation-aware attention layer uses hidden dimension $4d_{\mathrm{m}}$ with GELU activation and dropout $0.1$. 
We include self-loops in the attention softmax, following standard self-attention; for $u=v$, the relative offsets are zero, $c_{uv}=1$, and $d_{uv}=0$. 
All edge features $\mathbf{e}_{uv}$ are computed once from the initial node representations and reused across relation-aware attention layers. 
This keeps the edge semantics tied to the original layer-token-feature graph while allowing node states to be updated through multi-hop relational propagation.

For the final detector, we use a graph with $|\mathbf{T}^{\mathrm{sel}}|=3$, $|\mathbf{L}^{\mathrm{sel}}|=3$, and $B=64$, giving $N=576$ graph nodes. 
The graph is fully connected, so the relation-aware encoder processes $N^2$ directed node pairs per example. 
Because the selected token and layer windows are small, this cost remains tractable in practice, while still allowing the model to capture interactions across layers, tokens, and feature blocks.
The sensitivity analysis in Appendix~\ref{app:obs_token_layer_window} further shows that increasing the number of selected tokens or layers yields only marginal additional gains.

\subsection{Details of Meta-Learning Training}
\label{app:meta_learning_details}

This subsection expands the meta-learning training strategy used in Section~\ref{sec:method}. 
Our implementation is adapted from Feature-Critic \citep{li2019feature}, which learns an auxiliary loss by simulating domain shift among source domains. 
In the original formulation, source domains are randomly split into virtual training and validation domains, and the auxiliary loss is optimized so that an update using this auxiliary loss improves validation-domain performance. 
In \method{}, we instantiate this idea for cross-model hallucination detection by treating each source LLM as one domain. 
Thus, a held-out source LLM in each meta-episode serves as a pseudo-unseen target model.

\paragraph{Detector decomposition.}
We write the detector as
\begin{equation}
    \hat{y}
    =
    D_{\Theta}(\mathbf{Q},\mathbf{A};\mathbb{P}_{\boldsymbol{\theta}_i})
    =
    C_{\phi}
    \big(
    G_{\psi}(\mathbf{Q},\mathbf{A};\mathbb{P}_{\boldsymbol{\theta}_i})
    \big),
\end{equation}
where $G_{\psi}$ denotes the graph encoder, including hidden-state restructuring, graph construction, relation-aware attention, and attention pooling. 
It outputs the graph-level feature
\begin{equation}
    \mathbf{z}
    =
    G_{\psi}(\mathbf{Q},\mathbf{A};\mathbb{P}_{\boldsymbol{\theta}_i})
    \in\mathbb{R}^{d_{\mathrm{m}}},
\end{equation}
and $C_{\phi}$ is the binary hallucination classifier. 
The full learnable parameter set is $\Theta=(\psi,\phi)$. 
Unlike the heterogeneous classification setting in \citet{li2019feature}, all source LLMs in our problem share the same binary label space, so we use a single shared classifier $C_{\phi}$ rather than domain-specific classifiers.

\paragraph{Meta-train and meta-validation split.}
Let $\mathcal{M}_{\mathrm{src}}$ denote the set of source LLMs used during training, and let $\mathcal{D}_i$ be the labeled hallucination-detection data associated with source LLM $\mathbb{P}_{\boldsymbol{\theta}_i}$. 
At each meta-training iteration, we randomly sample one source LLM as the pseudo-target model,
\begin{equation}
    i_{\mathrm{val}} \sim \mathrm{Unif}(\mathcal{M}_{\mathrm{src}}),
\end{equation}
and define
\begin{equation}
    \mathcal{M}_{\mathrm{val}}=\{i_{\mathrm{val}}\},
    \qquad
    \mathcal{M}_{\mathrm{trn}}
    =
    \mathcal{M}_{\mathrm{src}}
    \setminus
    \mathcal{M}_{\mathrm{val}}.
\end{equation}
Mini-batches are sampled from all LLMs in $\mathcal{M}_{\mathrm{trn}}$ and from the held-out LLM in $\mathcal{M}_{\mathrm{val}}$. 
This leave-one-source-model-out split matches the test condition of CMHD: the detector must perform well on an LLM whose hidden-state distribution was not used for the current update.

\paragraph{Supervised loss.}
For a mini-batch $\mathcal{B}$, the binary cross-entropy loss is
\begin{equation}
    \mathcal{L}_{\mathrm{bce}}(\mathcal{B};\psi,\phi)
    =
    \frac{1}{|\mathcal{B}|}
    \sum_{(\mathbf{Q},\mathbf{A},y,i)\in\mathcal{B}}
    \ell^{\mathrm{bce}}
    \left(
    C_{\phi}
    \big(
    G_{\psi}(\mathbf{Q},\mathbf{A};\mathbb{P}_{\boldsymbol{\theta}_i})
    \big),
    y
    \right).
\end{equation}
In practice, mini-batches are class-balanced within each source LLM whenever possible. 
When exact balancing is not possible, we use a positive-class weight in the binary cross-entropy computed from the training split.

\paragraph{Feature-Critic auxiliary loss.}
Following \citet{li2019feature}, we introduce a learned non-negative auxiliary loss, denoted by $h_{\omega}$. 
The critic is applied to the graph-level features produced by $G_{\psi}$, rather than directly to the detector parameters. 
For a mini-batch $\mathcal{B}_i$ from source LLM $i$, let
\begin{equation}
    \mathbf{Z}_i
    =
    \left[
    \mathbf{z}_{i,1};
    \ldots;
    \mathbf{z}_{i,|\mathcal{B}_i|}
    \right]
    \in
    \mathbb{R}^{|\mathcal{B}_i|\times d_{\mathrm{m}}},
\end{equation}
where $\mathbf{z}_{i,r}=G_{\psi}(\mathbf{Q}_{i,r},\mathbf{A}_{i,r};\mathbb{P}_{\boldsymbol{\theta}_i})$. 
We use a permutation-invariant set-embedding critic,
\begin{equation}
    h_{\omega}(\mathbf{Z}_i)
    =
    \mathrm{Softplus}
    \left(
    \mathbf{w}_{\omega}^{\top}
    \frac{1}{|\mathcal{B}_i|}
    \sum_{r=1}^{|\mathcal{B}_i|}
    \rho_{\omega}
    \big(
    \mathrm{LN}(\mathbf{z}_{i,r})
    \big)
    +
    b_{\omega}
    \right),
\end{equation}
where $\rho_{\omega}$ is a small MLP, $\mathrm{LN}$ denotes layer normalization, and the final Softplus ensures that the critic output is non-negative. 
The auxiliary loss over the meta-training LLMs is
\begin{equation}
    \mathcal{L}_{\mathrm{aux}}
    (\mathcal{B}_{\mathrm{trn}};\psi,\omega)
    =
    \frac{1}{|\mathcal{M}_{\mathrm{trn}}|}
    \sum_{i\in\mathcal{M}_{\mathrm{trn}}}
    h_{\omega}(\mathbf{Z}_i).
\end{equation}
This design follows the set-embedding variant of Feature-Critic, which is more parameter-efficient than covariance-based critics for high-dimensional features.

\paragraph{Meta-objective.}
The critic should be useful only if an update using it improves performance on the pseudo-target LLM. 
To measure this, we compare two one-step virtual updates of the graph encoder parameters. 
The first update uses only the supervised loss:
\begin{equation}
    \psi^{\mathrm{old}}
    =
    \psi
    -
    \alpha
    \nabla_{\psi}
    \mathcal{L}_{\mathrm{bce}}
    (\mathcal{B}_{\mathrm{trn}};\psi,\phi).
\end{equation}
The second update uses both the supervised loss and the learned auxiliary loss:
\begin{equation}
    \psi^{\mathrm{new}}
    =
    \psi
    -
    \alpha
    \nabla_{\psi}
    \left[
    \mathcal{L}_{\mathrm{bce}}
    (\mathcal{B}_{\mathrm{trn}};\psi,\phi)
    +
    \lambda_{\mathrm{fc}}
    \mathcal{L}_{\mathrm{aux}}
    (\mathcal{B}_{\mathrm{trn}};\psi,\omega)
    \right],
\end{equation}
where $\alpha$ is the virtual inner-step size and $\lambda_{\mathrm{fc}}$ controls the strength of the auxiliary loss. 
The classifier parameters $\phi$ are kept fixed in these virtual updates, because the critic is intended to improve the transferable graph representation rather than to overfit the binary decision head.

Given a validation mini-batch $\mathcal{B}_{\mathrm{val}}$ from the held-out source LLM, we compute
\begin{align}
    \mathcal{L}_{\mathrm{val}}^{\mathrm{old}}
    &=
    \mathcal{L}_{\mathrm{bce}}
    (\mathcal{B}_{\mathrm{val}};\psi^{\mathrm{old}},\phi),\\
    \mathcal{L}_{\mathrm{val}}^{\mathrm{new}}
    &=
    \mathcal{L}_{\mathrm{bce}}
    (\mathcal{B}_{\mathrm{val}};\psi^{\mathrm{new}},\phi).
\end{align}
The meta-loss for updating the critic is
\begin{equation}
    \mathcal{L}_{\mathrm{meta}}(\omega)
    =
    \tanh
    \left(
    \mathcal{L}_{\mathrm{val}}^{\mathrm{new}}
    -
    \mathcal{L}_{\mathrm{val}}^{\mathrm{old}}
    \right).
\end{equation}
Minimizing this loss encourages the critic to produce auxiliary gradients that make $\psi^{\mathrm{new}}$ outperform $\psi^{\mathrm{old}}$ on the pseudo-target LLM. 
The hyperbolic tangent follows the utility transformation used in Feature-Critic and prevents very large validation-loss differences from producing unstable critic updates.

\paragraph{Outer optimization.}
After computing the meta-loss, we update the critic parameters by
\begin{equation}
    \omega
    \leftarrow
    \omega
    -
    \eta_{\omega}
    \nabla_{\omega}
    \mathcal{L}_{\mathrm{meta}}(\omega),
\end{equation}
where gradients are backpropagated through the differentiable virtual update that produces $\psi^{\mathrm{new}}$. 
We then update the actual detector parameters on the meta-training mini-batches:
\begin{align}
    \psi
    &\leftarrow
    \psi
    -
    \eta_{\psi}
    \nabla_{\psi}
    \left[
    \mathcal{L}_{\mathrm{bce}}
    (\mathcal{B}_{\mathrm{trn}};\psi,\phi)
    +
    \lambda_{\mathrm{fc}}
    \mathcal{L}_{\mathrm{aux}}
    (\mathcal{B}_{\mathrm{trn}};\psi,\omega)
    \right],\\
    \phi
    &\leftarrow
    \phi
    -
    \eta_{\phi}
    \nabla_{\phi}
    \mathcal{L}_{\mathrm{bce}}
    (\mathcal{B}_{\mathrm{trn}};\psi,\phi).
\end{align}
When updating $(\psi,\phi)$, the critic parameters $\omega$ are treated as fixed. 
When updating $\omega$, the detector parameters are not updated by the meta-loss. 
This separation keeps the critic as a learned training signal rather than an additional prediction module at test time. 
At inference, the critic is discarded and only $D_{\Theta}$ is used.

\paragraph{All-source consolidation.}
After meta-training, we run a short consolidation stage over all source LLMs without any pseudo-target split. 
The critic is frozen, and the detector is trained on the union of all source-model data using
\begin{equation}
    \mathcal{L}_{\mathrm{final}}
    =
    \mathcal{L}_{\mathrm{bce}}
    +
    \lambda_{\mathrm{fc}}
    \mathcal{L}_{\mathrm{aux}}.
\end{equation}
This stage uses no target-model data. 
It simply lets the detector absorb all labeled source-model information after the critic has learned a generalization-promoting auxiliary objective.

\paragraph{Implementation details.}
Unless otherwise stated, we use one virtual inner update per meta-iteration. 
We set the meta-validation split to one held-out source LLM, i.e., $|\mathcal{M}_{\mathrm{val}}|=1$, and use all remaining source LLMs as meta-training domains. 
The per-source mini-batch size is $32$ for both meta-training and meta-validation. 
We use $\lambda_{\mathrm{fc}}=0.1$ and virtual step size $\alpha=5\times10^{-4}$. 
The critic MLP $\rho_{\omega}$ has two layers with hidden dimension $256$, GELU activation, dropout $0.1$, and output dimension $128$. 
The detector parameters $(\psi,\phi)$ are optimized with AdamW using learning rate $2\times10^{-4}$ and weight decay $10^{-4}$, while the critic uses AdamW with learning rate $1\times10^{-4}$ and the same weight decay. 
We clip gradients to norm $1.0$. 
For stability, the first $5\%$ of training updates use only the supervised binary cross-entropy loss; the Feature-Critic auxiliary loss and meta-loss are enabled afterward. 
The final all-source consolidation stage uses the last $10\%$ of the total training budget.

\subsection{Details for the Cross-Model Alignment, Transfer, and Score-Distribution Diagnostics}
\label{app:obs_alignment_transfer_details}
This subsection provides the experimental details for Figures~\ref{fig:obs_alignment_detection} and~\ref{fig:obs_kde_scores}. 
We use a mixed benchmark pool constructed from TruthfulQA~\citep{lin2022truthfulqa}, HaluEval~\citep{li-etal-2023-halueval}, TriviaQA~\citep{joshi-etal-2017-triviaqa}, CoQA~\citep{reddy-etal-2019-coqa}, GSM8K~\citep{cobbe2021training}, MBPP~\citep{austin2021program}, and xLAM~\citep{liu2024apigen}. 
For each benchmark, we sample up to 1,000 examples; for MBPP, we use all available examples because the usable set contains fewer than 1,000 examples in our setting. 
The same question pool is used across models whenever possible, so that both neighborhood alignment and score-distribution comparisons are computed over matched evaluation examples.

We evaluate five representative directed source--target model pairs: Llama3.2-3B$\rightarrow$Llama3.1-8B, Qwen3-14B$\rightarrow$Qwen3-8B, Llama3.1-8B$\rightarrow$Qwen3-8B, Llama3.3-70B$\rightarrow$Qwen3.5-35B, and Llama3.2-3B$\rightarrow$Ministral3-8B. 
These pairs cover same-family transfer, cross-family transfer, scale-matched transfer, large-model transfer, and scale-mismatched transfer. 
Because hidden dimensions differ across models, we first map all extracted hidden states to a shared padded dimension $d_{\max}=8192$ before constructing raw or relational features. 
For a hidden state $\mathbf{h}_{\boldsymbol{\theta}_i,t}^{\ell}\in\mathbb{R}^{d_i}$, the common-dimensional vector is obtained by the same deterministic index mapping used in the main method: each original coordinate is placed into a fixed position in the $d_{\max}$-dimensional vector, and all remaining positions are filled with zeros. 
This step keeps feature dimensions consistent across LLMs without learning an alignment from target labels.

For each example and each model, we extract hidden states from the selected generated-token and layer windows, apply the common-dimensional mapping, and then construct the raw-state, token-level, cross-layer token, within-token feature, and joint relational features described in Section~\ref{sec:observations}. 
Raw-state features are obtained by pooling the mapped hidden states over the same selected token--layer window. 
Relational features are constructed after the same mapping, so that token, layer, feature-block, and joint relations are computed in a common feature space.

For local-neighborhood alignment in Figure~\ref{fig:obs_alignment_detection}, we compute mutual $k$NN overlap with $k=10$. 
Given two models in a pair, we build the $k$ nearest-neighbor set of each example under the feature space induced by each model, using inner-product similarity after feature normalization. 
The reported alignment is the average normalized overlap between the two neighbor sets over all examples. 
This gives a model-pair-level measure of whether two LLMs induce similar local example geometry under the same feature construction.

For transfer AUROC in Figure~\ref{fig:obs_alignment_detection}, we train a detector on the source-model features and evaluate it directly on the target-model features. 
The detector is a lightweight two-layer MLP with one hidden layer, ReLU activation, dropout, and a final scalar logit trained with binary cross-entropy. 
We use the same MLP architecture and optimization protocol for all feature types, so differences in transfer AUROC reflect the feature construction rather than a change in classifier capacity. 
Training uses only labels from the source model, while target-model labels are used only for evaluation. 
All reported bar values are averaged over multiple random seeds, and error bars show the corresponding seed-level variation.

Figure~\ref{fig:obs_kde_scores} further visualizes the detector score distributions for one representative transfer pair, Llama3.1-8B$\rightarrow$Qwen3-8B. 
For each feature construction, we use the same source-trained MLP detector described above and record its scalar detector scores on the target-model evaluation examples. 
We then estimate separate kernel density curves for factual and hallucinated examples. 
The solid curve denotes factual examples and the dashed curve denotes hallucinated examples. 
The AUROC annotated in each panel is recomputed from the finite sample of target scores plotted in that panel, and may therefore differ slightly from the corresponding entry of the full transfer matrix. 
Thus, Figure~\ref{fig:obs_kde_scores} provides a distribution-level view of the same transfer behavior summarized by AUROC in Figure~\ref{fig:obs_alignment_detection}.

In addition to raw and relational features, Figure~\ref{fig:obs_kde_scores} includes two training-free reference baselines, Perplexity and Semantic Entropy. 
These methods are evaluated directly on the same target examples and do not use the source-model training labels. 
They are included only as score-distribution references: Perplexity represents a simple model-confidence baseline, while Semantic Entropy represents a stronger uncertainty-based hallucination detector. 
Comparing their distributions with the relational-feature detectors helps contextualize how much separability is obtained by simple relational transfer before introducing the full \method{} training framework.

\subsection{Observation Pipeline for Relational Features}
\label{app:obs_pipeline}

This subsection provides the full diagnostic pipeline used in Section~\ref{sec:joint_relations}. 
The goal of this observation study is not to introduce the final learnable detector, but to test whether simple handcrafted relational structures already contain transferable hallucination signals across LLMs. 
Therefore, all edge weights in this subsection are fixed by non-learnable functions, and the only supervised component used for predictive transfer is a lightweight MLP detector trained after the relational features are extracted.

\paragraph{Shared preprocessing and notation.}
For each LLM $\mathbb{P}_{\boldsymbol{\theta}_i}$, we extract hidden states
$\mathbf{h}_{\boldsymbol{\theta}_i,t}^{\ell}(\mathbf{Q},\mathbf{A})\in\mathbb{R}^{d_i}$
from selected answer-token positions and selected layers. 
The selected token set is denoted by
$\mathbf{T}^{\mathrm{sel}}=\{t_1,\ldots,t_r\}$, and the selected layer set is denoted by
$\mathbf{L}^{\mathrm{sel}}_i=\{\ell_1,\ldots,\ell_m\}$, where $r=|\mathbf{T}^{\mathrm{sel}}|$ and $m=|\mathbf{L}^{\mathrm{sel}}_i|$. 
The number and positions of selected tokens and layers are specified in Appendix~\ref{app:obs_token_layer_window}. 
Since different LLMs may have different hidden dimensions, we first map all hidden states into the common padded space $\mathbb{R}^{d^\star}$ using the same hidden-state restructuring operation as in Section~\ref{sec:method}. 
For notational simplicity, we denote the mapped hidden state by
\begin{equation}
    \bar{\mathbf{h}}_{\boldsymbol{\theta}_i,t}^{\ell}
    =
    \mathrm{Map}_{i}
    \left(
    \mathbf{h}_{\boldsymbol{\theta}_i,t}^{\ell}
    \right)
    \in\mathbb{R}^{d^\star}.
\end{equation}
For feature-block and joint relations, the mapped dimension is partitioned into $B$ equal-size contiguous blocks
$\{\mathcal{I}_b\}_{b=1}^{B}$ as in Eq.~\eqref{eq:block_partition}. 
The choice of $B$ is discussed in Appendix~\ref{app:feature_block_granularity}. 
We choose $d^\star$ to be divisible by $B$.

\paragraph{Generic relation-weighted graph readout.}
All relational features follow the same graph construction and pooling template. 
Given a relation type $r\in\{\mathrm{tok},\mathrm{lyr},\mathrm{feat},\mathrm{joint}\}$, we construct a graph
\begin{equation}
    \mathcal{G}_{i}^{r}(\mathbf{Q},\mathbf{A})
    =
    \left(
    \mathcal{V}_{i}^{r},
    \mathcal{E}_{i}^{r}
    \right),
\end{equation}
where each node $u\in\mathcal{V}_{i}^{r}$ has a node feature $\mathbf{x}_u^{r}$, and each directed edge $(u,v)$ has a non-negative relation weight
$R_i^{r}(u,v;\mathbf{Q},\mathbf{A})$. 
For the diagnostic graph readout, we use a fully connected graph without self-loops. 
The edge weights are row-normalized as
\begin{equation}
    \alpha_i^{r}(u,v)
    =
    \frac{
    R_i^{r}(u,v;\mathbf{Q},\mathbf{A})
    }{
    \sum_{v'\in\mathcal{V}_{i}^{r},\,v'\neq u}
    R_i^{r}(u,v';\mathbf{Q},\mathbf{A})
    +
    \varepsilon
    },
    \qquad v\neq u,
\end{equation}
where $\varepsilon=10^{-8}$ is used for numerical stability. 
Each node then receives one relation-weighted aggregation step,
\begin{equation}
    \mathbf{m}_u^{r}
    =
    \sum_{v\in\mathcal{V}_{i}^{r},\,v\neq u}
    \alpha_i^{r}(u,v)\mathbf{x}_v^{r}.
\end{equation}
Finally, we obtain a graph-level feature by mean pooling:
\begin{equation}
    \mathbf{z}_{i}^{r}(\mathbf{Q},\mathbf{A})
    =
    \frac{1}{|\mathcal{V}_{i}^{r}|}
    \sum_{u\in\mathcal{V}_{i}^{r}}
    \mathbf{m}_u^{r}.
\end{equation}
This pooling operation is invariant to the enumeration order of the nodes, while the edge weights still encode the intended token, layer, and block relations. 
The same readout template is used for all relational features below; they differ only in the definition of nodes and edge weights.

\paragraph{Token-level relation.}
For token-level relations, the graph nodes are the selected tokens in $\mathbf{T}^{\mathrm{sel}}$ at a diagnostic layer $\ell^\star$. 
We choose $\ell^\star$ by the same relative-depth rule used in Appendix~\ref{app:obs_token_layer_window}. 
Each token node has feature
\begin{equation}
    \mathbf{x}_{t}^{\mathrm{tok}}
    =
    \bar{\mathbf{h}}_{\boldsymbol{\theta}_i,t}^{\ell^\star}.
\end{equation}
For two selected tokens $t$ and $\tilde{t}$, the edge weight is
\begin{equation}
    R_i^{\mathrm{tok}}(t,\tilde{t};\mathbf{Q},\mathbf{A})
    =
    s^{\mathrm{tok}}(t,\tilde{t})
    \left\|
    \bar{\mathbf{h}}_{\boldsymbol{\theta}_i,t}^{\ell^\star}
    -
    \bar{\mathbf{h}}_{\boldsymbol{\theta}_i,\tilde{t}}^{\ell^\star}
    \right\|_2,
\end{equation}
where
\begin{equation}
    s^{\mathrm{tok}}(t,\tilde{t})
    =
    \exp
    \left(
    -
    \frac{|t-\tilde{t}|}{\tau_{\mathrm{tok}}}
    \right).
\end{equation}
We use $\tau_{\mathrm{tok}}=2.0$ in selected-token index units and keep it fixed across all experiments.

\paragraph{Cross-layer token relation.}
For cross-layer token relations, each node corresponds to one selected layer $\ell\in\mathbf{L}^{\mathrm{sel}}_i$. 
The node feature is the vectorized stack of the selected token representations at that layer:
\begin{equation}
    \mathbf{x}_{\ell}^{\mathrm{lyr}}
    =
    \mathrm{vec}
    \left(
    \left[
    \bar{\mathbf{h}}_{\boldsymbol{\theta}_i,t_1}^{\ell},
    \ldots,
    \bar{\mathbf{h}}_{\boldsymbol{\theta}_i,t_r}^{\ell}
    \right]^{\top}
    \right)
    \in\mathbb{R}^{r d^\star}.
\end{equation}
For two selected layers $\ell$ and $\tilde{\ell}$, the relation weight is
\begin{equation}
    R_i^{\mathrm{lyr}}(\ell,\tilde{\ell};\mathbf{Q},\mathbf{A})
    =
    s^{\mathrm{lyr}}(\ell,\tilde{\ell})
    \left(
    1-
    \cos
    \left(
    \mathbf{x}_{\ell}^{\mathrm{lyr}},
    \mathbf{x}_{\tilde{\ell}}^{\mathrm{lyr}}
    \right)
    \right),
\end{equation}
where
\begin{equation}
    s^{\mathrm{lyr}}(\ell,\tilde{\ell})
    =
    \exp
    \left(
    -
    \frac{|\ell-\tilde{\ell}|}{\tau_{\mathrm{lyr}}}
    \right).
\end{equation}
We use $\tau_{\mathrm{lyr}}=2.0$ in selected-layer index units and keep it fixed. 
The cosine-distance term captures representational change across depth, while the decay factor gives slightly larger weight to nearby layers.

\paragraph{Within-token feature relation.}
For within-token feature relations, each node corresponds to one feature block $b\in\{1,\ldots,B\}$. 
For every block, we collect its hidden-state block over all selected layers and selected tokens:
\begin{equation}
    \mathbf{V}_{\boldsymbol{\theta}_i,b}(\mathbf{Q},\mathbf{A})
    =
    \left[
    \bar{\mathbf{h}}_{\boldsymbol{\theta}_i,t_1}^{\ell_1}[\mathcal{I}_b],
    \ldots,
    \bar{\mathbf{h}}_{\boldsymbol{\theta}_i,t_r}^{\ell_1}[\mathcal{I}_b],
    \ldots,
    \bar{\mathbf{h}}_{\boldsymbol{\theta}_i,t_1}^{\ell_m}[\mathcal{I}_b],
    \ldots,
    \bar{\mathbf{h}}_{\boldsymbol{\theta}_i,t_r}^{\ell_m}[\mathcal{I}_b]
    \right]^{\top}
    \in
    \mathbb{R}^{mr\times d^\star/B}.
\end{equation}
The block-node feature is
\begin{equation}
    \mathbf{x}_{b}^{\mathrm{feat}}
    =
    \mathrm{vec}
    \left(
    \mathbf{V}_{\boldsymbol{\theta}_i,b}(\mathbf{Q},\mathbf{A})
    \right)
    \in
    \mathbb{R}^{mr d^\star/B}.
\end{equation}
For two blocks $b$ and $\tilde{b}$, the relation weight is
\begin{equation}
    R_i^{\mathrm{feat}}(b,\tilde{b};\mathbf{Q},\mathbf{A})
    =
    \left\|
    \mathbf{x}_{b}^{\mathrm{feat}}
    -
    \mathbf{x}_{\tilde{b}}^{\mathrm{feat}}
    \right\|_2.
\end{equation}
This relation measures how differently two feature blocks respond over the same selected token-layer window.

\paragraph{Joint layer-token-feature relation.}
For joint relations, each node corresponds to a triple
$u=(\ell,t,b)$ with $\ell\in\mathbf{L}^{\mathrm{sel}}_i$, $t\in\mathbf{T}^{\mathrm{sel}}$, and $b\in\{1,\ldots,B\}$. 
The node feature is the corresponding hidden-state block:
\begin{equation}
    \mathbf{x}_{u}^{\mathrm{joint}}
    =
    \bar{\mathbf{h}}_{\boldsymbol{\theta}_i,t}^{\ell}[\mathcal{I}_b]
    \in
    \mathbb{R}^{d^\star/B}.
\end{equation}
For two nodes $u=(\ell,t,b)$ and $v=(\tilde{\ell},\tilde{t},\tilde{b})$, the relation weight is
\begin{equation}
    R_i^{\mathrm{joint}}(u,v;\mathbf{Q},\mathbf{A})
    =
    s^{\mathrm{lyr}}(\ell,\tilde{\ell})
    s^{\mathrm{tok}}(t,\tilde{t})
    \left\|
    \bar{\mathbf{h}}_{\boldsymbol{\theta}_i,t}^{\ell}[\mathcal{I}_b]
    -
    \bar{\mathbf{h}}_{\boldsymbol{\theta}_i,\tilde{t}}^{\tilde{\ell}}[\mathcal{I}_{\tilde{b}}]
    \right\|_2 .
\end{equation}
This joint graph combines the three axes considered separately above: depth, token position, and feature-block structure. 
It is deliberately handcrafted in the observation study, so that any cross-model transfer observed here can be attributed to relational structure rather than to a learned relation module.

\paragraph{Raw hidden-state baseline.}
To ensure that the comparison between raw representations and relational features is controlled, the raw-state baseline uses the same selected token-layer window and the same common hidden-state mapping. 
Specifically, we define
\begin{equation}
\label{app:eq:raw_state_pooling}
    \mathbf{z}_{i}^{\mathrm{raw}}(\mathbf{Q},\mathbf{A})
    =
    \frac{1}{mr}
    \sum_{\ell\in\mathbf{L}^{\mathrm{sel}}_i}
    \sum_{t\in\mathbf{T}^{\mathrm{sel}}}
    \bar{\mathbf{h}}_{\boldsymbol{\theta}_i,t}^{\ell}.
\end{equation}
Thus, improvements from relational features cannot be explained simply by using more tokens, more layers, or a larger mapped representation space.

\paragraph{Cross-model structural alignment.}
For structural alignment, we compute the feature vectors
$\mathbf{z}_{i}^{r}(\mathbf{Q}^s,\mathbf{A}_i^s)$ and
$\mathbf{z}_{j}^{r}(\mathbf{Q}^s,\mathbf{A}_j^s)$ for the same set of prompts under two LLMs $i$ and $j$. 
For each model, we construct a $k$NN graph over examples using inner-product similarity between the corresponding feature vectors. 
We then report mutual $k$NN overlap as the cross-model alignment score. 
The detailed definition and the choice of $k$ are given in Appendix~\ref{app:knn_alignment}. 
No hallucination labels are used in this structural-alignment computation.

\paragraph{Cross-model predictive transfer.}
For predictive transfer, we train a lightweight MLP detector on the source-model features and evaluate it directly on the target-model features. 
For each ordered source-target pair $(i,j)$ and each feature type $r$, the MLP takes
$\mathbf{z}_{i}^{r}(\mathbf{Q},\mathbf{A}_i)$ as input and predicts the binary truthfulness label. 
The same trained detector is then applied to
$\mathbf{z}_{j}^{r}(\mathbf{Q},\mathbf{A}_j)$ without updating its parameters. 
The detector is a two-layer MLP:
\begin{equation}
    p(y=1\mid \mathbf{z})
    =
    \sigma
    \left(
    \mathbf{w}_2^{\top}
    \mathrm{Dropout}
    \left(
    \mathrm{GELU}
    \left(
    \mathbf{W}_1\mathbf{z}+\mathbf{b}_1
    \right)
    \right)
    +
    b_2
    \right),
\end{equation}
with hidden dimension $256$ and dropout rate $0.1$. 
We train it with binary cross-entropy using AdamW, learning rate $10^{-3}$, weight decay $10^{-4}$, batch size $256$, and early stopping on source-validation AUROC. 
All reported AUROC values are averaged over three random seeds. 
The same MLP architecture and optimization protocol are used for raw and relational features.

\paragraph{Scope of the observation pipeline.}
The observation pipeline uses fixed handcrafted relations and a simple post-hoc MLP only to diagnose whether different relational structures are stable and predictive across models. 
It does not use the learnable relation-aware attention, attention pooling, or meta-learning strategy of \method{}. 
Those components are introduced later in Section~\ref{sec:method} to turn the empirical observation into a trainable cross-model detector.

\subsection{Baseline Implementation Details}
\label{app:baseline_details}

This subsection describes the baselines used in our experiments and how we adapt them to the cross-model hallucination detection (CMHD) setting. 
The main complication is that many training-based hallucination detectors were originally designed for a single LLM, where training and testing features live in the same representation space. 
In CMHD, however, the detector is trained on source LLMs and evaluated on unseen target LLMs, whose hidden dimensions, layer counts, head counts, and representation gauges may differ. 
Therefore, some baselines cannot be applied directly without a compatibility step. 
For such methods, we use the minimal adaptation needed to make them executable under CMHD, while keeping their scoring rule as close as possible to the original design. 
These adaptations should be understood as evaluation-time compatibility operations rather than improvements intended by the original baseline authors.

\paragraph{General evaluation protocol.}
All baselines are evaluated on the same generated answers, labels, and source--target splits as \method{}. 
For every scalar hallucination score, we report AUROC on the target model. 
If a method produces a confidence score rather than a hallucination score, we reverse its sign so that larger values consistently indicate higher hallucination likelihood. 
When the score direction is ambiguous, the direction is selected using source-validation data only, never using target labels. 
For methods requiring thresholds, thresholds are selected on source-validation data, but AUROC is computed from continuous scores whenever available. 
For sampling-based baselines, we use the same sampling budget across methods: unless otherwise stated, we draw $K_{\mathrm{gen}}=10$ additional generations per question with temperature $0.5$ and nucleus sampling $p=0.95$. 
This keeps test-time computation comparable across uncertainty and self-consistency baselines.

\paragraph{Perplexity.}
Perplexity is a training-free likelihood-based baseline. 
Given a question $\mathbf{Q}$ and an answer $\mathbf{A}=(a_1,\ldots,a_T)$ generated by the evaluated LLM, we compute the length-normalized negative log-likelihood
\begin{equation}
    s_{\mathrm{ppl}}(\mathbf{Q},\mathbf{A})
    =
    -\frac{1}{T}
    \sum_{t=1}^{T}
    \log p_{\boldsymbol{\theta}}(a_t \mid \mathbf{Q},a_{<t}).
\end{equation}
Larger values indicate that the model assigns lower probability to its own answer, which is treated as a higher hallucination score. 
This baseline does not require training or representation alignment, and can therefore be applied directly to each target LLM. 
However, since perplexity is sensitive to surface form, answer length, and tokenization, we use length normalization and compute perplexity with the same tokenizer and model that generated the answer.

\paragraph{Semantic Entropy.}
Semantic Entropy estimates uncertainty in meaning space rather than token-sequence space~\citep{farquhar2024detecting}. 
For each question, we sample $K_{\mathrm{gen}}$ additional answers from the same LLM and group the sampled answers into semantic-equivalence clusters using a fixed entailment or LLM-judge module. 
Let $\mathcal{C}$ denote the resulting set of semantic clusters and let $p(c)$ be the normalized probability mass assigned to cluster $c$. 
The score is
\begin{equation}
    s_{\mathrm{se}}(\mathbf{Q})
    =
    -\sum_{c\in\mathcal{C}} p(c)\log p(c).
\end{equation}
A high semantic entropy indicates that the model produces many semantically distinct answers, suggesting uncertainty and a higher hallucination risk. 
This method is training-free and does not rely on hidden-state dimensions, so it is directly applicable to unseen target LLMs. 
For fairness, we use the same generation budget and semantic-equivalence module for all models.

\paragraph{EigenScore.}
EigenScore is a white-box, training-free self-consistency baseline that uses internal states from multiple sampled generations~\citep{chen2024inside}. 
For each question, we sample $K_{\mathrm{gen}}$ answers and extract one sentence-level hidden representation from each sampled answer, following the original method. 
Let $\mathbf{r}_1,\ldots,\mathbf{r}_{K_{\mathrm{gen}}}$ denote these representations and let $\mathbf{\Sigma}$ be their empirical covariance matrix. 
EigenScore is computed from the spectrum of $\mathbf{\Sigma}$; larger spectral dispersion indicates that the generated answers occupy a more diverse internal-state subspace, which is taken as higher hallucination risk. 
Since EigenScore is computed within a single target model at test time and does not train a source-model probe, it does not require cross-model hidden-dimension mapping. 
We therefore apply it directly on each target LLM's hidden states.

\paragraph{Lexical Similarity.}
Lexical Similarity is a black-box self-consistency baseline. 
It samples multiple answers to the same question and measures their pairwise surface-form similarity using a lexical overlap metric such as ROUGE-L or token-level F1. 
For each evaluated answer, we compute its average lexical similarity to the sampled alternatives and use the negative average similarity as the hallucination score:
\begin{equation}
    s_{\mathrm{lex}}(\mathbf{Q},\mathbf{A})
    =
    -\frac{1}{K_{\mathrm{gen}}}
    \sum_{k=1}^{K_{\mathrm{gen}}}
    \mathrm{Sim}_{\mathrm{lex}}(\mathbf{A},\mathbf{A}^{(k)}).
\end{equation}
Low lexical agreement suggests that the model is unstable under resampling, which is treated as higher hallucination risk. 
This method does not depend on hidden states or attention maps, and thus requires no cross-model mapping. 
We keep the same sampling budget as Semantic Entropy to ensure a fair comparison.

\paragraph{Verbalize.}
Verbalize is a prompting-based confidence baseline that asks the LLM to explicitly report its confidence in the generated answer~\citep{lin2022teaching,lin2024generating}. 
After obtaining an answer $\mathbf{A}$, we query the same LLM with a fixed confidence prompt:
\begin{quote}
\small
\texttt{Q: \{question\} A: \{answer\} The proposed answer is true with a confidence value (0--100) of}
\end{quote}
We parse the generated number as a confidence score and use its negative value as the hallucination score. 
If the model does not produce a valid number, we reprompt once with a stricter numerical-output instruction; remaining invalid outputs are assigned the median confidence on the source-validation set. 
This baseline is training-free and black-box, so it is naturally applicable to unseen LLMs. 
However, verbalized confidence is known to be affected by calibration and instruction-following behavior, so we treat it as a behavioral baseline rather than a representation-based detector.

\paragraph{InterrogateLLM.}
InterrogateLLM is a zero-resource hallucination detector that checks whether an answer contains enough reliable information to reconstruct the original query~\citep{yehuda-etal-2024-interrogatellm}. 
Given an answer $\mathbf{A}$, the model is asked to generate a reverse question $\tilde{\mathbf{Q}}$ that would have led to $\mathbf{A}$. 
The original question $\mathbf{Q}$ and reconstructed question $\tilde{\mathbf{Q}}$ are then compared. 
If the answer is faithful, the reconstructed question should be semantically close to the original one; if the answer is hallucinated or drifts from the requested information, reconstruction becomes less reliable. 
We implement the comparison using the same semantic-similarity module across all models and use
\begin{equation}
    s_{\mathrm{int}}(\mathbf{Q},\mathbf{A})
    =
    -\mathrm{Sim}_{\mathrm{sem}}(\mathbf{Q},\tilde{\mathbf{Q}})
\end{equation}
as the hallucination score. 
For tasks where the original query is not naturally reconstructable from the answer alone, such as code generation or agent-tool tasks, we use a task-aware reconstruction prompt that asks the model to reconstruct the user's requested objective rather than the exact surface form of the question. 
This modification is necessary because the original InterrogateLLM protocol was designed primarily for factual question answering; it may deviate from the method's intended operating condition, but it makes the baseline applicable across our heterogeneous benchmark suite.

\paragraph{MM.}
MM denotes the Mass-Mean probe, a simple activation-space truthfulness probe~\citep{marks2023geometry}. 
Given hidden representations from truthful and hallucinated examples, it computes the class means
\begin{equation}
    \boldsymbol{\mu}_{+}
    =
    \frac{1}{|\mathcal{S}_{+}|}
    \sum_{\mathbf{r}\in\mathcal{S}_{+}}\mathbf{r},
    \qquad
    \boldsymbol{\mu}_{-}
    =
    \frac{1}{|\mathcal{S}_{-}|}
    \sum_{\mathbf{r}\in\mathcal{S}_{-}}\mathbf{r},
\end{equation}
and defines the truthfulness direction as
\begin{equation}
    \mathbf{w}_{\mathrm{MM}}
    =
    \boldsymbol{\mu}_{+}
    -
    \boldsymbol{\mu}_{-}.
\end{equation}
The score of a test representation $\mathbf{r}$ is then given by projection onto this direction. 
The original MM probe assumes that training and testing activations live in the same model-specific vector space. 
This assumption fails in CMHD because source and target LLMs may have different hidden dimensions. 
Therefore, for MM we first map hidden states into the same padded space $\mathbb{R}^{d^\star}$ using the hidden-state restructuring operation described in Section~\ref{sec:method}. 
We then train the mass-mean direction on source-model mapped hidden states and evaluate it on target-model mapped hidden states. 
No relation-aware graph construction or meta-learning is used for this baseline.

\paragraph{SEP.}
Semantic Entropy Probes (SEP) train a lightweight probe to approximate Semantic Entropy from a single generation's hidden state~\citep{kossen2024semantic}. 
The purpose is to avoid the high test-time cost of sampling multiple generations required by Semantic Entropy. 
In our implementation, we first compute Semantic Entropy targets on source-model data using the sampling procedure described above. 
Then, a probe is trained to predict these entropy targets from hidden representations of the corresponding single generated answer. 
At test time, the probe predicts a semantic-entropy score from one target-model generation. 
As with MM, the original probe is not directly cross-model because its input dimension is tied to the hidden states of the model on which it was trained. 
We therefore apply the same hidden-state mapping into $\mathbb{R}^{d^\star}$ before training and testing the SEP probe. 
This adaptation makes SEP executable in CMHD, but it should be viewed as a compatibility modification rather than part of the original SEP design.

\paragraph{SAPLMA.}
SAPLMA trains a classifier on LLM activation values to predict whether a generated statement is true or false~\citep{azaria2023internal}. 
It is a representative training-based hidden-state detector. 
For each question-answer pair, we extract the hidden representation of the final answer token from a fixed relative-depth layer. 
Specifically, when a method specifies a single middle layer, we use the layer closest to half of the model depth; this avoids relying on a layer index that only exists for one architecture. 
The extracted hidden state is mapped into the common padded space $\mathbb{R}^{d^\star}$ and used as input to a lightweight MLP classifier. 
The classifier is trained on source LLMs and evaluated directly on the mapped hidden states of the target LLM. 
This mapping is necessary because SAPLMA's original formulation assumes a fixed activation dimension, whereas CMHD evaluates transfer across architectures with different hidden sizes.

\paragraph{LapEigvals.}
LapEigvals is an attention-map-based hallucination detector that interprets attention maps as graph adjacency matrices and uses Laplacian eigenvalues as spectral features~\citep{binkowski-etal-2025-hallucination}. 
Unlike MM, SEP, and SAPLMA, this method does not operate on hidden-state vectors. 
It instead extracts attention maps from the LLM and computes graph spectral statistics from them. 
Consequently, we do not apply hidden-state padding to LapEigvals. 
The cross-model compatibility issue is different: LLMs may have different numbers of layers and attention heads, and their attention tensors may therefore have different shapes. 
To make LapEigvals applicable to CMHD, we map each model's attention information to a common attention-feature format. 
We first select the same number of relative-depth layers and the same selected answer-token window for all models. 
For each selected layer and head, we form the attention submatrix over the selected tokens, symmetrize it, construct its normalized graph Laplacian, and extract the smallest non-trivial eigenvalues. 
To handle different head counts, we aggregate head-wise spectral features using mean and standard deviation within each selected layer. 
The resulting fixed-dimensional attention-spectral vector is used to train a source-model classifier and evaluate on target models. 
This preserves the key idea of LapEigvals, namely detecting hallucinations from spectral properties of attention graphs, while removing architecture-specific tensor-shape dependencies.

\paragraph{Caveat on cross-model adaptations.}
The hidden-state mapping for MM, SEP, and SAPLMA, and the attention-feature mapping for LapEigvals, are introduced only because these baselines are otherwise not directly executable in the CMHD setting. 
They should not be interpreted as improvements to the original methods or as design choices advocated by their authors. 
Rather, they provide the fairest feasible way to compare single-model training-based detectors against a method explicitly designed for cross-model generalization. 
In contrast, training-free baselines such as Perplexity, Semantic Entropy, EigenScore, Lexical Similarity, Verbalize, and InterrogateLLM do not train a source-model representation classifier, so they are applied directly to each target LLM under a unified scoring and sampling protocol.

\subsection{LLM-as-Judge Labeling Protocol}
\label{app:judge_protocol}

We assign hallucination labels using an LLM-as-Judge protocol rather than lexical-overlap heuristics.
Recent re-evaluation work has shown that overlap-based criteria such as ROUGE-L can introduce substantial label noise in hallucination detection, because semantically correct answers may receive low overlap scores while fluent but incorrect answers may still achieve misleading lexical similarity.
We therefore use a semantically aware judge under a fixed protocol and deterministic decoding.

\begin{tcolorbox}[colback=white,colframe=black!12,boxrule=0.4pt,sharp corners,
  top=4pt,bottom=4pt,left=6pt,right=6pt]
\textbf{Protocol summary.}
The evaluator is GPT-4o under a fixed prompt and structured output format.
For each sample, the judge receives: (i) the question, (ii) an optional context, (iii) a list of acceptable truths, (iv) an optional list of known traps (incorrect answers), and (v) the model-generated answer.
The judge returns one of three intermediate categories: \texttt{correct}, \texttt{hallucination}, or \texttt{abstention}.
For the binary detector studied in this paper, \texttt{correct} and \texttt{abstention} are mapped to the positive class, while \texttt{hallucination} is mapped to the negative class.
\end{tcolorbox}

\paragraph{Inputs to the Judge.}
Each evaluation instance is represented by a structured record containing the question, optional supporting context, one or more acceptable truths, optional known traps, and the answer generated by the model under evaluation.
The acceptable truths specify the semantic content that should be accepted as correct, allowing multiple valid phrasings or equivalent answers.
Known traps list common or particularly misleading incorrect answers when such information is available.
This design reduces dependence on exact surface form and makes the evaluation more robust to paraphrase.

\paragraph{Decision Rules.}
The judge is instructed to output exactly one of three categories.
It returns \texttt{correct} if the model answer semantically matches at least one acceptable truth.
It returns \texttt{hallucination} if the answer matches a known trap, contradicts the acceptable truths, or introduces fabricated factual content not supported by the provided truth specification.
It returns \texttt{abstention} if the model explicitly states that it does not know the answer, lacks sufficient information, or refuses to answer on epistemic grounds.
The judge also produces a brief natural-language justification, which is stored for inspection but not used as model input.

\paragraph{Three-Way Intermediate Labels and Binary Mapping.}
We use a three-way intermediate decision layer to improve evaluator precision.
In particular, separating \texttt{abstention} from \texttt{hallucination} prevents explicit uncertainty from being conflated with confident factual error.
For the detector studied in this paper, however, the final supervision is binary.
Accordingly, we map both \texttt{correct} and \texttt{abstention} to label $1$ and map \texttt{hallucination} to label $0$.
This mapping matches the operational goal of the detector: to distinguish fabricated or unsupported factual claims from non-hallucinatory behavior.

\paragraph{Implementation Details.}
The evaluator uses deterministic decoding with temperature $0$ and structured output parsing.
To improve robustness in large-scale asynchronous evaluation, we use bounded retries for transient API failures and isolate samples that still fail after the retry budget is exhausted.
Only successfully parsed judge outputs are retained in the labeled dataset.
The exact prompt template, few-shot examples, and implementation code are included in the released repository.

\paragraph{Why Three-Way Judging Helps.}
The three-way design is not introduced because the final task is ternary, but because the intermediate decision boundary becomes cleaner.
A direct binary judge is more likely to confuse explicit uncertainty with factual error, especially for difficult or underspecified questions.
By first distinguishing \texttt{correct}, \texttt{hallucination}, and \texttt{abstention}, and only then collapsing to a binary label, we obtain a labeling procedure that better reflects the semantic distinction relevant to hallucination detection.

\subsection{Dataset Details and Preprocessing}
\label{app:dataset_details}

This appendix describes the seven benchmarks used in our experiments and the preprocessing pipeline that converts their heterogeneous formats into the unified input-output schema used throughout the paper. 
The datasets are selected to cover complementary hallucination-detection settings: open-domain factual QA, context-grounded QA, conversational QA, mathematical reasoning, code generation, and agent-style tool calling. 
For every benchmark, we first shuffle the original items with a fixed random seed and keep at most $10{,}000$ items. 
If the selected split contains fewer than $10{,}000$ usable items, we keep all available items. 
This case mainly affects benchmarks whose official or selected split is smaller than the cap, such as MBPP. 
We then split the selected items at the \emph{question/task level} into training, validation, and test partitions using an $80/10/10$ ratio. 
The split is performed before LLM generation, so the same question or task does not appear in more than one partition.

For each dataset item, every evaluated LLM receives the same processed prompt and generates its own answer. 
Thus, for a fixed benchmark item indexed by $s$, different models may produce different answers $\mathbf{A}_i^s$, while the underlying question, context, and reference information are shared. 
The generated answer is then labeled as truthful if it is semantically aligned with one of the accepted references or explicitly abstains, and hallucinated if it commits to an answer that matches none of them, following the setup in Section~\ref{sec:setups}. 
The detector training and evaluation are then performed on model-specific datasets of the form in Eq.~\eqref{dataset_format-i}. 
In the leave-one-family-out experiments, training-based methods use only source-family examples for training and validation, and all reported results are computed on the held-out target-family test examples.

\subsubsection{Unified Dataset Schema}
\label{app:dataset_schema}

The raw datasets differ substantially in their field names, task formats, and reference annotations. 
To avoid dataset-specific prompt and evaluation logic, we convert every raw item into a shared structured schema. 
Each processed item stores the following fields:

\begin{itemize}
    \item \texttt{task\_type}: records the task family, such as \texttt{qa}, \texttt{reasoning}, \texttt{coding}, or \texttt{agent\_action}.
    \item \texttt{system\_instruction}: stores task-specific generation requirements, such as answering strictly from a provided context, reasoning step by step, writing valid Python code, or producing a JSON tool call.
    \item \texttt{context}: stores passages, stories, tool descriptions, tests, or other background information needed by the task.
    \item \texttt{question}: stores the actual user-facing query or problem statement.
    \item \texttt{choices}: is used for multiple-choice tasks; for the seven benchmarks in the main experiments, it is empty because all selected tasks are open-ended, reasoning, coding, or structured-generation tasks.
    \item \texttt{ground\_truths}: stores all accepted references, including answer aliases when available.
    \item \texttt{incorrect\_answers}: stores known false or hallucinated answers when the original benchmark provides them. These known incorrect answers are never included in the prompt; they are retained only as evaluation-side traps or reference signals for judging whether a generated answer matches a known hallucination pattern.
\end{itemize}

For reproducibility, each processed JSONL item also keeps the original raw dataset row under an \texttt{original\_doc} field. 
This makes it possible to trace every processed example back to the original benchmark record. 
The prompt given to an LLM is constructed from the schema fields, not directly from the raw dataset row. 
When \texttt{system\_instruction} is non-empty, it is placed before the user query. 
When \texttt{context} is non-empty, it is placed before the question as task evidence or constraints. 
The reference fields \texttt{ground\_truths} and \texttt{incorrect\_answers} are not shown to the model during generation.

\subsubsection{TruthfulQA}
\label{app:dataset_truthfulqa}

TruthfulQA \citep{lin2022truthfulqa} evaluates whether a language model gives truthful answers to questions that are likely to elicit common misconceptions or false imitative responses. 
It is therefore a direct benchmark for factual hallucination and false-belief amplification. 
We use the generation setting, where each item contains a natural-language question, a list of correct answers, and a list of incorrect answers. 
This makes TruthfulQA especially useful for our setting because it provides both accepted factual references and known misleading alternatives.

In our schema, TruthfulQA is treated as an open-ended QA task:
\begin{center}
\texttt{task\_type}=\texttt{qa}.
\end{center}
The original question is mapped to \texttt{question}. 
No additional context is used, so \texttt{context} is set to the empty string. 
The original list of correct answers is copied into \texttt{ground\_truths}, preserving multiple acceptable phrasings. 
The original list of incorrect answers is copied into \texttt{incorrect\_answers}. 
During answer evaluation, a generated answer is considered truthful if it is semantically aligned with any accepted correct answer and does not match the known incorrect alternatives. 
Keeping the incorrect-answer list is important because many TruthfulQA questions are adversarial: a fluent but common answer may be precisely the hallucinated answer that the dataset is designed to expose.

\subsubsection{HaluEval}
\label{app:dataset_halueval}

HaluEval \citep{li-etal-2023-halueval} is designed for hallucination evaluation and contains examples in which models must answer questions based on provided knowledge. 
We use the QA subset. 
Each item provides a knowledge passage, a question, a right answer, and a hallucinated answer. 
This makes the dataset a natural fit for context-grounded hallucination detection: the model must use the supplied evidence rather than produce unsupported statements.

In our schema, HaluEval is treated as a context-grounded QA task:
\begin{center}
\texttt{task\_type}=\texttt{qa}.
\end{center}
The knowledge field is mapped to \texttt{context}, and the question field is mapped to \texttt{question}. 
We use the instruction:
\begin{promptbox}
Please answer the question based strictly on the provided context.
\end{promptbox}
The right answer is stored as a singleton list in \texttt{ground\_truths}. 
The hallucinated answer is stored as a singleton list in \texttt{incorrect\_answers}. 
The hallucinated answer is not shown to the LLM. 
It is preserved only for evaluation, where it helps identify generated answers that are fluent but inconsistent with the provided knowledge. 
This design lets HaluEval test whether the detector can distinguish context-supported answers from answers that sound plausible but are not grounded in the evidence.

\subsubsection{TriviaQA}
\label{app:dataset_triviaqa}

TriviaQA \citep{joshi-etal-2017-triviaqa} is an open-domain factual QA benchmark. 
Questions often have short factual answers, but the same answer may appear in multiple acceptable surface forms. 
This makes the dataset useful for evaluating whether hallucination detectors generalize beyond adversarial truthfulness questions to ordinary knowledge-intensive factual QA.

We use the no-context reading-comprehension configuration, so each item is processed as open-ended QA without an additional passage:
\begin{center}
\texttt{task\_type}=\texttt{qa}.
\end{center}
The original question is mapped to \texttt{question}, and \texttt{context} is set to the empty string. 
For references, we use the answer alias list provided by the dataset whenever available. 
These aliases are stored in \texttt{ground\_truths}. 
If no alias list is available, we fall back to the canonical answer value. 
This alias-preserving processing is important because a generated answer may be semantically correct even when it does not exactly match the canonical surface form. 
No dataset-provided hallucinated answer is available, so \texttt{incorrect\_answers} is empty.

\subsubsection{CoQA}
\label{app:dataset_coqa}

CoQA \citep{reddy-etal-2019-coqa} is a conversational question-answering benchmark grounded in a story. 
Each raw item contains one story and a sequence of conversational questions and answers. 
CoQA is included because it evaluates context-grounded answering in a conversational style, where the model must extract information from a passage rather than rely only on parametric knowledge.

In the main experiments, we use one QA instance from each CoQA story to avoid introducing multiple highly correlated examples from the same story into different splits. 
Specifically, we use the first question-answer turn of each story. 
The story is mapped to \texttt{context}, the first question is mapped to \texttt{question}, and the corresponding first answer is stored in \texttt{ground\_truths}. 
The task is treated as:
\begin{center}
\texttt{task\_type}=\texttt{qa}.
\end{center}
No multiple-choice options are used, and \texttt{incorrect\_answers} is empty. 
This processing keeps one top-level example per story, reducing the risk that different turns from the same story create near-duplicate training and test instances. 
It also gives a clean context-grounded QA setting that complements HaluEval: HaluEval focuses on hallucinated answers against explicit knowledge, while CoQA focuses on reading a narrative context and producing the correct answer.

\subsubsection{GSM8K}
\label{app:dataset_gsm8k}

GSM8K \citep{cobbe2021training} is a grade-school mathematical reasoning benchmark. 
Each item contains a natural-language math word problem and a reference solution. 
We include GSM8K because reasoning hallucinations differ from factual QA hallucinations: an answer may fail because of an incorrect intermediate calculation, an invalid reasoning step, or a wrong final value. 
This benchmark therefore tests whether internal representations encode transferable signals for reasoning correctness.

In our schema, GSM8K is treated as a reasoning task:
\begin{center}
\texttt{task\_type}=\texttt{reasoning}.
\end{center}
The problem statement is mapped to \texttt{question}, and no external context is used. 
We use the instruction:
\begin{promptbox}
Please think step by step and provide the final answer at the end.
\end{promptbox}
The dataset's reference answer is stored in \texttt{ground\_truths}. 
The reference answer may contain both the reasoning trace and the final numeric answer, and we preserve the original string for traceability. 
During evaluation, the final answer is compared against the reference answer, with semantic or normalized numeric matching used when appropriate. 
No dataset-provided incorrect answers are available, so \texttt{incorrect\_answers} is empty. 
By keeping GSM8K separate in the main and ablation tables, we can examine whether methods that work for factual QA also transfer to reasoning-oriented hallucination detection.

\subsubsection{MBPP}
\label{app:dataset_mbpp}

MBPP \citep{austin2021program} is a Python programming benchmark consisting of short natural-language programming tasks, reference code, and assertion-style tests. 
It is included because code correctness is not identical to natural-language factuality. 
A generated program may look plausible but fail on edge cases, violate the required function signature, or implement a subtly different behavior. 
MBPP therefore provides a useful stress test for hallucination detection in code-generation settings.

In our schema, MBPP is treated as a coding task:
\begin{center}
\texttt{task\_type}=\texttt{coding}.
\end{center}
The natural-language programming requirement is mapped to \texttt{question}. 
The provided assertion tests are concatenated and placed in \texttt{context}:
\begin{promptbox}
Your code must pass the following tests:
\end{promptbox}
followed by the test list. We use the instruction:
\begin{promptbox}
You are an expert Python programmer. Write a Python function to solve the problem. Your code must pass the provided assertion tests.
\end{promptbox}
The reference code is stored in \texttt{ground\_truths}. 
The tests are included in the prompt because they are part of the programming task specification and provide functional constraints. 
They are also retained for evaluation-side checking. 
When execution-based checking is available, generated code is evaluated against the provided assertions; otherwise, the reference code and tests are used by the evaluator to judge functional equivalence. 
No dataset-provided incorrect programs are available, so \texttt{incorrect\_answers} is empty.

MBPP has fewer available items than our $10{,}000$-item cap in the configuration used here, so we keep all available examples after shuffling and apply the same train/validation/test split ratio. 
Because the number of training examples is much smaller than in the other benchmarks, MBPP is expected to be a more fragile setting for training-based detectors. 
This is why the main text reports MBPP separately rather than only as part of a pooled average.

\subsubsection{xLAM Function Calling}
\label{app:dataset_xlam}

The xLAM function-calling benchmark \citep{liu2024apigen} evaluates agent-style tool use. 
Each item contains a user query, a set of available tools, and the expected tool-call answer. 
Unlike ordinary QA, the output is often a structured action rather than a free-form natural-language answer. 
We include xLAM because hallucination in agentic settings often appears as selecting a nonexistent or inappropriate tool, filling incorrect arguments, or producing a tool call that does not satisfy the user's request. 
This setting tests whether hallucination detectors transfer to structured action generation.

In our schema, xLAM is treated as an agent-action task:
\begin{center}
\texttt{task\_type}=\texttt{agent\_action}.
\end{center}
The available tool descriptions are placed in \texttt{context}, prefixed by:
\begin{promptbox}
Available Tools:
\end{promptbox}
The user query is mapped to \texttt{question}. 
We use the instruction:
\begin{promptbox}
You are a helpful assistant with access to various tools. Based on the User's question, select the appropriate tool from the Context and output the exact tool call in JSON format. If no tool is needed, answer directly.
\end{promptbox}
The expected tool-call output is stored in \texttt{ground\_truths}. 
Because tool-call outputs are structured, we canonicalize JSON strings when possible before evaluation, so superficial formatting differences do not count as hallucinations. 
The evaluator checks whether the generated tool name and arguments are semantically equivalent to the reference action. 
No dataset-provided incorrect tool calls are used, so \texttt{incorrect\_answers} is empty.

xLAM complements the QA and coding benchmarks by testing hallucination detection for agent-oriented behavior. 
In this setting, correctness depends not only on factual content but also on whether the generated action is executable and appropriate for the tool inventory. 
This makes xLAM useful for assessing whether cross-model detectors capture broader internal signals of task-grounded correctness.

\subsubsection{Summary of Dataset Roles}
\label{app:dataset_roles}

The seven benchmarks are intentionally heterogeneous. 
TruthfulQA focuses on adversarial factual truthfulness and common misconceptions. 
HaluEval tests context-grounded hallucination against provided knowledge and includes explicit hallucinated answers. 
TriviaQA tests open-domain factual recall with multiple valid answer aliases. 
CoQA tests context-grounded conversational QA over stories. 
GSM8K tests mathematical reasoning correctness. 
MBPP tests functional code generation and execution-oriented correctness. 
xLAM tests structured agent actions and tool-call validity.

This diversity is important for the cross-model hallucination detection setting. 
A detector that transfers only on one type of data may be exploiting task-specific artifacts rather than model-general hallucination signals. 
By evaluating across factual QA, context-grounded QA, reasoning, code, and agent-action settings, we test whether the proposed detector captures internal signals that remain useful across both model families and task formats.

\section{Related Work}
\label{app:related_work}

\paragraph{Hallucination Detection.}
A large body of work studies hallucination detection for LLM outputs using signals from output probabilities, answer consistency, internal activations, or attention structures \citep{10.1145/3571730,minaee2024large}. 
Uncertainty-based methods estimate reliability from likelihood, perplexity, entropy, or semantic uncertainty, and often avoid task-specific detector training \citep{farquhar2024detecting}. 
Consistency-based methods sample multiple responses or construct auxiliary queries, then infer hallucination risk from disagreement, contradiction, or answer instability \citep{manakul-etal-2023-selfcheckgpt}. 
Another line trains lightweight detectors on hidden states, activation patterns, or semantic uncertainty features to predict truthfulness from a single forward pass \citep{azaria2023internal,kossen2024semantic,chen2024inside,internalinspector24}. 
Recent attention-based approaches further explore structural statistics of attention maps for hallucination detection \citep{binkowski-etal-2025-hallucination}. 
Although these methods are effective in many settings, most of them are designed for a fixed LLM or rely on output-level proxies, and therefore do not directly address detector transfer across unseen LLMs.

\paragraph{Robustness and Cross-Model Generalization.}
Our work is also related to broader studies on robustness, distributional generalization, and representation mismatch in modern AI systems. 
Theoretical and empirical studies on OOD robustness investigate when reliable separation is possible and how stable signals can be exploited under distribution shift \citep{NEURIPS2022_f0e91b13,FANG2025121989}. 
In parallel, representation similarity research studies how neural networks differ in their intermediate activations and how representations can be compared across layers, training runs, or architectures \citep{kornblith2019similarity, klabunde2025similarity}. 
These works indicate that models with similar functionality may still have substantially different internal representation spaces, making direct transfer of learned probes difficult. 
CMHD follows this motivation but focuses on hallucination detection across LLMs, where a detector trained on source models must remain effective on target models with different architectures, scales, or training procedures.

\paragraph{Relational and Graph-based Representation Learning.}
Relational modeling provides a natural way to move beyond isolated hidden-state vectors. 
Graph neural networks and message-passing frameworks aggregate information through node-edge interactions and produce graph-level representations through pooling or readout operations \citep{gilmer2017neural}. 
Relation-aware attention further incorporates positional or edge information into attention computation, allowing the model to learn interactions conditioned on relations rather than treating all pairs uniformly \citep{shaw2018self}. 
Attention pooling and related readout mechanisms summarize variable-size structures into compact representations while reducing sensitivity to node ordering \citep{lin2017structured}. 
These ideas are particularly relevant to LLM internals, where hallucination evidence may appear not only in individual token states but also in relations across tokens, layers, and feature groups.

Building on these threads, our work formulates \emph{cross-model hallucination detection} and studies relational structures as transferable signals across LLMs. 
Instead of directly training detectors on model-specific hidden vectors, \method{} represents hidden states as joint layer-token-feature graphs, learns relation-aware graph representations, and uses meta-learning to encourage transfer to unseen target models. 
This design connects hallucination detection with cross-model robustness and relational representation learning, while targeting a setting that has received limited attention in prior work.

\section{Limitations}
\label{app:limitations}
\method{} is designed for white-box or partially transparent LLM settings where hidden states can be accessed. 
It therefore cannot be directly applied to purely black-box APIs that expose only generated text or token probabilities. 
This limits its use in scenarios where model internals are unavailable. 
However, hallucination detection is often deployed in controlled enterprise or platform-side settings, where the service provider has access to model activations or can instrument the inference stack. 
In such deployments, the white-box requirement is less restrictive, and \method{} can be used as an internal reliability module for monitoring generated answers.

\section{Computational Infrastructure}
\label{app:compute}

All experiments were conducted on a server equipped with four NVIDIA H100 GPUs. 
We use this setup because LLM inference itself requires substantial GPU memory, especially for the 70B model, and hidden-state extraction further increases memory usage. 
Although most models other than the 70B model can be run with less restrictive hardware, we use the same 4$\times$H100 environment throughout to keep the extraction and evaluation pipeline consistent. 
After hidden representations are extracted, detector training and evaluation are much lighter and can in principle be performed on a consumer-grade GPU such as an NVIDIA RTX 5090; nevertheless, we keep the full experimental pipeline on the same 4$\times$H100 setup for consistency.

\section{Licenses for Existing Assets}
\label{app:licenses}

All models and datasets utilized in our extensive cross-family transfer experiments are open-source or available for research purposes. Table~\ref{tab:asset_licenses} details the specific licenses for the evaluated large language models and benchmarks. We strictly adhered to the usage terms prescribed by their respective creators. Notably, for datasets with mixed licenses like CoQA, and research-only assets like xLAM, we ensured our empirical evaluation process complies with their non-commercial and academic usage guidelines.

\begin{table}[h!]
    \centering
    \caption{Summary of licenses for all existing models and datasets used in this paper.}
    \label{tab:asset_licenses}
    \begin{adjustbox}{max width=0.8\textwidth}
    \begin{tabular}{lllp{6.5cm}}
        \toprule
        \textbf{Asset Name} & \textbf{Type} & \textbf{License} & \textbf{Notes} \\
        \midrule
        Llama-3.1-8B & Model & Llama 3.1 Community & Follows Meta's specific community agreement. \\
        Llama-3.2-3B & Model & Llama 3.2 Community & Follows Meta's specific community agreement. \\
        Llama-3.3-70B & Model & Llama 3.3 Community & Follows Meta's specific community agreement. \\
        Ministral-3-8B/14B & Model & Apache 2.0 & Mistral AI open-source license. \\
        Qwen3-8B/14B & Model & Apache 2.0 & Alibaba Cloud open-source license. \\
        Qwen3.5-35B-A3B & Model & Apache 2.0 & Alibaba Cloud open-source license. \\
        \midrule
        TruthfulQA & Dataset & Apache 2.0 & Stated in the official GitHub repository. \\
        HaluEval & Dataset & MIT & Stated in the official GitHub repository. \\
        TriviaQA & Dataset & Apache 2.0 & Stated in the official GitHub repository. \\
        CoQA & Dataset & Mixed License & Includes CC BY-SA 4.0 (Wiki/Literature) and Apache 2.0 (CNN). \\
        GSM8K & Dataset & MIT & Stated in the OpenAI GitHub repository. \\
        MBPP & Dataset & CC BY 4.0 & Stated in the official Google dataset release. \\
        xLAM (APIGen) & Dataset & CC BY 4.0 & Released for research purposes only (official data card). \\
        \bottomrule
    \end{tabular}
    \end{adjustbox}
\end{table}

\section{Observation Analysis Details}
\label{app:observation_details}

\subsection{Alignment and Transfer Are Related but Decoupled}
\label{app:obs_alignment_transfer_decoupling}

\begin{figure}[t]
    \centering
    \includegraphics[width=\linewidth]{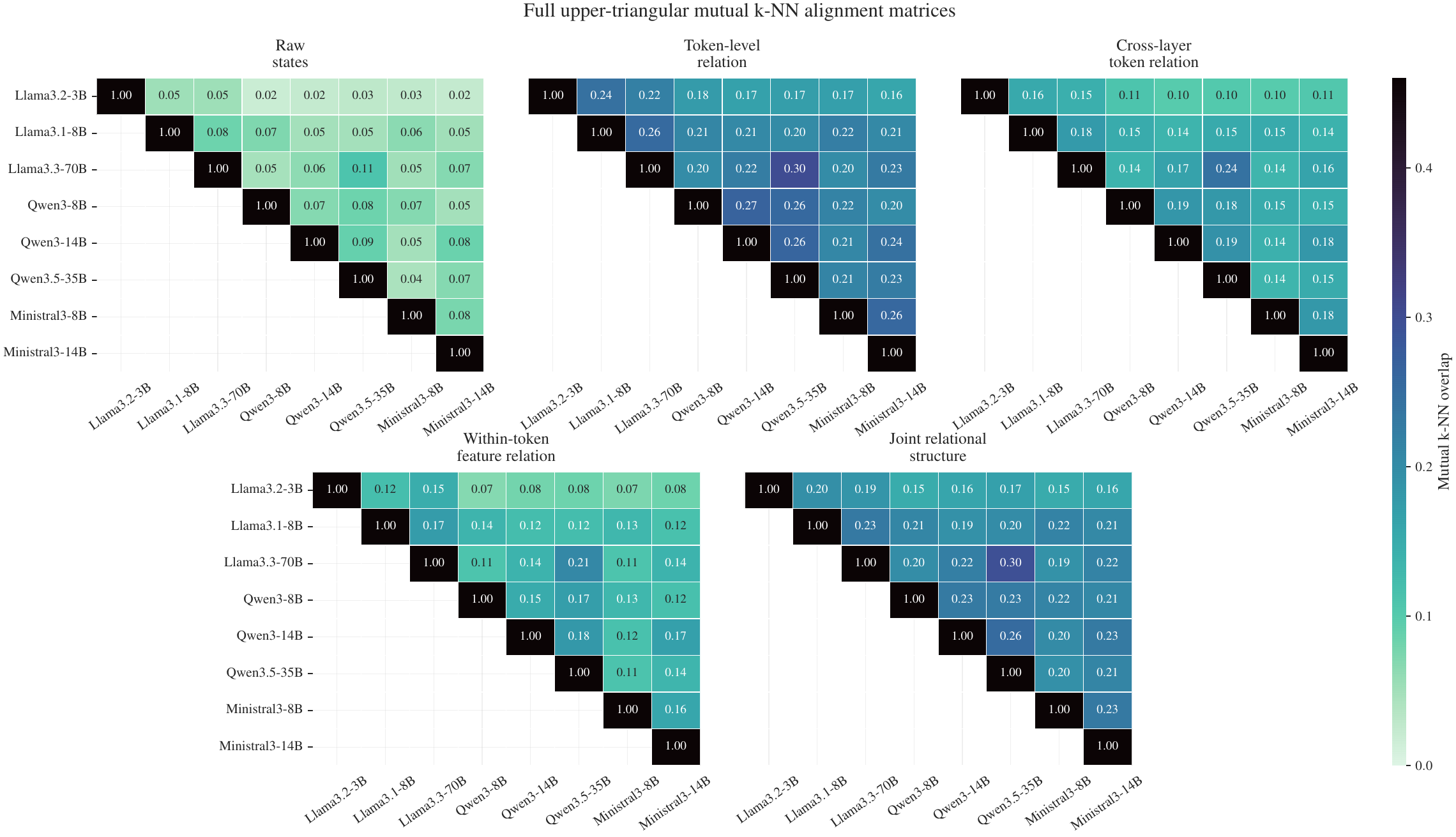}
    \caption{
    Full mutual $k$ nearest neighbor alignment matrices across all model pairs.
    For each feature view, we compute the average overlap between the $k$ nearest neighbor sets induced by two models over the same evaluation examples. Since this metric is symmetric, we show the upper triangular matrix and set the diagonal to one. Across the full model suite, relational views consistently yield stronger local-neighborhood agreement than raw hidden states, supporting the representative trends shown in the main text.
    }
    \label{fig:app_alignment_matrices}  
\end{figure}

\begin{figure}[t]
    \centering
    \includegraphics[width=\linewidth]{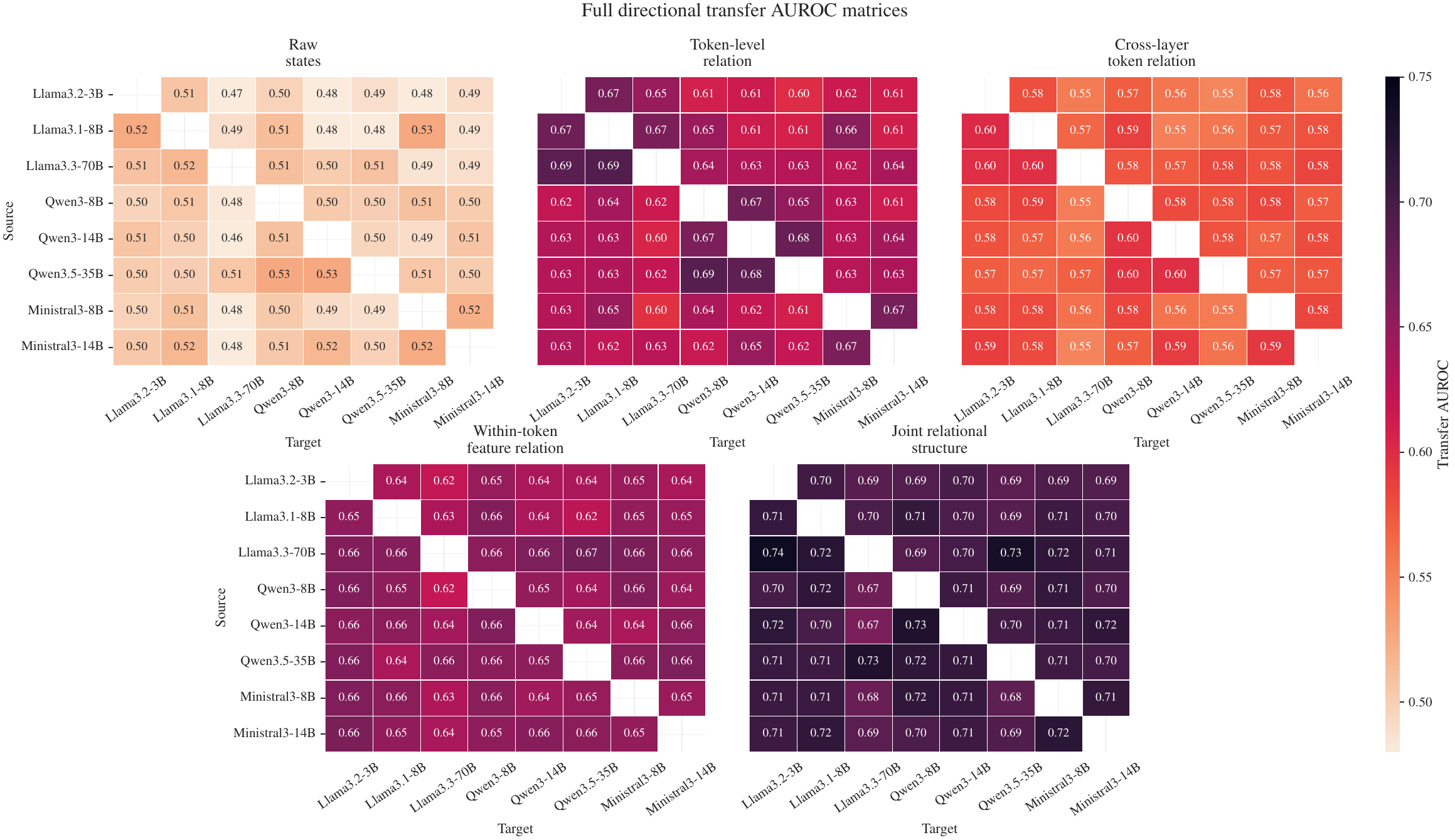}
    \caption{
    Full directional cross-model hallucination transfer matrices.
    Rows denote source models used for training the detector, and columns denote target models used for evaluation. Unlike representational alignment, transfer AUROC is directional, so we show the full matrix. Raw hidden states remain close to chance for most source-to-target pairs, whereas relational views produce substantially stronger transfer across model families and scales.
    }
    \label{fig:app_detection_matrices}
\end{figure}

\begin{figure}[t]
    \centering
    \includegraphics[width=\linewidth]{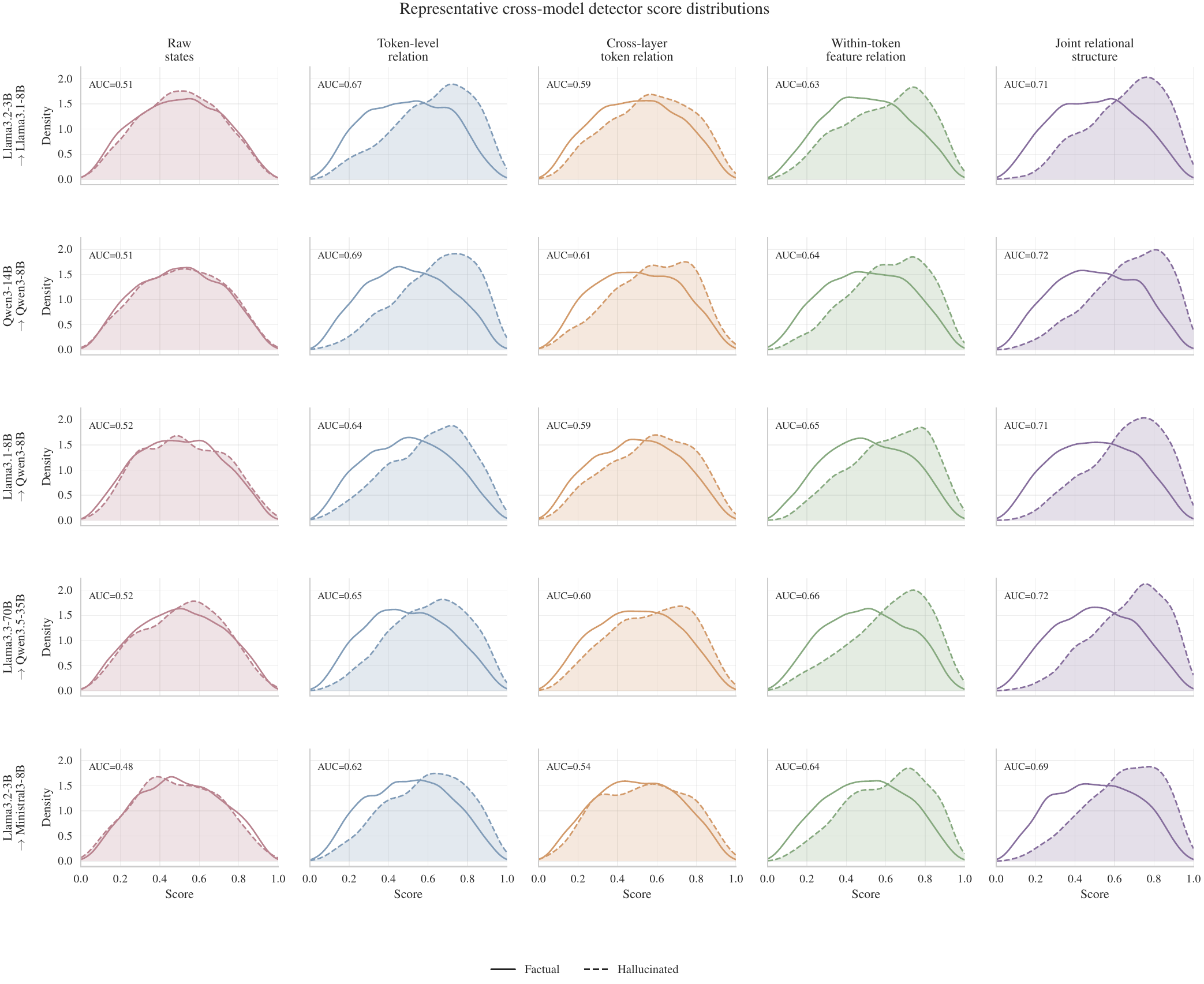}
    \caption{
    Representative detector score distributions across source-to-target model pairs.
    Each row corresponds to one representative transfer pair, and each column corresponds to one feature view. Curves show the score distributions of factual and hallucinated examples on the target model, and the annotated AUROC is recomputed from the plotted sample. The main text KDE plot is one representative case; here we show that the same qualitative pattern holds across the selected same-family, cross-family, scale-matched, and scale-mismatched settings.
    }
    \label{fig:app_kde_representative}
\end{figure}

\begin{figure}[t]
    \centering
    \includegraphics[width=\linewidth]{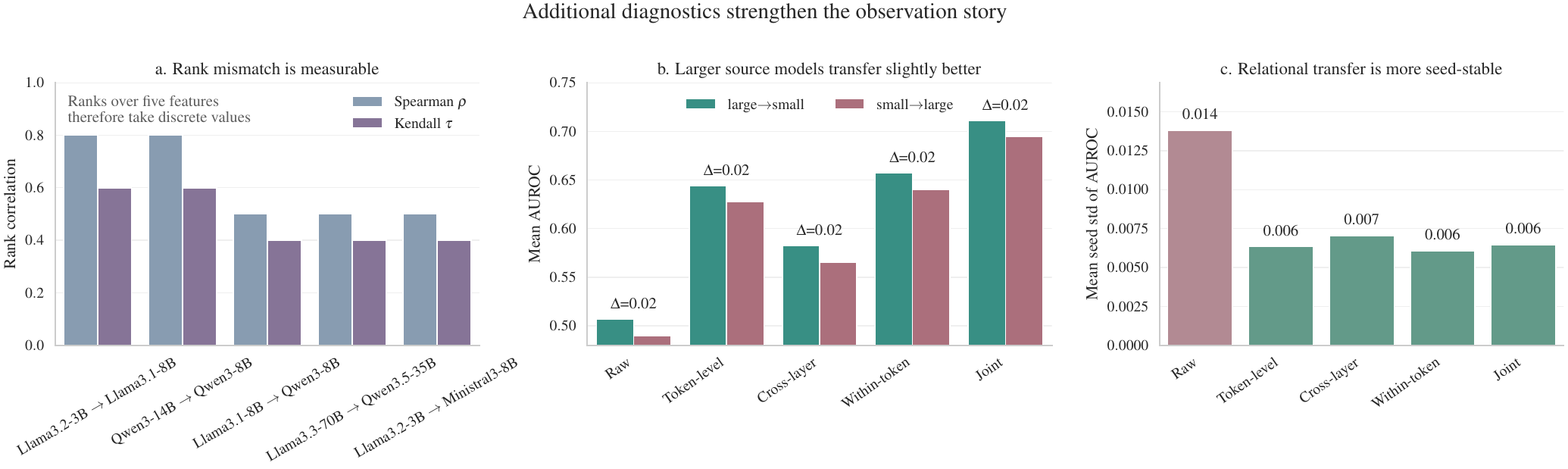}
    \caption{
    Additional diagnostics for the observation analysis.
    (A) Rank correlations between feature-view orderings induced by mutual $k$ nearest neighbor alignment and by transfer AUROC. The moderate correlations indicate that alignment and downstream utility are related but not equivalent. Because the rank correlations are computed over five feature views, the possible Spearman and Kendall values are discrete.
    (B) Cross-scale transfer is mildly asymmetric: detectors trained on larger source models transfer slightly better to smaller target models than the reverse direction.
    (C) Raw hidden states exhibit higher seed variance in transfer AUROC, while relational views are more stable across random seeds.
    }
    \label{fig:app_quant_checks}
\end{figure}

This section expands the observational analysis in Section~\ref{sec:observations}. 
The goal is to examine whether cross-model representational alignment is sufficient to explain downstream hallucination-detector transfer. 
We study five feature views: raw hidden states, token-level relations, cross-layer token relations, within-token feature relations, and joint relational structures. 
All relational views in this section are fixed and hand-designed diagnostic views. 
They are not trained end-to-end and do not use the relation-aware attention module introduced in Section~\ref{sec:method}. 
They are used only to test what kinds of internal structures remain stable and useful across LLMs.

\paragraph{Feature views used in the diagnostic analysis.}
For each model $\mathbb{P}_{\boldsymbol{\theta}_i}$ and each generated example $(\mathbf{Q},\mathbf{A}_i)$, we extract hidden states 
$\mathbf{h}_{\boldsymbol{\theta}_i,t}^{\ell}(\mathbf{Q},\mathbf{A}_i)$ 
as defined in Section~\ref{sec:setups}. 
Since different LLMs have different hidden dimensions, we use the same scattered-padding map as in Eq.~\eqref{eq:index_map} and Eq.~\eqref{eq:block_partition}. 
The padded hidden state is denoted by
\[
    \bar{\mathbf{h}}_{\boldsymbol{\theta}_i,t}^{\ell}(\mathbf{Q},\mathbf{A}_i)
    \in\mathbb{R}^{d^{\star}}.
\]
Unless otherwise stated, all feature-block operations in this appendix are applied to the padded hidden states 
$\bar{\mathbf{h}}_{\boldsymbol{\theta}_i,t}^{\ell}$. 
The scattered-padding map is injective for each model, so within-model dot products, cosine similarities, and Euclidean distances are preserved after padding.

The raw-state view uses the same selected hidden states as the relational diagnostics, but pools them directly without relation-weighted propagation. 
Let $\mathbf{T}^{\mathrm{sel}}=\{t_1,\ldots,t_r\}$ denote the selected token positions and 
$\mathbf{L}^{\mathrm{sel}}=\{\ell_1,\ldots,\ell_{m}\}$ denote the selected layers, with $m=|\mathbf{L}^{\mathrm{sel}}|$. 
Following Eq.~(\ref{app:eq:raw_state_pooling}), we define
\begin{equation}
\label{app:eq:raw_state_view}
    \mathbf{z}_{i}^{\mathrm{raw}}(\mathbf{Q},\mathbf{A}_i)
    =
    \frac{1}{mr}
    \sum_{\ell\in\mathbf{L}^{\mathrm{sel}}_i}
    \sum_{t\in\mathbf{T}^{\mathrm{sel}}}
    \bar{\mathbf{h}}_{\boldsymbol{\theta}_i,t}^{\ell}(\mathbf{Q},\mathbf{A}_i).
\end{equation}
This view keeps the selected hidden-state coordinates in the unified representation space, but it does not explicitly encode relations among tokens, layers, or feature blocks.

For the four relational views, we use a common relation-weighted propagation and mean-readout template. 
This notation is only used to describe the fixed diagnostic pipeline compactly. 
For a relation type $q$, let $\mathcal{V}_{i}^{q}$ denote its node set, 
$\mathbf{x}_{i,u}^{q}(\mathbf{Q},\mathbf{A}_i)$ the feature of node $u\in\mathcal{V}_{i}^{q}$, and 
$R_{i}^{q}(u,v;\mathbf{Q},\mathbf{A}_i)$ the hand-designed relation value between nodes $u$ and $v$. 
The graph-level relational feature is
\begin{equation}
\label{app:eq:general_relation_readout}
    \mathbf{z}_{i}^{q}(\mathbf{Q},\mathbf{A}_i)
    =
    \frac{1}{|\mathcal{V}_{i}^{q}|}
    \sum_{u\in\mathcal{V}_{i}^{q}}
    \sum_{v\in\mathcal{V}_{i}^{q},\,v\neq u}
    \frac{
        R_{i}^{q}(u,v;\mathbf{Q},\mathbf{A}_i)
    }{
        \epsilon+
        \sum_{v'\in\mathcal{V}_{i}^{q},\,v'\neq u}
        R_{i}^{q}(u,v';\mathbf{Q},\mathbf{A}_i)
    }
    \mathbf{x}_{i,v}^{q}(\mathbf{Q},\mathbf{A}_i),
\end{equation}
where $\epsilon>0$ is a small numerical-stability constant. 
Here $R_i^q$ is treated as a relation magnitude rather than a similarity score. 
Thus, this diagnostic readout emphasizes nodes that have larger relative pairwise differences from the receiving node. 
The final mean readout is permutation-invariant over node orderings, which is useful because the same relational construction may be implemented with different internal node orders across LLMs.

The token-level relation view uses selected tokens as nodes. 
For notational simplicity, we write $\mathbf{z}_{i}^{\mathrm{tok}}$ for the token-level feature 
$\mathbf{z}_{i}^{\mathrm{tok},\ell}$ computed at the selected diagnostic layer $\ell$. 
The node set is 
$\mathcal{V}_{i}^{\mathrm{tok}}=\mathbf{T}^{\mathrm{sel}}$, and the node feature is
\[
    \mathbf{x}_{i,t}^{\mathrm{tok}}(\mathbf{Q},\mathbf{A}_i)
    =
    \bar{\mathbf{h}}_{\boldsymbol{\theta}_i,t}^{\ell}(\mathbf{Q},\mathbf{A}_i).
\]
The edge relation follows the token-distance relation in Eq.~\eqref{Token-Relation-1}, implemented on padded hidden states:
\begin{equation}
\label{app:eq:tok_relation_padded}
     R_{i}^{\mathrm{tok}}(t,\tilde{t};\mathbf{Q},\mathbf{A}_i)
    =
    s^{\mathrm{tok}}(t,\tilde{t})
    \big\|
        \bar{\mathbf{h}}_{\boldsymbol{\theta}_i,t}^{\ell}(\mathbf{Q},\mathbf{A}_i)
        -
        \bar{\mathbf{h}}_{\boldsymbol{\theta}_i,\tilde{t}}^{\ell}(\mathbf{Q},\mathbf{A}_i)
    \big\|_2 .
\end{equation}
Applying Eq.~\eqref{app:eq:general_relation_readout} with this node set and relation value gives 
$\mathbf{z}_{i}^{\mathrm{tok}}(\mathbf{Q},\mathbf{A}_i)$.

The cross-layer token relation view uses selected layers as nodes. 
For each layer $\ell\in\mathbf{L}^{\mathrm{sel}}$, we stack the padded hidden states of all selected tokens:
\begin{equation}
\label{app:eq:padded_layer_matrix}
    \bar{\mathbf{H}}_{\boldsymbol{\theta}_i}^{\ell,\mathrm{sel}}(\mathbf{Q},\mathbf{A}_i)
    =
    \big[
        \bar{\mathbf{h}}_{\boldsymbol{\theta}_i,t_1}^{\ell}(\mathbf{Q},\mathbf{A}_i),
        \ldots,
        \bar{\mathbf{h}}_{\boldsymbol{\theta}_i,t_r}^{\ell}(\mathbf{Q},\mathbf{A}_i)
    \big]^{\top}
    \in\mathbb{R}^{r\times d^{\star}}.
\end{equation}
The node associated with layer $\ell$ is represented by
\[
    \mathbf{x}_{i,\ell}^{\mathrm{lyr}}(\mathbf{Q},\mathbf{A}_i)
    =
    \mathrm{vec}
    \left(
        \bar{\mathbf{H}}_{\boldsymbol{\theta}_i}^{\ell,\mathrm{sel}}(\mathbf{Q},\mathbf{A}_i)
    \right).
\]
The relation between two layer nodes is the depth-decayed cosine distance:
\begin{equation}
\label{app:eq:padded_layer_relation}
    R_{i}^{\mathrm{lyr}}(\ell,\tilde{\ell};\mathbf{Q},\mathbf{A}_i)
    =
    s^{\mathrm{lyr}}(\ell,\tilde{\ell})\,
    \big(
        1-
        \cos
        \big(
            \mathrm{vec}(\bar{\mathbf{H}}_{\boldsymbol{\theta}_i}^{\ell,\mathrm{sel}}),
            \mathrm{vec}(\bar{\mathbf{H}}_{\boldsymbol{\theta}_i}^{\tilde{\ell},\mathrm{sel}})
        \big)
    \big),
\end{equation}
where $s^{\mathrm{lyr}}(\ell,\tilde{\ell})=\exp(-|\ell-\tilde{\ell}|/\tau_{\mathrm{lyr}})$ is the depth-decay factor defined in Section~\ref{sec:joint_relations}. 
Applying Eq.~\eqref{app:eq:general_relation_readout} over 
$\mathcal{V}_{i}^{\mathrm{lyr}}=\mathbf{L}^{\mathrm{sel}}$ gives 
$\mathbf{z}_{i}^{\mathrm{lyr}}(\mathbf{Q},\mathbf{A}_i)$.

The within-token feature relation view uses feature blocks as nodes. 
The padded hidden dimension is partitioned into $B$ equal-size contiguous blocks with index sets $\mathcal{I}_b$ as in Eq.~\eqref{eq:block_partition}. 
For each block $b$ and each selected layer $\ell$, we first collect the $b$-th hidden-state block of all selected tokens:
\begin{equation}
\label{app:eq:padded_feat_layer_matrix}
    \bar{\mathbf{V}}_{\boldsymbol{\theta}_i,b}^{\ell}(\mathbf{Q},\mathbf{A}_i)
    =
    \big[
        \bar{\mathbf{h}}_{\boldsymbol{\theta}_i,t_1}^{\ell}(\mathbf{Q},\mathbf{A}_i)[\mathcal{I}_b],
        \ldots,
        \bar{\mathbf{h}}_{\boldsymbol{\theta}_i,t_r}^{\ell}(\mathbf{Q},\mathbf{A}_i)[\mathcal{I}_b]
    \big]^{\top}
    \in \mathbb{R}^{r\times |\mathcal{I}_b|}.
\end{equation}
We then stack these token-block matrices vertically over the selected layers:
\begin{equation}
\label{app:eq:padded_feat_matrix}
    \bar{\mathbf{V}}_{\boldsymbol{\theta}_i,b}(\mathbf{Q},\mathbf{A}_i)
    =
    \big[
        (\bar{\mathbf{V}}_{\boldsymbol{\theta}_i,b}^{\ell_1}(\mathbf{Q},\mathbf{A}_i))^{\top},
        \ldots,
        (\bar{\mathbf{V}}_{\boldsymbol{\theta}_i,b}^{\ell_{|\mathbf{L}^{\mathrm{sel}}|}}(\mathbf{Q},\mathbf{A}_i))^{\top}
    \big]^{\top}
    \in \mathbb{R}^{|\mathbf{L}^{\mathrm{sel}}|r\times|\mathcal{I}_b|}.
\end{equation}
The node feature for block $b$ is
\[
    \mathbf{x}_{i,b}^{\mathrm{feat}}(\mathbf{Q},\mathbf{A}_i)
    =
    \mathrm{vec}
    \left(
        \bar{\mathbf{V}}_{\boldsymbol{\theta}_i,b}(\mathbf{Q},\mathbf{A}_i)
    \right).
\]
The relation between two block nodes is
\begin{equation}
\label{app:eq:padded_feat_relation}
    R_{i}^{\mathrm{feat}}(b,\tilde{b};\mathbf{Q},\mathbf{A}_i)
    =
    \big\|
        \mathrm{vec}(\bar{\mathbf{V}}_{\boldsymbol{\theta}_i,b}(\mathbf{Q},\mathbf{A}_i))
        -
        \mathrm{vec}(\bar{\mathbf{V}}_{\boldsymbol{\theta}_i,\tilde{b}}(\mathbf{Q},\mathbf{A}_i))
    \big\|_2.
\end{equation}
No decay term is applied, since feature-block indices do not carry a natural ordering. 
Applying Eq.~\eqref{app:eq:general_relation_readout} over 
$\mathcal{V}_{i}^{\mathrm{feat}}=\{1,\ldots,B\}$ gives 
$\mathbf{z}_{i}^{\mathrm{feat}}(\mathbf{Q},\mathbf{A}_i)$.

The joint relation view uses each layer-token-feature triple $(\ell,t,b)$ as a node. 
The node set is
\[
    \mathcal{V}_{i}^{\mathrm{joint}}
    =
    \mathbf{L}^{\mathrm{sel}}\times \mathbf{T}^{\mathrm{sel}}\times \{1,\ldots,B\}.
\]
The node feature is
\[
    \mathbf{x}_{i,(\ell,t,b)}^{\mathrm{joint}}(\mathbf{Q},\mathbf{A}_i)
    =
    \bar{\mathbf{h}}_{\boldsymbol{\theta}_i,t}^{\ell}(\mathbf{Q},\mathbf{A}_i)[\mathcal{I}_b].
\]
The relation between two joint nodes is the layer- and token-decayed block distance:
\begin{equation}
\label{app:eq:padded_joint_relation}
    R_{i}^{\mathrm{joint}}\big((\ell,t,b),(\tilde{\ell},\tilde{t},\tilde{b});\mathbf{Q},\mathbf{A}_i\big)
    =
    s^{\mathrm{lyr}}(\ell,\tilde{\ell})\,s^{\mathrm{tok}}(t,\tilde{t})\,
    \big\|
        \bar{\mathbf{h}}_{\boldsymbol{\theta}_i,t}^{\ell}(\mathbf{Q},\mathbf{A}_i)[\mathcal{I}_b]
        -
        \bar{\mathbf{h}}_{\boldsymbol{\theta}_i,\tilde{t}}^{\tilde{\ell}}(\mathbf{Q},\mathbf{A}_i)[\mathcal{I}_{\tilde{b}}]
    \big\|_2.
\end{equation}
Applying Eq.~\eqref{app:eq:general_relation_readout} over 
$\mathcal{V}_{i}^{\mathrm{joint}}$ gives 
$\mathbf{z}_{i}^{\mathrm{joint}}(\mathbf{Q},\mathbf{A}_i)$.

\paragraph{Structural alignment.}
For each feature view 
$q\in\{\mathrm{raw},\mathrm{tok},\mathrm{lyr},\mathrm{feat},\mathrm{joint}\}$, 
we compute mutual $k$ nearest-neighbor alignment between two LLMs 
$\mathbb{P}_{\boldsymbol{\theta}_i}$ and $\mathbb{P}_{\boldsymbol{\theta}_j}$ over the same evaluation questions. 
For sample index $s$, let 
$\mathcal{N}_{k}^{i,q}(s)$ denote the indices of the $k$ nearest neighbors of the $s$-th example under feature view $q$ for model $i$:
\[
\mathcal{N}_{k}^{i,q}(s)
=
\operatorname{Top}k_{s' \in \{1,\ldots,n\}\setminus\{s\}}
\,
\mathrm{sim}_{i}^{q}(s,s'),
\]
where
\[
\mathrm{sim}_{i}^{q}(s,s')
=
\left\langle
\mathbf{z}_{i}^{q}(\mathbf{Q}^{s},\mathbf{A}_{i}^{s}),
\mathbf{z}_{i}^{q}(\mathbf{Q}^{s'},\mathbf{A}_{i}^{s'})
\right\rangle .
\]
The mutual $k$NN alignment is then
\begin{equation}
\label{app:eq:general_mnn}
m_{\mathrm{NN}}^{q}(i,j)
    =
    \frac{1}{nk}\sum_{s=1}^{n}
    \big|\mathcal{N}_{k}^{i,q}(s)\cap\mathcal{N}_{k}^{j,q}(s)\big|.
\end{equation}
This quantity measures whether the local neighborhood structure induced by a feature view is preserved across two LLMs. 
Since it is symmetric in $i$ and $j$, Figure~\ref{fig:app_alignment_matrices} shows the upper-triangular matrix and sets the diagonal entries to one by convention.

\paragraph{Predictive transfer.}
Structural alignment alone does not fully determine whether a feature view is useful for hallucination detection. 
We therefore also measure directional source-to-target detector transfer. 
For a source model $\mathbb{P}_{\boldsymbol{\theta}_i}$ and a target model $\mathbb{P}_{\boldsymbol{\theta}_j}$, an MLP detector is trained on
\[
    \{(\mathbf{z}_{i}^{q}(\mathbf{Q}^{s},\mathbf{A}_{i}^{s}),y_i^s)\}_{s=1}^{n}
\]
from $\mathcal{D}_i$ and evaluated directly on
\[
    \{(\mathbf{z}_{j}^{q}(\mathbf{Q}^{s},\mathbf{A}_{j}^{s}),y_j^s)\}_{s=1}^{n}
\]
from $\mathcal{D}_j$. 
The resulting AUROC is denoted by 
$a_{\mathrm{tr}}^{q}(i,j)$. 
Unlike mutual $k$NN alignment, transfer AUROC is directional because the detector is trained on one model and evaluated on another. 
Accordingly, Figure~\ref{fig:app_detection_matrices} reports the full source-to-target matrices.

\paragraph{Score-distribution diagnostic.}
To examine why AUROC changes across feature views, we also visualize the detector score distributions on the target model. 
The detector score is oriented so that larger values indicate stronger evidence for the hallucinated label. 
For a trained detector from source $i$ to target $j$, the predicted scores on factual and hallucinated target examples are plotted separately. 
If the two distributions strongly overlap, the detector cannot distinguish the two labels after transfer. 
If the distributions separate, the detector has learned a feature direction that remains discriminative on the target model. 
Figure~\ref{fig:app_kde_representative} shows that raw hidden states yield highly overlapping score distributions, whereas relational views, especially the joint relational structure, produce clearer separation between factual and hallucinated examples.

\FloatBarrier
\begin{wrapfigure}{R}{0.45\textwidth}
    \vspace{-1em} 
    \centering

    \includegraphics[width=\linewidth]{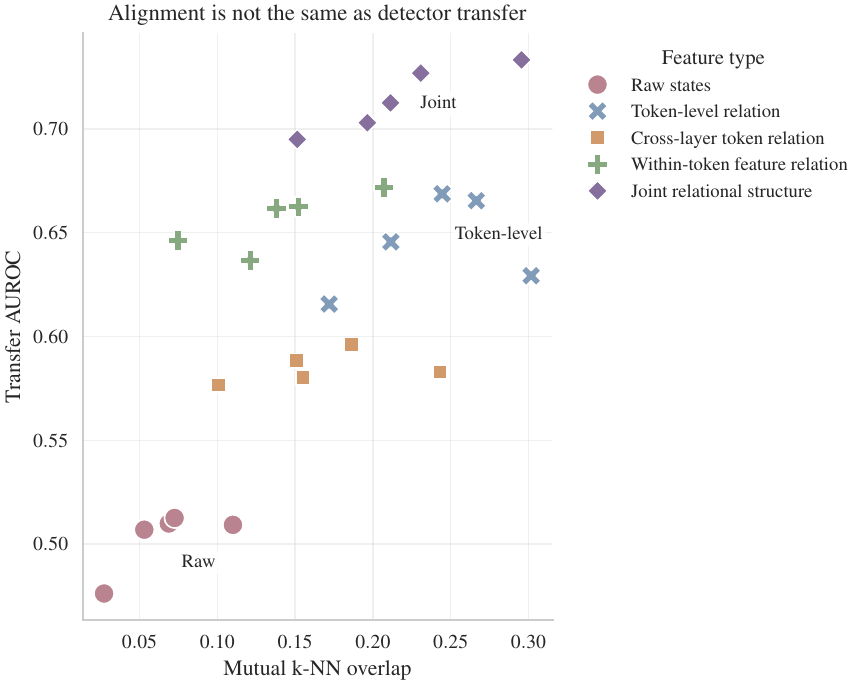}
    \caption{
        Alignment--utility decoupling.
        Each point denotes one relational construction on a representative source-to-target pair. 
        Token-level relations often achieve high mutual $k$NN alignment, whereas joint relational structures yield stronger transfer AUROC. 
        The most stable relational structure is not necessarily the most useful for hallucination detection.
    }
    \label{fig:obs_alignment_utility}
    \vspace{-1em} 
\end{wrapfigure}

\paragraph{Rank diagnostic for alignment--utility mismatch.}
Finally, we quantify whether the feature ordering induced by alignment agrees with the ordering induced by transfer AUROC. 
For each source-target pair $(i,j)$, we rank the five feature views by 
$m_{\mathrm{NN}}^{q}(i,j)$ and by $a_{\mathrm{tr}}^{q}(i,j)$, and compute Spearman's $\rho$ and Kendall's $\tau$ between the two rankings. 
When ties occur, we use standard average ranks. 
Because there are only five feature views, these rank-correlation values are discrete; nevertheless, they provide a compact diagnostic of whether alignment and downstream utility agree. 
Figure~\ref{fig:app_quant_checks}\textcolor{red}{a} shows that the correlations are positive but moderate, indicating that alignment and detector transfer are related but not equivalent.

\paragraph{Observations.}
Figures~\ref{fig:obs_alignment_utility}, \ref{fig:app_alignment_matrices}, \ref{fig:app_detection_matrices}, \ref{fig:app_kde_representative}, and \ref{fig:app_quant_checks}\textcolor{red}{a} support three observations.

First, relational views are more stable across LLMs than raw hidden states. 
In the full mutual $k$NN matrices, raw-state features produce low neighborhood overlap across model pairs, while token-level, cross-layer, within-token, and joint relational views consistently increase local-neighborhood agreement. 
This suggests that relational features preserve cross-model structure that is largely absent from raw hidden coordinates.

Second, stronger alignment does not always imply stronger detector transfer. 
Token-level relations often achieve high mutual $k$NN overlap, but the joint relational structure yields stronger transfer AUROC. 
This mismatch is also visible in the rank-correlation diagnostics: the feature rankings induced by alignment and transfer are positively related, but they are not identical. 
One reason is that mutual $k$NN alignment measures local structural consistency over unlabeled examples, whereas AUROC depends on whether the transferred feature geometry separates factual and hallucinated examples on the target model.

Third, among the fixed hand-designed views considered here, the joint relational structure is the most consistently useful diagnostic for transfer. 
It combines layer, token, and feature-block relations, and therefore captures a richer class of cross-model patterns than any single relation axis. 
The score-distribution plots are consistent with this interpretation: joint relational features produce clearer separation between factual and hallucinated target examples across representative transfer pairs. 
Thus, the fixed handcrafted analysis motivates the learnable relation-aware model in Section~\ref{sec:method}: useful transfer signals are relational, but the most aligned handcrafted view is not necessarily the most discriminative one. 
These diagnostics motivate moving from fixed handcrafted relation graphs to learnable, input-dependent relation aggregation.

\subsection{Handcrafted Relation Priors Help but Saturate}
\label{app:obs_handcrafted_relation_priors}

\begin{figure}[t]
    \centering
    \includegraphics[width=\linewidth]{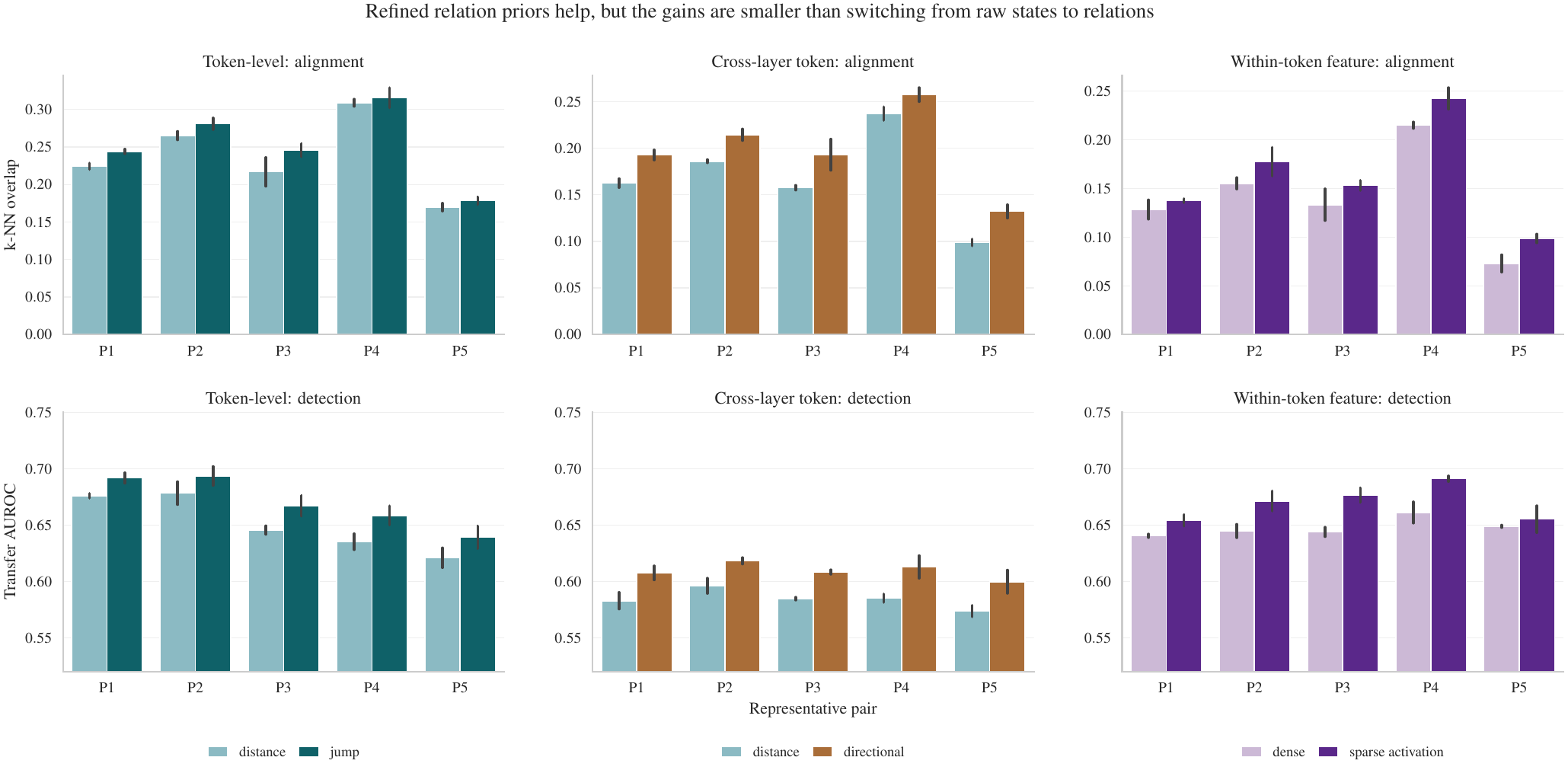}
    \caption{
    Hand-designed relation edges improve transfer, but their gains saturate.
    Simple relation edges are compared with refined handcrafted variants for token, layer, and intra-block relations. Across representative model pairs, refined edges improve both local neighborhood alignment and source-to-target AUROC. The improvements are smaller than the gain from moving from raw hidden states to relational views and are not uniform across relation types. Fixed graphs are useful diagnostics, but their instance-independent structural bias motivates learned relation aggregation. Here P1 to P5 denote Llama-3B$\rightarrow$Llama-8B, Qwen-14B$\rightarrow$Qwen-8B, Llama-8B$\rightarrow$Qwen-8B, Llama-70B$\rightarrow$Qwen-35B-A3B, and Llama-3B$\rightarrow$Ministral-8B.
    }
    \label{fig:obs_edge_refinement}
\end{figure}

Section~\ref{sec:observations} uses deliberately simple hand-designed relations to test whether relational structure provides transferable hallucination signals. 
These simple relations already improve both cross-model alignment and detector transfer, but they leave open a natural question: can stronger handcrafted edge priors remove the need for learned relation aggregation? 
This section studies that question by comparing the simple relations in Section~\ref{sec:observations} with more refined fixed variants. 
The node sets, relation-weighted readout, detector architecture, and evaluation protocol are kept the same as in Section~\ref{app:obs_alignment_transfer_decoupling}; only the edge relation values are changed. 
Thus, any difference in Figure~\ref{fig:obs_edge_refinement} reflects the effect of the handcrafted edge prior itself.

\paragraph{Simple relations used as the reference.}
We consider three relation families from Section~\ref{sec:observations}: token-level relations, cross-layer token relations, and within-token feature relations. 
For each family, the reference edge is the simple relation used in the main observational analysis.

For token-level relations, the simple edge between two selected tokens $t$ and $\tilde{t}$ at a selected diagnostic layer $\ell$ is the token-distance relation
\begin{equation}
\label{app:eq:simple_token_relation_reference}
     R_{i}^{\mathrm{tok}}(t,\tilde{t};\mathbf{Q},\mathbf{A}_i)
    =
    s^{\mathrm{tok}}(t,\tilde{t})
    \big\|
        \bar{\mathbf{h}}_{\boldsymbol{\theta}_i,t}^{\ell}(\mathbf{Q},\mathbf{A}_i)
        -
        \bar{\mathbf{h}}_{\boldsymbol{\theta}_i,\tilde{t}}^{\ell}(\mathbf{Q},\mathbf{A}_i)
    \big\|_2 .
\end{equation}
This relation captures the magnitude of hidden-state separation between two token positions, modulated by their positional distance.

For cross-layer token relations, the simple edge between two selected layers $\ell$ and $\tilde{\ell}$ is the depth-decayed cosine distance between their selected-token summaries:
\begin{equation}
\label{app:eq:simple_layer_relation_reference}
    R_{i}^{\mathrm{lyr}}(\ell,\tilde{\ell};\mathbf{Q},\mathbf{A}_i)
    =
    s^{\mathrm{lyr}}(\ell,\tilde{\ell})
    \big(
        1-
        \cos
        \big(
            \mathrm{vec}(\bar{\mathbf{H}}_{\boldsymbol{\theta}_i}^{\ell,\mathrm{sel}}),
            \mathrm{vec}(\bar{\mathbf{H}}_{\boldsymbol{\theta}_i}^{\tilde{\ell},\mathrm{sel}})
        \big)
    \big),
\end{equation}
where $\bar{\mathbf{H}}_{\boldsymbol{\theta}_i}^{\ell,\mathrm{sel}}$ is defined in Eq.~\eqref{app:eq:padded_layer_matrix}. 
This relation compares the layer summaries themselves, but it does not explicitly encode how the hidden states evolve from one layer to the next.

For within-token feature relations, the simple edge between two feature blocks $b$ and $\tilde{b}$ is the dense Euclidean distance
\begin{equation}
\label{app:eq:simple_feature_relation_reference}
    R_{i}^{\mathrm{feat}}(b,\tilde{b};\mathbf{Q},\mathbf{A}_i)
    =
    \big\|
        \mathrm{vec}(\bar{\mathbf{V}}_{\boldsymbol{\theta}_i,b}(\mathbf{Q},\mathbf{A}_i))
        -
        \mathrm{vec}(\bar{\mathbf{V}}_{\boldsymbol{\theta}_i,\tilde{b}}(\mathbf{Q},\mathbf{A}_i))
    \big\|_2,
\end{equation}
where $\bar{\mathbf{V}}_{\boldsymbol{\theta}_i,b}$ is defined in Eq.~\eqref{app:eq:padded_feat_matrix}. 
This relation uses all coordinates in each block equally.

\paragraph{Refined token relation: state-jump edges.}
The simple token relation in Eq.~\eqref{app:eq:simple_token_relation_reference} compares token hidden states directly. 
A more structured token relation can also incorporate how the hidden state changes as generation moves from one token to the next. 
For a token position $t$, let $t^{-}$ denote its immediate predecessor in the token sequence. 
We define the token-state jump at layer $\ell$ by
\begin{equation}
\label{app:eq:token_jump_vector}
    \Delta_{i,t}^{\ell}(\mathbf{Q},\mathbf{A}_i)
    =
    \bar{\mathbf{h}}_{\boldsymbol{\theta}_i,t}^{\ell}(\mathbf{Q},\mathbf{A}_i)
    -
    \bar{\mathbf{h}}_{\boldsymbol{\theta}_i,t^{-}}^{\ell}(\mathbf{Q},\mathbf{A}_i).
\end{equation}
If the predecessor is not available within the selected sequence, we use the nearest available previous token; this affects only boundary positions. 
The jump-refined token relation is then
\begin{equation}
\label{app:eq:token_jump_relation}
    R_{i}^{\mathrm{tok}\text{-}\mathrm{jump}}(t,\tilde{t};\mathbf{Q},\mathbf{A}_i)
    =
    s^{\mathrm{tok}}(t,\tilde{t})
    \left(
    \big\|
        \bar{\mathbf{h}}_{\boldsymbol{\theta}_i,t}^{\ell}
        -
        \bar{\mathbf{h}}_{\boldsymbol{\theta}_i,\tilde{t}}^{\ell}
    \big\|_2
    +
    \lambda_{\mathrm{jump}}
    \big\|
        \Delta_{i,t}^{\ell}
        -
        \Delta_{i,\tilde{t}}^{\ell}
    \big\|_2
    \right),
\end{equation}
where $\lambda_{\mathrm{jump}}\geq 0$ controls the contribution of the state-jump term. 
The first term is the original token-state distance, while the second term compares local token-to-token transition patterns. 
Thus, this refined edge tests whether transfer improves when the token relation captures not only where token states are located, but also how they change along the generated sequence.

\paragraph{Refined cross-layer relation: directional layer-evolution edges.}
The simple cross-layer relation in Eq.~\eqref{app:eq:simple_layer_relation_reference} compares the layer summaries themselves. 
However, two layers can have similar summaries while participating in different layer-to-layer transformations. 
To test whether layer-evolution directions provide additional transferable signal, we define a directional layer update.

For each selected layer $\ell\in\mathbf{L}^{\mathrm{sel}}$, let $\ell^{+}$ be the next selected layer larger than $\ell$ when it exists, and let $\ell^{-}$ be the previous selected layer otherwise. 
We define
\begin{equation}
\label{app:eq:layer_direction_vector}
    \Delta_{i}^{\ell}(\mathbf{Q},\mathbf{A}_i)
    =
    \begin{cases}
    \mathrm{vec}
    \left(
        \bar{\mathbf{H}}_{\boldsymbol{\theta}_i}^{\ell^{+},\mathrm{sel}}(\mathbf{Q},\mathbf{A}_i)
        -
        \bar{\mathbf{H}}_{\boldsymbol{\theta}_i}^{\ell,\mathrm{sel}}(\mathbf{Q},\mathbf{A}_i)
    \right),
    & \text{if } \ell^{+} \text{ exists},\\[1mm]
    \mathrm{vec}
    \left(
        \bar{\mathbf{H}}_{\boldsymbol{\theta}_i}^{\ell,\mathrm{sel}}(\mathbf{Q},\mathbf{A}_i)
        -
        \bar{\mathbf{H}}_{\boldsymbol{\theta}_i}^{\ell^{-},\mathrm{sel}}(\mathbf{Q},\mathbf{A}_i)
    \right),
    & \text{otherwise}.
    \end{cases}
\end{equation}
This vector summarizes the local direction of representation change around layer $\ell$. 
The directional cross-layer relation is
\begin{equation}
\label{app:eq:layer_direction_relation}
    R_{i}^{\mathrm{lyr}\text{-}\mathrm{dir}}(\ell,\tilde{\ell};\mathbf{Q},\mathbf{A}_i)
    =
    s^{\mathrm{lyr}}(\ell,\tilde{\ell})
    \left[
    \big(
        1-
        \cos
        \big(
            \mathrm{vec}(\bar{\mathbf{H}}_{\boldsymbol{\theta}_i}^{\ell,\mathrm{sel}}),
            \mathrm{vec}(\bar{\mathbf{H}}_{\boldsymbol{\theta}_i}^{\tilde{\ell},\mathrm{sel}})
        \big)
    \big)
    +
    \lambda_{\mathrm{dir}}
    \big(
        1-
        \cos
        \big(
            \Delta_{i}^{\ell},
            \Delta_{i}^{\tilde{\ell}}
        \big)
    \big)
    \right],
\end{equation}
where $\lambda_{\mathrm{dir}}\geq 0$ controls the contribution of the directional term. 
This edge keeps the original layer-summary comparison but additionally measures whether two layers induce similar local transformation directions. 
It therefore tests whether cross-model transfer benefits from handcrafted information about layerwise evolution, beyond static layer-summary distance.

\paragraph{Refined within-token feature relation: sparse-activation edges.}
The simple within-token feature relation in Eq.~\eqref{app:eq:simple_feature_relation_reference} treats all coordinates in each feature block as dense signals. 
However, feature blocks can contain many weakly active or noisy coordinates. 
To test whether emphasizing high-activation coordinates improves transfer, we define a sparse-activation variant.

Let
\[
    \mathbf{v}_{i,b}(\mathbf{Q},\mathbf{A}_i)
    =
    \mathrm{vec}
    \left(
        \bar{\mathbf{V}}_{\boldsymbol{\theta}_i,b}(\mathbf{Q},\mathbf{A}_i)
    \right)
\]
be the dense vector for block $b$. 
Let $\mathbf{m}_{i,b}\in\{0,1\}^{\dim(\mathbf{v}_{i,b})}$ be a binary activation mask that keeps the largest coordinates of 
$|\mathbf{v}_{i,b}|$ and sets the remaining coordinates to zero. 
The sparsified block vector is
\begin{equation}
\label{app:eq:sparse_block_vector}
    \mathrm{sp}(\mathbf{v}_{i,b})
    =
    \mathbf{m}_{i,b}\odot \mathbf{v}_{i,b},
\end{equation}
where $\odot$ denotes element-wise multiplication. 
The sparse-activation relation between two feature blocks is
\begin{equation}
\label{app:eq:sparse_feature_relation}
    R_{i}^{\mathrm{feat}\text{-}\mathrm{sp}}(b,\tilde{b};\mathbf{Q},\mathbf{A}_i)
    =
    \big\|
        \mathrm{sp}(\mathbf{v}_{i,b}(\mathbf{Q},\mathbf{A}_i))
        -
        \mathrm{sp}(\mathbf{v}_{i,\tilde{b}}(\mathbf{Q},\mathbf{A}_i))
    \big\|_2.
\end{equation}
This relation compares feature blocks after suppressing weak activations, and therefore tests whether transfer improves when the handcrafted relation emphasizes more active coordinates.

\paragraph{Evaluation protocol.}
For each refined relation, we keep the same node set and readout as its simple counterpart. 
Specifically, token-distance and token-jump relations are both evaluated on the token-node graph; dense and sparse-activation feature relations are both evaluated on the feature-block graph; layer-summary and directional layer-evolution relations are both evaluated on the layer-node graph. 
After computing the corresponding graph-level features with Eq.~\eqref{app:eq:general_relation_readout}, we evaluate the two criteria from Section~\ref{app:obs_alignment_transfer_decoupling}: mutual $k$NN alignment and directional cross-model AUROC.

Figure~\ref{fig:obs_edge_refinement} reports results on five representative source-to-target transfer pairs, denoted P1--P5. 
These pairs cover same-family transfer, cross-family transfer, scale-matched transfer, and scale-mismatched transfer. 
For each relation family, the lighter bar denotes the simple reference relation and the darker bar denotes the refined handcrafted relation.

\paragraph{Observations.}
Figure~\ref{fig:obs_edge_refinement} shows that refined handcrafted priors generally improve both local-neighborhood alignment and transfer AUROC. 
Token-state jump edges improve over token-distance edges, suggesting that local state-transition patterns contain additional transferable information beyond static token-state separation. 
Directional layer-evolution edges improve over static layer-summary edges, indicating that layer-to-layer transformation directions provide useful complementary information. 
Sparse-activation feature edges improve over dense feature-block distances in several settings, suggesting that suppressing weak activations can reduce noise in feature-block relations.

However, the gains from these refined handcrafted priors are modest compared with the larger improvement obtained by moving from raw hidden states to relational views in the first place. 
The improvements are also not uniform across relation families and transfer pairs. 
For example, the refined token relation and refined layer relation do not yield the same magnitude of gain, and the feature-block refinement behaves differently from the token and layer refinements. 
This indicates that better handcrafted priors can help, but choosing a single fixed relation formula is unlikely to fully capture the relation structure needed for robust cross-model hallucination detection.

\paragraph{Implication.}
These results suggest that fixed relation priors are useful diagnostics but have limited headroom. 
More sophisticated handcrafted edges can encode additional intuitions, such as token-state transitions, layerwise evolution directions, or sparse activation patterns, but their benefits saturate and remain relation-type dependent. 
This motivates the learnable relation-aware aggregation in Section~\ref{sec:method}. 
Rather than manually selecting one refined edge formula for all examples and all model pairs, \method{} learns how to combine positional and content-based relation signals in an input-dependent way.

\subsection{Relation-Channel Importance Is Input-Dependent}
\label{app:obs_channel_importance}

\begin{figure}[t]
    \centering
    \includegraphics[width=\linewidth]{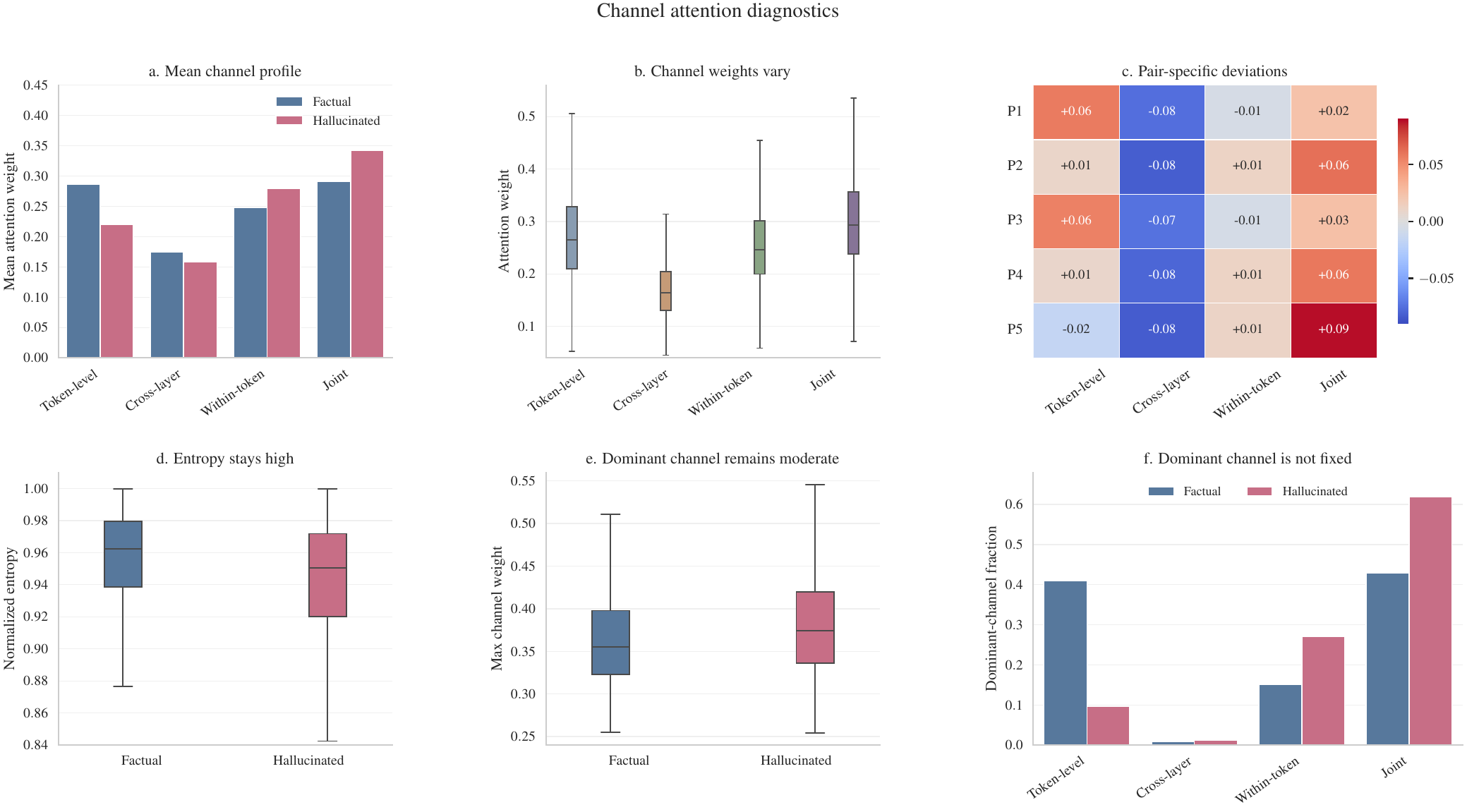}
    \caption{
    Channel attention diagnostics.
    A lightweight channel probe is trained on source-model relational features and evaluated on target-model examples, with all relational features frozen. The four channel weights sum to one for each example. Panels (A) and (B) show label-conditioned mean profiles and per-example channel-weight variation. Panel (C) shows pair-specific deviations after subtracting each pair's mean channel weight, so colors emphasize relative channel preference within each transfer pair. Panels (D) and (E) report normalized entropy and maximum channel weight on separate axes. Panel (F) shows the dominant channel frequency. Together, these diagnostics indicate that the useful relation mixture is not well described by a single fixed channel weighting.
    }
    \label{fig:app_channel_attention}
\end{figure}

Sections~\ref{app:obs_alignment_transfer_decoupling} and~\ref{app:obs_handcrafted_relation_priors} show that different relational views capture different transferable structures. 
However, these results do not imply that a single relation view should be used for every example. 
A hallucinated answer may differ from a factual answer through token-level inconsistency, abnormal layer evolution, feature-block changes, or a combination of these effects. 
Therefore, we further ask whether the importance of different relation channels is fixed, or whether it changes across examples, labels, and source-target model pairs.

\paragraph{Diagnostic channel-gated probe.}
We use the four fixed relational features from Section~\ref{app:obs_alignment_transfer_decoupling}:
\[
    \mathbf{z}_{i}^{\mathrm{tok}}(\mathbf{Q},\mathbf{A}_i),\quad
    \mathbf{z}_{i}^{\mathrm{lyr}}(\mathbf{Q},\mathbf{A}_i),\quad
    \mathbf{z}_{i}^{\mathrm{feat}}(\mathbf{Q},\mathbf{A}_i),\quad
    \mathbf{z}_{i}^{\mathrm{joint}}(\mathbf{Q},\mathbf{A}_i).
\]
These four views are treated as four relation channels. 
Let 
\[
    \mathcal{C}=\{\mathrm{tok},\mathrm{lyr},\mathrm{feat},\mathrm{joint}\}
\]
denote the set of channels. 
Since the four relational features can have different dimensions, we first map each channel into the same diagnostic dimension:
\begin{equation}
\label{app:eq:channel_projection}
    \mathbf{r}_{i}^{q}(\mathbf{Q},\mathbf{A}_i)
    =
    g_{q}\!\left(
        \mathbf{z}_{i}^{q}(\mathbf{Q},\mathbf{A}_i)
    \right)
    \in\mathbb{R}^{d_{\mathrm{c}}},
    \qquad q\in\mathcal{C},
\end{equation}
where $g_q(\cdot)$ is a small learnable projection for channel $q$. 
The projected channel vectors are then assigned input-dependent channel weights. 
For each channel $q$, we compute a channel score
\begin{equation}
\label{app:eq:channel_score}
    a_{i}^{q}(\mathbf{Q},\mathbf{A}_i)
    =
    \mathbf{w}_{\mathrm{c}}^{\top}
    \tanh
    \left(
        \mathbf{W}_{\mathrm{c}}
        \mathbf{r}_{i}^{q}(\mathbf{Q},\mathbf{A}_i)
        +
        \mathbf{b}_{\mathrm{c}}
    \right),
\end{equation}
and normalize the scores over the four channels:
\begin{equation}
\label{app:eq:channel_weight}
    \alpha_{i}^{q}(\mathbf{Q},\mathbf{A}_i)
    =
    \frac{
        \exp(a_{i}^{q}(\mathbf{Q},\mathbf{A}_i))
    }{
        \sum_{q'\in\mathcal{C}}
        \exp(a_{i}^{q'}(\mathbf{Q},\mathbf{A}_i))
    },
    \qquad q\in\mathcal{C}.
\end{equation}
The fused diagnostic representation is
\begin{equation}
\label{app:eq:channel_fusion}
    \mathbf{z}_{i}^{\mathrm{mix}}(\mathbf{Q},\mathbf{A}_i)
    =
    \sum_{q\in\mathcal{C}}
    \alpha_{i}^{q}(\mathbf{Q},\mathbf{A}_i)
    \mathbf{r}_{i}^{q}(\mathbf{Q},\mathbf{A}_i).
\end{equation}
A binary MLP detector is then trained on 
$\mathbf{z}_{i}^{\mathrm{mix}}(\mathbf{Q},\mathbf{A}_i)$ 
using the same source-to-target transfer protocol as in Section~\ref{app:obs_alignment_transfer_decoupling}. 
For a source model $\mathbb{P}_{\boldsymbol{\theta}_i}$ and a target model $\mathbb{P}_{\boldsymbol{\theta}_j}$, the detector and channel-gating module are trained on $\mathcal{D}_i$ and evaluated on $\mathcal{D}_j$. 
After training, we record the channel weights 
$\alpha_{j}^{q}(\mathbf{Q}^{s},\mathbf{A}_{j}^{s})$ 
on target examples.

This channel-gated probe is used only for analysis. 
The weights $\alpha_i^q$ should not be interpreted as causal explanations. 
They are diagnostic quantities that describe how the probe allocates mass across the fixed relational channels when forming its prediction.

\paragraph{Label-conditioned channel profiles.}
We first examine whether factual and hallucinated examples use the same channel mixture. 
Recall that $y_j^s=1$ indicates that the generated answer is truthful, and $y_j^s=0$ indicates hallucination. 
For each label $y\in\{0,1\}$ and channel $q\in\mathcal{C}$, we compute the label-conditioned mean channel weight on the target model:
\begin{equation}
\label{app:eq:label_conditioned_channel_profile}
    \bar{\alpha}_{j,y}^{q}
    =
    \frac{1}{|\mathcal{S}_{j,y}|}
    \sum_{s\in\mathcal{S}_{j,y}}
    \alpha_{j}^{q}(\mathbf{Q}^{s},\mathbf{A}_{j}^{s}),
    \qquad
    \mathcal{S}_{j,y}
    =
    \{s:y_j^s=y\}.
\end{equation}
If relation-channel importance were fixed, the profiles 
$\{\bar{\alpha}_{j,0}^{q}\}_{q\in\mathcal{C}}$ and 
$\{\bar{\alpha}_{j,1}^{q}\}_{q\in\mathcal{C}}$ 
would be nearly identical. 
Differences between the two profiles indicate that the detector uses relational information differently for hallucinated and factual examples.

\paragraph{Per-example variation.}
Average profiles alone may hide substantial example-level variation. 
For each target example, we therefore analyze the full channel-weight vector
\[
    \boldsymbol{\alpha}_{j}^{s}
    =
    \left[
    \alpha_{j}^{\mathrm{tok}}(\mathbf{Q}^{s},\mathbf{A}_{j}^{s}),
    \alpha_{j}^{\mathrm{lyr}}(\mathbf{Q}^{s},\mathbf{A}_{j}^{s}),
    \alpha_{j}^{\mathrm{feat}}(\mathbf{Q}^{s},\mathbf{A}_{j}^{s}),
    \alpha_{j}^{\mathrm{joint}}(\mathbf{Q}^{s},\mathbf{A}_{j}^{s})
    \right].
\]
We report the distribution of these weights across examples. 
If the probe used a fixed relation view, then almost all examples would place their largest weight on the same channel and the per-example weight distribution would be nearly constant. 
In contrast, broad distributions indicate that the relevant relational signal varies across inputs.

\paragraph{Pair-specific channel deviations.}
The same relation channel may also have different utility for different source-target model pairs. 
For each transfer pair $(i,j)$, we compute the average target-side channel profile
\begin{equation}
\label{app:eq:pair_channel_profile}
    \bar{\alpha}_{i\rightarrow j}^{q}
    =
    \frac{1}{n}
    \sum_{s=1}^{n}
    \alpha_{j}^{q}(\mathbf{Q}^{s},\mathbf{A}_{j}^{s}),
    \qquad q\in\mathcal{C}.
\end{equation}
We then center each profile by the uniform channel weight:
\begin{equation}
\label{app:eq:pair_channel_deviation}
    \delta_{i\rightarrow j}^{q}
    =
    \bar{\alpha}_{i\rightarrow j}^{q}
    -
    \frac{1}{|\mathcal{C}|}
    \sum_{q'\in\mathcal{C}}
    \bar{\alpha}_{i\rightarrow j}^{q'}
    =
    \bar{\alpha}_{i\rightarrow j}^{q}-\frac{1}{|\mathcal{C}|},
\end{equation}
where the second equality holds because the channel weights of a transfer pair sum to one. 
A positive $\delta_{i\rightarrow j}^{q}$ means that the pair allocates above-uniform weight to channel $q$, and differences between rows show how channel preferences vary across transfer pairs. 
This diagnostic is useful because different LLM families may differ in their layer dynamics, token-level geometry, or feature-block organization.

\paragraph{Entropy, maximum weight, and dominant channel.}
We further summarize the sharpness of the channel distribution. 
For each example, the normalized channel entropy is
\begin{equation}
\label{app:eq:channel_entropy}
    H_{j}^{s}
    =
    -
    \frac{1}{\log|\mathcal{C}|}
    \sum_{q\in\mathcal{C}}
    \alpha_{j}^{q}(\mathbf{Q}^{s},\mathbf{A}_{j}^{s})
    \log
    \left(
        \alpha_{j}^{q}(\mathbf{Q}^{s},\mathbf{A}_{j}^{s})
        +
        \epsilon
    \right),
\end{equation}
where $\epsilon>0$ is a small numerical-stability constant. 
A value close to zero would indicate that almost all weight is assigned to a single channel, while a value close to one would indicate a nearly uniform mixture. 
We also compute the maximum channel weight
\begin{equation}
\label{app:eq:max_channel_weight}
    M_{j}^{s}
    =
    \max_{q\in\mathcal{C}}
    \alpha_{j}^{q}(\mathbf{Q}^{s},\mathbf{A}_{j}^{s}),
\end{equation}
and the dominant channel
\begin{equation}
\label{app:eq:dominant_channel}
    q_{j}^{s}
    =
    \arg\max_{q\in\mathcal{C}}
    \alpha_{j}^{q}(\mathbf{Q}^{s},\mathbf{A}_{j}^{s}).
\end{equation}
The empirical frequency of $q_j^s$ over examples indicates whether one relation channel dominates the diagnostic probe or whether different channels are selected for different inputs.

\paragraph{Observations.}
Figure~\ref{fig:app_channel_attention} supports four observations.

First, channel importance differs between factual and hallucinated examples. 
The label-conditioned profiles are not identical: hallucinated and factual examples allocate different average weights to the token, layer, feature, and joint channels. 
This suggests that hallucination-related signals are not confined to a single relational axis. 
Instead, the detector uses different mixtures of relation channels depending on the label-conditioned structure of the example.

Second, the channel weights vary substantially across individual examples. 
The per-example distributions show that the probe does not assign a constant weight vector to all inputs. 
Some examples emphasize token-level relations, while others place more mass on feature-block, layer, or joint relations. 
This variation is consistent with the view that hallucinations can manifest through different internal patterns depending on the question, generated answer, and target model.

Third, channel usage is source-target dependent. 
The pair-specific deviations show that different transfer pairs shift weight toward different channels. 
This is expected because LLMs with different architectures, scales, or training procedures need not share the same token geometry, layer evolution, or feature-block organization. 
Thus, a fixed channel mixture that works well for one source-target pair may be suboptimal for another.

Fourth, the probe does not collapse to a single relation channel. 
The normalized entropy remains away from zero, the maximum channel weight is not consistently close to one, and the dominant-channel frequency is distributed across multiple channels. 
This indicates that the diagnostic probe combines relation channels rather than using only one of them. 
At the same time, the entropy is not exactly one, so the mixture is not purely uniform. 
The probe therefore learns nontrivial, input-dependent channel preferences.

\paragraph{Implication.}
These diagnostics support the need for input-dependent relation aggregation. 
The earlier sections show that relational views are more transferable than raw hidden states, and that refined handcrafted priors can help but saturate. 
The channel-weight analysis adds a complementary point: even among useful relation views, the relevant channel is not fixed across labels, examples, or source-target pairs. 
Therefore, a detector that commits to one handcrafted relation view, or to one fixed mixture of relation channels, is unlikely to be optimal across heterogeneous LLMs. 
This motivates the relation-aware attention mechanism in Section~\ref{sec:method}, where positional and content-based relation signals are combined adaptively for each input rather than selected by a fixed manual rule.

\subsection{Additional Robustness and Transfer Diagnostics}
\label{app:obs_robustness_transfer_diagnostics}

We provide two additional diagnostics to examine whether the transfer advantage of relational views depends on a particular source-target scale direction or on detector training randomness. 
Both diagnostics use the same five feature views as Section~\ref{app:obs_alignment_transfer_decoupling}: raw hidden states, token-level relations, cross-layer token relations, within-token feature relations, and joint relational structures. 
For a source model $\mathbb{P}_{\boldsymbol{\theta}_i}$ and a target model $\mathbb{P}_{\boldsymbol{\theta}_j}$, we denote the cross-model transfer AUROC of feature view $q$ by $a_{\mathrm{tr}}^{q}(i,j)$.

\paragraph{Scale direction.}
Since cross-model transfer is directional, training on a larger source model and testing on a smaller target model may not behave the same as the reverse direction. 
Let $\mathrm{size}(i)$ denote the parameter scale of $\mathbb{P}_{\boldsymbol{\theta}_i}$. 
We compare two sets of directed transfer pairs:
\[
    \mathcal{P}_{\mathrm{large}\rightarrow\mathrm{small}}
    =
    \{(i,j):\mathrm{size}(i)>\mathrm{size}(j)\},
    \qquad
    \mathcal{P}_{\mathrm{small}\rightarrow\mathrm{large}}
    =
    \{(i,j):\mathrm{size}(i)<\mathrm{size}(j)\}.
\]
For each feature view $q$, we compute the average AUROC in both directions and report their difference:
\[
    \Delta_{\mathrm{scale}}^{q}
    =
    \frac{1}{|\mathcal{P}_{\mathrm{large}\rightarrow\mathrm{small}}|}
    \sum_{(i,j)\in\mathcal{P}_{\mathrm{large}\rightarrow\mathrm{small}}}
    a_{\mathrm{tr}}^{q}(i,j)
    -
    \frac{1}{|\mathcal{P}_{\mathrm{small}\rightarrow\mathrm{large}}|}
    \sum_{(i,j)\in\mathcal{P}_{\mathrm{small}\rightarrow\mathrm{large}}}
    a_{\mathrm{tr}}^{q}(i,j).
\]
Figure~\ref{fig:app_quant_checks}\textcolor{red}{b} shows that large-source to small-target transfer is slightly stronger than small-source to large-target transfer for most feature views. 
This is consistent with the intuition that larger source models may expose richer internal structures for training the detector. 
However, the gap is modest compared with the difference between raw hidden states and relational views. 
Thus, the main observation is not driven by a single favorable scale direction: relational features improve transfer in both directions, while scale direction only mildly modulates the absolute AUROC.

\paragraph{Seed stability.}
We also examine whether the transfer results are sensitive to the random seed used for training the downstream MLP detector. 
For each directed pair $(i,j)$ and feature view $q$, we train the detector with multiple random seeds and denote the resulting AUROC by $a_{\mathrm{tr}}^{q,r}(i,j)$, where $r$ indexes the seed. 
The seed standard deviation is
\[
    \sigma_{\mathrm{seed}}^{q}(i,j)
    =
    \left(
    \frac{1}{R-1}
    \sum_{r=1}^{R}
    \left(
        a_{\mathrm{tr}}^{q,r}(i,j)
        -
        \frac{1}{R}\sum_{r'=1}^{R}a_{\mathrm{tr}}^{q,r'}(i,j)
    \right)^2
    \right)^{1/2}.
\]
We then average $\sigma_{\mathrm{seed}}^{q}(i,j)$ over all evaluated directed transfer pairs. 
Figure~\ref{fig:app_quant_checks}\textcolor{red}{c} shows that raw hidden states have noticeably higher seed variance, whereas relational views yield more stable detector performance. 
This suggests that raw-state transfer often depends on unstable decision boundaries, while relational features provide a more consistent basis for cross-model training.

\paragraph{Implication.}
Together, these diagnostics support the robustness of the observational findings. 
The advantage of relational views is not explained by choosing an easy source-target scale direction, and it is not a consequence of lucky detector initialization. 
Large-to-small transfer is mildly easier than small-to-large transfer, but the effect is secondary. 
Relational views are both more transferable and more stable than raw hidden states, which further motivates using relational structure as the basis for cross-model hallucination detection.

\section{Additional Experimental Analysis}
\label{app:additional_exp_analysis}

\paragraph{Expanded analysis of the main cross-model results.}
Table~\ref{tab:main_cross_family_results} evaluates hallucination detectors under the leave-one-family-out cross-model setting, where training-based methods are trained on source model families and evaluated on a held-out target family. 
This setting directly tests whether a detector can transfer beyond the representation space and generation behavior of the models used for training. 
Overall, \method{} achieves the best AUROC across all target models and all seven benchmarks. 
These results show that the proposed relational graph discrimination framework remains effective across factual question answering, conversational question answering, mathematical reasoning, coding, and agent-oriented tasks.

The comparison with training-free methods reveals a clear pattern. 
Methods such as Perplexity, Semantic Entropy, EigenScore, Lexical Similarity, Verbalize, and InterrogateLLM do not train source-model-specific classifiers, and therefore largely avoid training-induced cross-model generalization issues. 
Among them, InterrogateLLM and Semantic Entropy are usually the strongest baselines, suggesting that consistency and uncertainty signals are useful model-agnostic indicators. 
Nevertheless, these methods rely on indirect output-level evidence, such as likelihood, semantic variability, lexical overlap, or self-evaluation. 
As a result, their performance remains below \method{} on all benchmarks in the averaged results. 
This indicates that output-level signals are useful but do not fully capture hallucination-relevant structure in the model's internal computation.

The representative training-based baselines show a different failure mode. 
MM, SEP, SAPLMA, and LapEigvals use supervised or structured signals for hallucination detection, but their cross-model performance is often close to random guessing. 
This does not imply that these methods are ineffective in easier within-model settings. 
Rather, the results suggest that detectors learned from source models can rely on regularities that do not transfer to unseen target models. 
For hidden-state-based probes, this is consistent with the cross-model semantic inconsistency discussed in Section~\ref{sec:intro}. 
For attention-based spectral statistics such as LapEigvals, the results suggest that fixed structural descriptors alone are not sufficient to provide robust cross-family discrimination. 
The contrast between training-free and training-based baselines highlights the central challenge of CMHD. 
Training-free methods are easier to apply across models but have limited discriminative strength, whereas supervised detectors can be discriminative but often become brittle under cross-model transfer. 
\method{} bridges this gap by learning from labeled source-model data while representing hidden states through relational graph structures that reduce reliance on model-specific raw features.

\paragraph{Dataset-level observations.}
The largest gains of \method{} appear on xLAM and GSM8K. 
On xLAM, \method{} reaches an average AUROC of $0.8564$, clearly exceeding the strongest baseline. 
This suggests that relational hidden-state structures are particularly informative for agent-oriented hallucination detection. 
In such settings, errors may involve tool selection, action arguments, or intermediate decision trajectories, which can induce structured deviations across tokens, layers, and feature groups. 
On GSM8K, \method{} also achieves strong performance, with an average AUROC of $0.8298$. 
This is consistent with the intuition that reasoning errors are not always isolated local mistakes, but may appear as structured changes along the internal reasoning trajectory.

The gains are also stable on standard QA and hallucination benchmarks. 
On TruthfulQA, HaluEval, TriviaQA, and CoQA, \method{} consistently outperforms both training-free and training-based baselines. 
This shows that the advantage of relational graph discrimination is not limited to specialized reasoning or agent tasks, but also applies to conventional factual and conversational question answering. 
Among these datasets, TriviaQA and CoQA are relatively favorable to lexical or consistency-based baselines, since answer correctness can sometimes be correlated with surface overlap or answer agreement. 
Even in these settings, \method{} maintains a clear advantage, suggesting that internal relational structures provide complementary information beyond output-level similarity.

MBPP is the most challenging benchmark for \method{}. 
Although \method{} still achieves the best average AUROC on MBPP, its margin over the strongest training-free baseline is smaller than on other datasets. 
This is expected for two reasons. 
First, MBPP contains fewer than one thousand programming problems, so the available supervision remains limited even after collecting generations from multiple LLMs. 
Second, code correctness is strongly execution-dependent. 
A generated program can be semantically wrong for reasons that are not fully reflected in natural-language hidden-state relations. 
Therefore, the smaller margin on MBPP should be interpreted as a limitation of the current representation signal rather than a contradiction of the overall trend.

\paragraph{Model-family and scale observations.}
The results also show that larger target models often yield stronger detection performance, although the trend is not strictly monotonic. 
Within the Llama family, Llama-3.3-70B generally achieves higher AUROC than smaller Llama models under \method{}. 
A similar pattern can be observed for larger Qwen and Ministral models in many settings. 
This may be because stronger LLMs develop richer and more structured internal representations, which provide more informative relational signals for hallucination detection. 
At the same time, local reversals still occur across datasets and model families. 
This is reasonable because model version, architecture, instruction tuning, and generation behavior can also affect hidden-state relations. 
Thus, the results support a cautious conclusion. 
Model scale is often helpful, but cross-model hallucination detection remains influenced by family-specific and version-specific factors.

\paragraph{Expanded analysis of ablation results.}
Table~\ref{tab:ablation_full} studies the two core components of \method{}, namely relation-aware attention and meta-learning. 
The main ablation table reports a compact view of the results. 
QA Avg. denotes the macro average over TruthfulQA, HaluEval, TriviaQA, and CoQA. 
GSM8K, MBPP, and xLAM are kept separately to represent reasoning, code, and agent-oriented settings, and All Avg. denotes the macro average over all seven benchmarks. 

\definecolor{SkyBlueRow}{RGB}{224,244,250}
\begin{table*}[t]
\caption{Full ablation results under the cross-family evaluation protocol. Values are AUROC. Best results are \textbf{bold} and second-best results are \underline{underlined}.}
\vspace{2mm}
\centering
\label{tab:ablation_full}
\resizebox{\linewidth}{!}{%
\begin{tabular}{lcccccccccccccccc}
\toprule
\multirow{4}{*}{\textbf{Method}} & \multicolumn{16}{c}{\textbf{Target Model}} \\
& \multicolumn{8}{c}{Llama3.2-3B} & \multicolumn{8}{c}{Llama3.1-8B} \\
\cmidrule(lr){2-9}\cmidrule(lr){10-17}
& \multicolumn{8}{c}{Mean AUROC$\uparrow$} & \multicolumn{8}{c}{Mean AUROC$\uparrow$} \\
& TruthfulQA & HaluEval & TriviaQA & CoQA & GSM8K & MBPP & xLAM & \textbf{Average} & TruthfulQA & HaluEval & TriviaQA & CoQA & GSM8K & MBPP & xLAM & \textbf{Average} \\
\midrule
\rowcolor{gray!18} \multicolumn{17}{l}{\textbf{Training-free methods}} \\
Perplexity & 0.6329 & 0.6464 & 0.6448 & 0.6296 & 0.6536 & 0.6188 & 0.6621 & 0.6412 & 0.6546 & 0.6452 & 0.6545 & 0.6395 & 0.6523 & 0.6024 & 0.6857 & 0.6477 \\
Semantic Entropy & 0.6862 & 0.7095 & 0.7006 & \underline{0.6934} & 0.7112 & 0.6571 & 0.7223 & 0.6972 & 0.6991 & 0.7113 & 0.7065 & \underline{0.6919} & 0.7258 & 0.6636 & 0.7438 & 0.7060 \\
EigenScore & 0.6636 & 0.6872 & 0.6767 & 0.6719 & 0.6772 & 0.6508 & 0.7045 & 0.6760 & 0.6985 & 0.6963 & 0.6803 & 0.6747 & 0.7068 & 0.6614 & 0.6957 & 0.6877 \\
Lexical Similarity & 0.6542 & 0.6747 & 0.6878 & 0.6786 & 0.6454 & 0.6225 & 0.6603 & 0.6605 & 0.6760 & 0.6681 & 0.6928 & 0.6628 & 0.6495 & 0.6236 & 0.6871 & 0.6657 \\
Verbalize & 0.5199 & 0.5306 & 0.5220 & 0.5290 & 0.5207 & 0.5069 & 0.5470 & 0.5252 & 0.5238 & 0.5398 & 0.5110 & 0.5050 & 0.5474 & 0.4884 & 0.5595 & 0.5250 \\
InterrogateLLM & 0.6845 & 0.7353 & 0.7012 & 0.6915 & 0.7177 & \underline{0.6915} & 0.7253 & 0.7067 & 0.7185 & 0.7211 & 0.7201 & 0.6881 & 0.7293 & \underline{0.6865} & 0.7469 & 0.7158 \\
\rowcolor{gray!18} \multicolumn{17}{l}{\textbf{Training-based methods}} \\
MM & 0.4857 & 0.5170 & 0.5129 & 0.4874 & 0.5061 & 0.5035 & 0.5200 & 0.5047 & 0.5127 & 0.5029 & 0.5247 & 0.5000 & 0.5141 & 0.4808 & 0.5052 & 0.5058 \\
SEP & 0.5079 & 0.5193 & 0.5373 & 0.5189 & 0.5300 & 0.5032 & 0.5290 & 0.5208 & 0.5191 & 0.5346 & 0.5142 & 0.5089 & 0.5382 & 0.4876 & 0.5395 & 0.5203 \\
SAPLMA & 0.4941 & 0.5191 & 0.5239 & 0.5390 & 0.5158 & 0.4901 & 0.5397 & 0.5174 & 0.5326 & 0.5311 & 0.5250 & 0.5013 & 0.5382 & 0.4818 & 0.5386 & 0.5212 \\
LapEigvals & 0.4817 & 0.5021 & 0.4740 & 0.4965 & 0.5264 & 0.4765 & 0.5295 & 0.4981 & 0.4981 & 0.4916 & 0.4712 & 0.4789 & 0.5083 & 0.4725 & 0.5225 & 0.4919 \\
\rowcolor{gray!18} \multicolumn{17}{l}{\textbf{Ablation variants}} \\
CHORD w/o RelAttn & 0.6511 & 0.7258 & 0.7033 & 0.6800 & 0.7241 & 0.6458 & 0.7259 & 0.6937 & 0.7154 & 0.7230 & 0.7115 & 0.6594 & 0.7403 & 0.6678 & 0.7916 & 0.7156 \\
CHORD w/o Meta & \underline{0.6883} & \underline{0.7457} & \underline{0.7194} & 0.6872 & \underline{0.7394} & 0.6563 & \underline{0.7594} & \underline{0.7137} & \underline{0.7339} & \underline{0.7696} & \underline{0.7442} & 0.6637 & \underline{0.7538} & 0.6693 & \underline{0.7975} & \underline{0.7331} \\
\rowcolor{gray!18} \multicolumn{17}{l}{\textbf{Proposed method}} \\
\rowcolor{SkyBlueRow} \textbf{CHORD} & \textbf{0.7766} & \textbf{0.7926} & \textbf{0.7991} & \textbf{0.7920} & \textbf{0.8193} & \textbf{0.7351} & \textbf{0.8437} & \textbf{0.7941} & \textbf{0.7989} & \textbf{0.8102} & \textbf{0.8156} & \textbf{0.7823} & \textbf{0.8362} & \textbf{0.7307} & \textbf{0.8650} & \textbf{0.8056} \\
\midrule[1.5pt]
\multicolumn{17}{c}{} \\[-2mm]
& \multicolumn{8}{c}{Llama3.3-70B} & \multicolumn{8}{c}{Qwen3-8B} \\
\cmidrule(lr){2-9}\cmidrule(lr){10-17}
\midrule
\rowcolor{gray!18} \multicolumn{17}{l}{\textbf{Training-free methods}} \\
Perplexity & 0.6575 & 0.6747 & 0.6717 & 0.6748 & 0.6765 & 0.6353 & 0.6931 & 0.6691 & 0.6345 & 0.6398 & 0.6781 & 0.6542 & 0.6653 & 0.6179 & 0.6726 & 0.6518 \\
Semantic Entropy & 0.7096 & 0.7366 & 0.7231 & \underline{0.7272} & 0.7410 & 0.6697 & 0.7406 & 0.7211 & 0.6840 & 0.6800 & 0.7127 & 0.6994 & 0.7143 & 0.6502 & 0.7228 & 0.6948 \\
EigenScore & 0.7076 & 0.7087 & 0.6961 & 0.7095 & 0.7094 & 0.6703 & 0.7093 & 0.7016 & 0.6785 & 0.6817 & 0.6869 & 0.6873 & 0.6813 & 0.6683 & 0.7085 & 0.6846 \\
Lexical Similarity & 0.6844 & 0.6797 & 0.7040 & 0.7088 & 0.6854 & 0.6432 & 0.6858 & 0.6845 & 0.6538 & 0.6538 & 0.7089 & 0.6966 & 0.6441 & 0.6401 & 0.6688 & 0.6666 \\
Verbalize & 0.5472 & 0.5531 & 0.5472 & 0.5385 & 0.5592 & 0.5139 & 0.5584 & 0.5454 & 0.5193 & 0.5090 & 0.5200 & 0.5170 & 0.5339 & 0.4969 & 0.5463 & 0.5203 \\
InterrogateLLM & 0.7343 & 0.7454 & 0.7472 & 0.7219 & 0.7473 & \underline{0.7181} & 0.7477 & 0.7374 & \underline{0.6950} & \underline{0.7077} & 0.7044 & 0.7173 & 0.7082 & \underline{0.6834} & 0.7483 & 0.7092 \\
\rowcolor{gray!18} \multicolumn{17}{l}{\textbf{Training-based methods}} \\
MM & 0.5095 & 0.5244 & 0.5288 & 0.5195 & 0.5259 & 0.4814 & 0.5153 & 0.5150 & 0.4923 & 0.4849 & 0.5044 & 0.5095 & 0.5046 & 0.4837 & 0.5171 & 0.4995 \\
SEP & 0.5117 & 0.5352 & 0.5400 & 0.5243 & 0.5263 & 0.4905 & 0.5297 & 0.5225 & 0.5147 & 0.5030 & 0.4993 & 0.5062 & 0.5159 & 0.4823 & 0.5228 & 0.5063 \\
SAPLMA & 0.5250 & 0.5177 & 0.5231 & 0.5364 & 0.5375 & 0.4840 & 0.5386 & 0.5232 & 0.4822 & 0.5223 & 0.5171 & 0.5093 & 0.5167 & 0.4922 & 0.5384 & 0.5112 \\
LapEigvals & 0.5083 & 0.4992 & 0.5167 & 0.5204 & 0.5132 & 0.4718 & 0.5280 & 0.5082 & 0.4831 & 0.4872 & 0.5141 & 0.4873 & 0.5063 & 0.4716 & 0.5292 & 0.4970 \\
\rowcolor{gray!18} \multicolumn{17}{l}{\textbf{Ablation variants}} \\
CHORD w/o RelAttn & 0.7420 & 0.7461 & 0.7515 & 0.7097 & 0.7646 & 0.7017 & 0.7821 & 0.7425 & 0.6634 & 0.6950 & 0.7147 & 0.7130 & 0.7233 & 0.6416 & 0.7602 & 0.7016 \\
CHORD w/o Meta & \underline{0.7552} & \underline{0.7652} & \underline{0.7860} & 0.7220 & \underline{0.8101} & 0.7151 & \underline{0.7886} & \underline{0.7632} & 0.6627 & 0.7019 & \underline{0.7531} & \underline{0.7256} & \underline{0.7444} & 0.6577 & \underline{0.7706} & \underline{0.7166} \\
\rowcolor{gray!18} \multicolumn{17}{l}{\textbf{Proposed method}} \\
\rowcolor{SkyBlueRow} \textbf{CHORD} & \textbf{0.8276} & \textbf{0.8374} & \textbf{0.8257} & \textbf{0.8302} & \textbf{0.8568} & \textbf{0.7605} & \textbf{0.8713} & \textbf{0.8299} & \textbf{0.7896} & \textbf{0.7941} & \textbf{0.8192} & \textbf{0.7991} & \textbf{0.8256} & \textbf{0.7481} & \textbf{0.8484} & \textbf{0.8034} \\
\midrule[1.5pt]
\multicolumn{17}{c}{} \\[-2mm]
& \multicolumn{8}{c}{Qwen3-14B} & \multicolumn{8}{c}{Qwen3.5-35B} \\
\cmidrule(lr){2-9}\cmidrule(lr){10-17}
\midrule
\rowcolor{gray!18} \multicolumn{17}{l}{\textbf{Training-free methods}} \\
Perplexity & 0.6408 & 0.6650 & 0.6442 & 0.6519 & 0.6363 & 0.6164 & 0.6851 & 0.6485 & 0.6460 & 0.6488 & 0.6751 & 0.6601 & 0.6614 & 0.6361 & 0.6732 & 0.6572 \\
Semantic Entropy & 0.6810 & 0.6968 & 0.6966 & 0.6958 & 0.7193 & 0.6438 & 0.7195 & 0.6933 & 0.7078 & 0.7110 & 0.7211 & 0.7052 & 0.7392 & 0.6738 & 0.7278 & 0.7123 \\
EigenScore & 0.6931 & 0.6891 & 0.6965 & 0.6748 & 0.7013 & 0.6556 & 0.7090 & 0.6885 & 0.7061 & 0.7018 & 0.7072 & 0.6854 & 0.7093 & 0.6648 & 0.7046 & 0.6970 \\
Lexical Similarity & 0.6498 & 0.6560 & 0.6779 & 0.6795 & 0.6529 & 0.6368 & 0.6834 & 0.6623 & 0.6752 & 0.6756 & 0.7050 & 0.7083 & 0.6583 & 0.6377 & 0.6730 & 0.6762 \\
Verbalize & 0.5349 & 0.5262 & 0.5274 & 0.5261 & 0.5329 & 0.5091 & 0.5578 & 0.5306 & 0.5286 & 0.5352 & 0.5511 & 0.5279 & 0.5460 & 0.5150 & 0.5550 & 0.5370 \\
InterrogateLLM & 0.7009 & 0.7127 & \underline{0.7062} & 0.7014 & 0.7496 & 0.6904 & 0.7435 & 0.7150 & 0.6978 & 0.6997 & 0.7248 & \underline{0.7273} & 0.7438 & \underline{0.6902} & 0.7376 & 0.7173 \\
\rowcolor{gray!18} \multicolumn{17}{l}{\textbf{Training-based methods}} \\
MM & 0.5033 & 0.5228 & 0.5270 & 0.4910 & 0.5120 & 0.4966 & 0.5269 & 0.5114 & 0.4844 & 0.4817 & 0.5243 & 0.5240 & 0.5280 & 0.4848 & 0.5116 & 0.5055 \\
SEP & 0.5277 & 0.5040 & 0.5072 & 0.5170 & 0.5258 & 0.4980 & 0.5317 & 0.5159 & 0.5395 & 0.5139 & 0.5284 & 0.5222 & 0.5301 & 0.4874 & 0.5350 & 0.5224 \\
SAPLMA & 0.5109 & 0.5192 & 0.5226 & 0.5385 & 0.5287 & 0.4916 & 0.5370 & 0.5212 & 0.5198 & 0.5174 & 0.5268 & 0.5303 & 0.5240 & 0.4868 & 0.5392 & 0.5206 \\
LapEigvals & 0.4840 & 0.5049 & 0.5113 & 0.4969 & 0.5062 & 0.4822 & 0.5263 & 0.5017 & 0.4712 & 0.4891 & 0.5108 & 0.4829 & 0.5289 & 0.4727 & 0.5291 & 0.4978 \\
\rowcolor{gray!18} \multicolumn{17}{l}{\textbf{Ablation variants}} \\
CHORD w/o RelAttn & 0.6538 & 0.7046 & 0.6887 & 0.6854 & 0.7645 & 0.6652 & 0.7594 & 0.7031 & 0.6988 & 0.7104 & 0.7213 & 0.7077 & 0.7609 & 0.6550 & 0.7640 & 0.7169 \\
CHORD w/o Meta & \underline{0.7108} & \underline{0.7158} & 0.6989 & \underline{0.7329} & \underline{0.7679} & \underline{0.6913} & \underline{0.7788} & \underline{0.7281} & \underline{0.7132} & \underline{0.7289} & \underline{0.7379} & 0.7214 & \underline{0.7661} & 0.6515 & \underline{0.7641} & \underline{0.7262} \\
\rowcolor{gray!18} \multicolumn{17}{l}{\textbf{Proposed method}} \\
\rowcolor{SkyBlueRow} \textbf{CHORD} & \textbf{0.7720} & \textbf{0.8237} & \textbf{0.8143} & \textbf{0.8070} & \textbf{0.8387} & \textbf{0.7248} & \textbf{0.8722} & \textbf{0.8075} & \textbf{0.8069} & \textbf{0.8104} & \textbf{0.8333} & \textbf{0.8027} & \textbf{0.8406} & \textbf{0.7207} & \textbf{0.8472} & \textbf{0.8088} \\
\midrule[1.5pt]
\multicolumn{17}{c}{} \\[-2mm]
& \multicolumn{8}{c}{Ministral3-8B} & \multicolumn{8}{c}{Ministral3-14B} \\
\cmidrule(lr){2-9}\cmidrule(lr){10-17}
\midrule
\rowcolor{gray!18} \multicolumn{17}{l}{\textbf{Training-free methods}} \\
Perplexity & 0.6443 & 0.6328 & 0.6576 & 0.6397 & 0.6528 & 0.6308 & 0.6785 & 0.6481 & 0.6335 & 0.6510 & 0.6471 & 0.6357 & 0.6526 & 0.6029 & 0.6635 & 0.6409 \\
Semantic Entropy & 0.6828 & 0.6800 & 0.7148 & 0.6867 & 0.7045 & 0.6424 & 0.7338 & 0.6921 & 0.6882 & 0.6925 & 0.7006 & 0.6930 & 0.7104 & 0.6534 & 0.7283 & 0.6952 \\
EigenScore & 0.6856 & 0.6729 & 0.6918 & 0.6788 & 0.6836 & 0.6519 & 0.7076 & 0.6817 & 0.6668 & 0.7015 & 0.6846 & 0.6746 & 0.6926 & 0.6509 & 0.7090 & 0.6829 \\
Lexical Similarity & 0.6530 & 0.6474 & 0.7075 & 0.6906 & 0.6479 & 0.6265 & 0.6698 & 0.6632 & 0.6632 & 0.6595 & 0.6945 & 0.6788 & 0.6610 & 0.6255 & 0.6682 & 0.6644 \\
Verbalize & 0.5019 & 0.5116 & 0.5440 & 0.5172 & 0.5230 & 0.5112 & 0.5528 & 0.5231 & 0.5143 & 0.5340 & 0.5158 & 0.5075 & 0.5231 & 0.5048 & 0.5506 & 0.5214 \\
InterrogateLLM & 0.6963 & \underline{0.7002} & 0.7030 & 0.6969 & 0.7254 & \underline{0.6961} & 0.7446 & 0.7089 & 0.6989 & 0.7295 & 0.7097 & 0.6862 & 0.7163 & \underline{0.6957} & 0.7432 & 0.7114 \\
\rowcolor{gray!18} \multicolumn{17}{l}{\textbf{Training-based methods}} \\
MM & 0.4807 & 0.4840 & 0.5294 & 0.5123 & 0.5298 & 0.4853 & 0.5285 & 0.5071 & 0.4866 & 0.5095 & 0.5123 & 0.4854 & 0.5284 & 0.4924 & 0.5232 & 0.5054 \\
SEP & 0.5232 & 0.5188 & 0.5162 & 0.4990 & 0.5033 & 0.5030 & 0.5288 & 0.5132 & 0.4860 & 0.5076 & 0.5123 & 0.4999 & 0.5186 & 0.4895 & 0.5391 & 0.5076 \\
SAPLMA & 0.4816 & 0.5275 & 0.5396 & 0.5073 & 0.5335 & 0.4944 & 0.5392 & 0.5176 & 0.4865 & 0.5232 & 0.4952 & 0.5091 & 0.5266 & 0.4839 & 0.5386 & 0.5090 \\
LapEigvals & 0.4708 & 0.4737 & 0.5019 & 0.5031 & 0.4983 & 0.4733 & 0.5295 & 0.4929 & 0.4847 & 0.4766 & 0.5179 & 0.4800 & 0.5058 & 0.4702 & 0.5161 & 0.4930 \\
\rowcolor{gray!18} \multicolumn{17}{l}{\textbf{Ablation variants}} \\
CHORD w/o RelAttn & 0.7051 & 0.6523 & 0.7346 & 0.6500 & 0.7148 & 0.6591 & 0.7514 & 0.6953 & 0.6816 & 0.7271 & 0.7152 & 0.6713 & 0.6829 & 0.6463 & 0.7328 & 0.6939 \\
CHORD w/o Meta & \underline{0.7084} & 0.6484 & \underline{0.7499} & \underline{0.7029} & \underline{0.7574} & 0.6821 & \underline{0.7738} & \underline{0.7176} & \underline{0.7195} & \underline{0.7341} & \underline{0.7235} & \underline{0.7115} & \underline{0.7186} & 0.6456 & \underline{0.7594} & \underline{0.7160} \\
\rowcolor{gray!18} \multicolumn{17}{l}{\textbf{Proposed method}} \\
\rowcolor{SkyBlueRow} \textbf{CHORD} & \textbf{0.7663} & \textbf{0.8005} & \textbf{0.7914} & \textbf{0.7836} & \textbf{0.8021} & \textbf{0.7263} & \textbf{0.8474} & \textbf{0.7882} & \textbf{0.7760} & \textbf{0.7988} & \textbf{0.8046} & \textbf{0.7933} & \textbf{0.8193} & \textbf{0.7237} & \textbf{0.8557} & \textbf{0.7959} \\
\bottomrule
\end{tabular}%
}
\vspace{-2mm}
\end{table*}

This design avoids repeating all baseline rows in the main text while preserving the main task-level differences.

The variant \method{} w/o RelAttn removes relation-aware attention while keeping the remaining framework. 
In this setting, edge-level relational information is no longer explicitly propagated into node representations. 
The performance drop is usually larger than that of the meta-learning ablation. 
This indicates that the improvement of \method{} does not come merely from hidden-state restructuring, padding, or graph node construction. 
Instead, explicitly propagating pairwise relational information is central to obtaining transferable graph-level features. 
In other words, the layer-token-feature graph is useful not only because it reorganizes hidden states, but also because its edges are used by the relation-aware encoder.

The variant \method{} w/o Meta keeps the relation-aware architecture but replaces the meta-learning objective with standard empirical risk minimization over source-family examples. 
This variant usually performs better than \method{} w/o RelAttn, which suggests that the relational architecture itself already provides useful structure. 
However, it still falls below full \method{} and shows less stable performance across target families. 
This supports the role of meta-learning in the cross-model setting. 
By repeatedly simulating held-out model transfer during training, the meta-learning objective discourages the detector from relying only on source-family-specific shortcuts and encourages it to use relational features that are more transferable.

Both ablated variants remain above ordinary training-based baselines in most aggregated results. 
This suggests that the unified relational formulation already addresses part of the cross-model mismatch. 
However, neither variant consistently dominates the strongest training-free baseline. 
This pattern is important because it shows that the two modules are complementary. 
Relation-aware attention provides the structural mechanism for absorbing edge-level relations into node representations, while meta-learning improves the stability of the learned detector under family-held-out transfer. 
Only their combination yields the consistent advantage of full \method{}.

The relational design pays off most clearly on GSM8K and xLAM: both ablated variants still exceed the strongest training-free baseline there ($0.7344$ and $0.7572$ versus $0.7297$ on GSM8K; $0.7584$ and $0.7740$ versus $0.7421$ on xLAM), and full \method{} attains its largest margins on these two benchmarks.
This suggests that reasoning-oriented and agent-oriented hallucinations benefit more from modeling relational internal structures. 
For GSM8K, errors often emerge through multi-step reasoning, where token and layer relations may encode the progression of incorrect reasoning. 
For xLAM, agent-oriented errors may involve tool choice, action construction, and trajectory consistency, which can also produce structured internal deviations. 
By contrast, MBPP is more fragile. 
The full model remains strongest, but ablated variants often move closer to strong training-free baselines. 
This is consistent with the limited training size and execution-dependent nature of code correctness discussed above. 
Overall, the ablation study confirms that relation-aware attention and meta-learning are both important, and that the stable performance of \method{} comes from their combination rather than either component alone.

\section{Experimental Extension}
\label{app:exp_exp_exp}

\subsection{Scale-Group Transfer}
\label{app:exp_scale_group_transfer}

The main experiments evaluate cross-model generalization under a leave-one-family-out protocol. 
Here we provide a complementary analysis that asks whether the detector also transfers across model scales. 
We partition the target models into three coarse scale groups: the small-scale group contains Llama3.2-3B, Llama3.1-8B, Qwen3-8B, and Ministral3-8B; the medium-scale group contains Qwen3-14B and Ministral3-14B; and the large-scale group contains Llama3.3-70B and Qwen3.5-35B. 
The groups are intended to probe transfer across scale ranges rather than to form a perfectly controlled size-only comparison, since available open-source models are not family-balanced at every parameter scale.

\begin{table*}[t]
\caption{Cross-size transfer results. Models are grouped by parameter scale; training-based methods are trained on source-size-group models and evaluated on target-size-group models. Values are AUROC averaged over target models in the target group. Best results are \textbf{bold} and second-best results are \underline{underlined}.}
\vspace{2mm}
\centering
\label{tab:cross_size_full}
\resizebox{\linewidth}{!}{%
\begin{tabular}{lcccccccccccccccc}
\toprule
\multirow{4}{*}{\textbf{Method}} & \multicolumn{16}{c}{\textbf{Scale-Group Transfer}} \\
& \multicolumn{8}{c}{Small$\rightarrow$Medium} & \multicolumn{8}{c}{Medium$\rightarrow$Small} \\
\cmidrule(lr){2-9}\cmidrule(lr){10-17}
& \multicolumn{8}{c}{Mean AUROC$\uparrow$} & \multicolumn{8}{c}{Mean AUROC$\uparrow$} \\
& TruthfulQA & HaluEval & TriviaQA & CoQA & GSM8K & MBPP & xLAM & \textbf{Average} & TruthfulQA & HaluEval & TriviaQA & CoQA & GSM8K & MBPP & xLAM & \textbf{Average} \\
\midrule
\rowcolor{gray!18} \multicolumn{17}{l}{\textbf{Training-free methods}} \\
Perplexity & 0.6371 & 0.6580 & 0.6457 & 0.6438 & 0.6444 & 0.6097 & 0.6743 & 0.6447 & 0.6416 & 0.6411 & 0.6588 & 0.6407 & 0.6560 & 0.6175 & 0.6747 & 0.6472 \\
Semantic Entropy & 0.6846 & 0.6946 & 0.6986 & \underline{0.6944} & 0.7148 & 0.6486 & 0.7239 & 0.6942 & 0.6880 & 0.6952 & \underline{0.7087} & 0.6928 & 0.7140 & 0.6533 & 0.7307 & 0.6975 \\
EigenScore & 0.6800 & 0.6953 & 0.6905 & 0.6747 & 0.6969 & 0.6532 & 0.7090 & 0.6857 & 0.6815 & 0.6845 & 0.6839 & 0.6782 & 0.6872 & 0.6581 & 0.7041 & 0.6825 \\
Lexical Similarity & 0.6565 & 0.6578 & 0.6862 & 0.6791 & 0.6570 & 0.6311 & 0.6758 & 0.6634 & 0.6593 & 0.6610 & 0.6992 & 0.6822 & 0.6467 & 0.6282 & 0.6715 & 0.6640 \\
Verbalize & 0.5246 & 0.5301 & 0.5216 & 0.5168 & 0.5280 & 0.5070 & 0.5542 & 0.5260 & 0.5162 & 0.5227 & 0.5242 & 0.5171 & 0.5312 & 0.5009 & 0.5514 & 0.5234 \\
InterrogateLLM & \underline{0.6999} & \underline{0.7211} & \underline{0.7080} & 0.6938 & \underline{0.7329} & \underline{0.6930} & \underline{0.7433} & \underline{0.7132} & \underline{0.6986} & \underline{0.7161} & 0.7072 & \underline{0.6985} & \underline{0.7202} & \underline{0.6894} & \underline{0.7413} & \underline{0.7102} \\
\rowcolor{gray!18} \multicolumn{17}{l}{\textbf{Training-based methods}} \\
MM & 0.5462 & 0.5881 & 0.5885 & 0.5490 & 0.6051 & 0.5384 & 0.6079 & 0.5747 & 0.5695 & 0.5681 & 0.5762 & 0.5854 & 0.5897 & 0.5487 & 0.6034 & 0.5773 \\
SEP & 0.5749 & 0.5799 & 0.5714 & 0.5829 & 0.6167 & 0.5559 & 0.6155 & 0.5853 & 0.6001 & 0.6005 & 0.6048 & 0.5915 & 0.6159 & 0.5543 & 0.6105 & 0.5968 \\
SAPLMA & 0.6014 & 0.5986 & 0.5956 & 0.6181 & 0.6305 & 0.5530 & 0.6363 & 0.6048 & 0.5885 & 0.6161 & 0.6151 & 0.6062 & 0.6196 & 0.5627 & 0.6315 & 0.6057 \\
LapEigvals & 0.5399 & 0.5568 & 0.5706 & 0.5592 & 0.5620 & 0.5182 & 0.5916 & 0.5569 & 0.5422 & 0.5568 & 0.5541 & 0.5770 & 0.5891 & 0.5197 & 0.5933 & 0.5618 \\
\rowcolor{gray!18} \multicolumn{17}{l}{\textbf{Proposed method}} \\
\rowcolor{SkyBlueRow} \textbf{CHORD} & \textbf{0.7429} & \textbf{0.7797} & \textbf{0.7791} & \textbf{0.7688} & \textbf{0.8146} & \textbf{0.7063} & \textbf{0.8407} & \textbf{0.7760} & \textbf{0.7530} & \textbf{0.7740} & \textbf{0.7707} & \textbf{0.7557} & \textbf{0.7977} & \textbf{0.7009} & \textbf{0.8298} & \textbf{0.7688} \\
\midrule[1.5pt]
\multicolumn{17}{c}{} \\[-2mm]
& \multicolumn{8}{c}{Small$\rightarrow$Large} & \multicolumn{8}{c}{Large$\rightarrow$Small} \\
\cmidrule(lr){2-9}\cmidrule(lr){10-17}
\midrule
\rowcolor{gray!18} \multicolumn{17}{l}{\textbf{Training-free methods}} \\
Perplexity & 0.6518 & 0.6618 & 0.6734 & 0.6674 & 0.6689 & 0.6357 & 0.6832 & 0.6632 & 0.6416 & 0.6411 & 0.6588 & 0.6407 & 0.6560 & 0.6175 & 0.6747 & 0.6472 \\
Semantic Entropy & 0.7087 & \underline{0.7238} & 0.7221 & 0.7162 & 0.7401 & 0.6717 & 0.7342 & 0.7167 & 0.6880 & 0.6952 & \underline{0.7087} & 0.6928 & 0.7140 & 0.6533 & 0.7307 & 0.6975 \\
EigenScore & 0.7068 & 0.7052 & 0.7017 & 0.6975 & 0.7094 & 0.6675 & 0.7069 & 0.6993 & 0.6815 & 0.6845 & 0.6839 & 0.6782 & 0.6872 & 0.6581 & 0.7041 & 0.6825 \\
Lexical Similarity & 0.6798 & 0.6776 & 0.7045 & 0.7086 & 0.6719 & 0.6404 & 0.6794 & 0.6803 & 0.6593 & 0.6610 & 0.6992 & 0.6822 & 0.6467 & 0.6282 & 0.6715 & 0.6640 \\
Verbalize & 0.5379 & 0.5442 & 0.5492 & 0.5332 & 0.5526 & 0.5145 & 0.5567 & 0.5412 & 0.5162 & 0.5227 & 0.5242 & 0.5171 & 0.5312 & 0.5009 & 0.5514 & 0.5234 \\
InterrogateLLM & \underline{0.7160} & 0.7226 & \underline{0.7360} & \underline{0.7246} & \underline{0.7455} & \underline{0.7042} & \underline{0.7427} & \underline{0.7274} & \underline{0.6986} & \underline{0.7161} & 0.7072 & \underline{0.6985} & \underline{0.7202} & \underline{0.6894} & \underline{0.7413} & \underline{0.7102} \\
\rowcolor{gray!18} \multicolumn{17}{l}{\textbf{Training-based methods}} \\
MM & 0.5353 & 0.5522 & 0.5816 & 0.5646 & 0.5857 & 0.5176 & 0.5795 & 0.5595 & 0.5649 & 0.5814 & 0.6076 & 0.5937 & 0.6073 & 0.5566 & 0.6055 & 0.5881 \\
SEP & 0.5872 & 0.5845 & 0.5987 & 0.5931 & 0.5990 & 0.5219 & 0.5995 & 0.5834 & 0.6144 & 0.6029 & 0.5877 & 0.5972 & 0.6209 & 0.5611 & 0.6273 & 0.6017 \\
SAPLMA & 0.5821 & 0.5906 & 0.5846 & 0.6047 & 0.5962 & 0.5306 & 0.6233 & 0.5874 & 0.6020 & 0.6346 & 0.6244 & 0.6277 & 0.6313 & 0.5630 & 0.6429 & 0.6180 \\
LapEigvals & 0.5316 & 0.5419 & 0.5554 & 0.5517 & 0.5789 & 0.5150 & 0.5886 & 0.5519 & 0.5539 & 0.5742 & 0.5608 & 0.5793 & 0.6030 & 0.5269 & 0.6246 & 0.5747 \\
\rowcolor{gray!18} \multicolumn{17}{l}{\textbf{Proposed method}} \\
\rowcolor{SkyBlueRow} \textbf{CHORD} & \textbf{0.7567} & \textbf{0.7580} & \textbf{0.7785} & \textbf{0.7580} & \textbf{0.7900} & \textbf{0.7150} & \textbf{0.8089} & \textbf{0.7664} & \textbf{0.7660} & \textbf{0.7882} & \textbf{0.7925} & \textbf{0.7746} & \textbf{0.8144} & \textbf{0.7126} & \textbf{0.8495} & \textbf{0.7854} \\
\midrule[1.5pt]
\multicolumn{17}{c}{} \\[-2mm]
& \multicolumn{8}{c}{Medium$\rightarrow$Large} & \multicolumn{8}{c}{Large$\rightarrow$Medium} \\
\cmidrule(lr){2-9}\cmidrule(lr){10-17}
\midrule
\rowcolor{gray!18} \multicolumn{17}{l}{\textbf{Training-free methods}} \\
Perplexity & 0.6518 & 0.6618 & 0.6734 & 0.6674 & 0.6689 & 0.6357 & 0.6832 & 0.6632 & 0.6371 & 0.6580 & 0.6457 & 0.6438 & 0.6444 & 0.6097 & 0.6743 & 0.6447 \\
Semantic Entropy & 0.7087 & \underline{0.7238} & 0.7221 & 0.7162 & 0.7401 & 0.6717 & 0.7342 & 0.7167 & 0.6846 & 0.6946 & 0.6986 & \underline{0.6944} & 0.7148 & 0.6486 & 0.7239 & 0.6942 \\
EigenScore & 0.7068 & 0.7052 & 0.7017 & 0.6975 & 0.7094 & 0.6675 & 0.7069 & 0.6993 & 0.6800 & 0.6953 & 0.6905 & 0.6747 & 0.6969 & 0.6532 & 0.7090 & 0.6857 \\
Lexical Similarity & 0.6798 & 0.6776 & 0.7045 & 0.7086 & 0.6719 & 0.6404 & 0.6794 & 0.6803 & 0.6565 & 0.6578 & 0.6862 & 0.6791 & 0.6570 & 0.6311 & 0.6758 & 0.6634 \\
Verbalize & 0.5379 & 0.5442 & 0.5492 & 0.5332 & 0.5526 & 0.5145 & 0.5567 & 0.5412 & 0.5246 & 0.5301 & 0.5216 & 0.5168 & 0.5280 & 0.5070 & 0.5542 & 0.5260 \\
InterrogateLLM & \underline{0.7160} & 0.7226 & \underline{0.7360} & \underline{0.7246} & \underline{0.7455} & \underline{0.7042} & \underline{0.7427} & \underline{0.7274} & \underline{0.6999} & \underline{0.7211} & \underline{0.7080} & 0.6938 & \underline{0.7329} & \underline{0.6930} & \underline{0.7433} & \underline{0.7132} \\
\rowcolor{gray!18} \multicolumn{17}{l}{\textbf{Training-based methods}} \\
MM & 0.5477 & 0.5621 & 0.5806 & 0.5781 & 0.5877 & 0.5176 & 0.5829 & 0.5653 & 0.5677 & 0.5966 & 0.5957 & 0.5714 & 0.5946 & 0.5563 & 0.6083 & 0.5844 \\
SEP & 0.5796 & 0.5934 & 0.5985 & 0.5986 & 0.5923 & 0.5442 & 0.6024 & 0.5870 & 0.5948 & 0.5847 & 0.5896 & 0.5868 & 0.6122 & 0.5523 & 0.6278 & 0.5926 \\
SAPLMA & 0.6130 & 0.5841 & 0.6004 & 0.6049 & 0.6164 & 0.5333 & 0.6129 & 0.5950 & 0.6049 & 0.6301 & 0.6012 & 0.6220 & 0.6125 & 0.5510 & 0.6380 & 0.6085 \\
LapEigvals & 0.5317 & 0.5450 & 0.5676 & 0.5564 & 0.5648 & 0.5155 & 0.5780 & 0.5513 & 0.5357 & 0.5572 & 0.5901 & 0.5810 & 0.5699 & 0.5301 & 0.6096 & 0.5677 \\
\rowcolor{gray!18} \multicolumn{17}{l}{\textbf{Proposed method}} \\
\rowcolor{SkyBlueRow} \textbf{CHORD} & \textbf{0.7699} & \textbf{0.7824} & \textbf{0.7923} & \textbf{0.7747} & \textbf{0.8195} & \textbf{0.7131} & \textbf{0.8261} & \textbf{0.7826} & \textbf{0.7463} & \textbf{0.7888} & \textbf{0.7805} & \textbf{0.7772} & \textbf{0.8142} & \textbf{0.7050} & \textbf{0.8498} & \textbf{0.7803} \\
\bottomrule
\end{tabular}%
}
\vspace{-2mm}
\end{table*}

For each ordered pair of distinct scale groups, training-based methods are trained on all models in the source scale group and evaluated on every model in the target scale group. 
For example, Small$\rightarrow$Large trains on the small-scale group and tests on the large-scale target models, while Large$\rightarrow$Small reverses this direction. 
Training-free methods do not use source models, so their values depend only on the target scale group; repeated values across different source groups with the same target group reflect this source-independent nature. 
Table~\ref{tab:cross_size_full} reports the full per-benchmark results for all directed scale-group transfers.

Overall, \method{} remains the strongest method across scale-group transfer directions. 
The result indicates that the relational detector is not tied to a narrow parameter scale: even when the source and target groups differ substantially in size, \method{} maintains a clear advantage over both training-free and ordinary training-based baselines. 
At the same time, the transfer directions are not identical. 
Transfer between the two extreme groups is asymmetric: Large$\rightarrow$Small reaches $0.7854$, compared with $0.7664$ for Small$\rightarrow$Large. 
This direction effect is not consistent across the remaining pairs, however: Small$\rightarrow$Medium ($0.7760$) slightly exceeds Medium$\rightarrow$Small ($0.7688$), and Medium$\rightarrow$Large ($0.7826$) slightly exceeds Large$\rightarrow$Medium ($0.7803$). 
Larger source models may therefore expose richer relational structure in some cases, but scale direction alone does not determine transfer difficulty.

The scale-group results also show that source composition matters in addition to scale. 
The small group contains more models and covers more model families, whereas the medium and large groups contain fewer models. 
Thus, absolute AUROC reflects both scale direction and the diversity of the source group. 
This is why the experiment should be interpreted as scale-group transfer rather than a causal isolation of parameter count. 
Despite this caveat, the main trend is stable: \method{} remains robust across all directed scale-group shifts, and its gains do not depend on training and testing within the same size regime.

The behavior of the baselines provides additional context. 
Training-free methods, especially Semantic Entropy and InterrogateLLM, remain competitive because they are naturally source-independent. 
However, they do not benefit from source-model training and therefore lag behind \method{} in most settings. 
Ordinary training-based baselines perform better here than in the leave-one-family-out main experiment, which is expected because each source scale group can include multiple model families and therefore provides a less severe distribution shift than a fully held-out family. 
Nevertheless, their performance remains clearly below \method{}, suggesting that simply training on a mixture of source-scale models is insufficient for stable cross-scale generalization.

The task-level results follow the same qualitative pattern as the main experiments. 
xLAM and GSM8K show strong gains for \method{}, indicating that relational internal features are especially useful for agent-oriented and reasoning-heavy hallucination detection. 
MBPP remains the most fragile benchmark, where the gap between \method{} and the strongest training-free baselines is smaller. 
This is consistent with the code setting being more dependent on execution semantics and fine-grained program correctness. 
Taken together, Table~\ref{tab:cross_size_full} shows that \method{}'s cross-model robustness extends beyond family transfer to scale-group transfer, while also revealing that transfer direction and source-group diversity modulate the absolute difficulty of the setting.

\subsection{Fine-Grained Edge Ablation}
\label{app:exp_edge_ablation}

We next analyze which relation channels contribute to \method{}'s performance. 
The main ablation in Table~\ref{tab:ablation_main} removes larger components such as relation-aware attention or meta-learning. 
Here we perform a finer-grained edge ablation by removing one relation channel at a time while keeping the rest of the method unchanged. 
Specifically, we compare full \method{} with three variants: \method{} w/o Token, \method{} w/o Layer, and \method{} w/o Intra. 
The first removes the token-level edge channel, the second removes the layer-level edge channel, and the third removes the intra-block feature edge channel. 
All other components, including hidden representation restructuring, the remaining edge channels, relation-aware encoding, pooling, and meta-learning, are kept the same. 
This design isolates whether each relation channel provides complementary information rather than testing whether the entire relational architecture is necessary.

\begin{table*}[t]
\caption{Full edge-ablation results under the cross-family evaluation protocol. Values are AUROC. Best results are \textbf{bold} and second-best results are \underline{underlined}.}
\vspace{2mm}
\centering
\label{tab:edge_ablation_full}
\resizebox{\linewidth}{!}{%
\begin{tabular}{lcccccccccccccccc}
\toprule
\multirow{4}{*}{\textbf{Method}} & \multicolumn{16}{c}{\textbf{Target Model}} \\
& \multicolumn{8}{c}{Llama3.2-3B} & \multicolumn{8}{c}{Llama3.1-8B} \\
\cmidrule(lr){2-9}\cmidrule(lr){10-17}
& \multicolumn{8}{c}{Mean AUROC$\uparrow$} & \multicolumn{8}{c}{Mean AUROC$\uparrow$} \\
& TruthfulQA & HaluEval & TriviaQA & CoQA & GSM8K & MBPP & xLAM & \textbf{Average} & TruthfulQA & HaluEval & TriviaQA & CoQA & GSM8K & MBPP & xLAM & \textbf{Average} \\
\midrule
\rowcolor{gray!18} \multicolumn{17}{l}{\textbf{Training-free methods}} \\
Perplexity & 0.6329 & 0.6464 & 0.6448 & 0.6296 & 0.6536 & 0.6188 & 0.6621 & 0.6412 & 0.6546 & 0.6452 & 0.6545 & 0.6395 & 0.6523 & 0.6024 & 0.6857 & 0.6477 \\
Semantic Entropy & 0.6862 & 0.7095 & 0.7006 & 0.6934 & 0.7112 & 0.6571 & 0.7223 & 0.6972 & 0.6991 & 0.7113 & 0.7065 & 0.6919 & 0.7258 & 0.6636 & 0.7438 & 0.7060 \\
EigenScore & 0.6636 & 0.6872 & 0.6767 & 0.6719 & 0.6772 & 0.6508 & 0.7045 & 0.6760 & 0.6985 & 0.6963 & 0.6803 & 0.6747 & 0.7068 & 0.6614 & 0.6957 & 0.6877 \\
Lexical Similarity & 0.6542 & 0.6747 & 0.6878 & 0.6786 & 0.6454 & 0.6225 & 0.6603 & 0.6605 & 0.6760 & 0.6681 & 0.6928 & 0.6628 & 0.6495 & 0.6236 & 0.6871 & 0.6657 \\
Verbalize & 0.5199 & 0.5306 & 0.5220 & 0.5290 & 0.5207 & 0.5069 & 0.5470 & 0.5252 & 0.5238 & 0.5398 & 0.5110 & 0.5050 & 0.5474 & 0.4884 & 0.5595 & 0.5250 \\
InterrogateLLM & 0.6845 & 0.7353 & 0.7012 & 0.6915 & 0.7177 & 0.6915 & 0.7253 & 0.7067 & 0.7185 & 0.7211 & 0.7201 & 0.6881 & 0.7293 & 0.6865 & 0.7469 & 0.7158 \\
\rowcolor{gray!18} \multicolumn{17}{l}{\textbf{Training-based methods}} \\
MM & 0.4857 & 0.5170 & 0.5129 & 0.4874 & 0.5061 & 0.5035 & 0.5200 & 0.5047 & 0.5127 & 0.5029 & 0.5247 & 0.5000 & 0.5141 & 0.4808 & 0.5052 & 0.5058 \\
SEP & 0.5079 & 0.5193 & 0.5373 & 0.5189 & 0.5300 & 0.5032 & 0.5290 & 0.5208 & 0.5191 & 0.5346 & 0.5142 & 0.5089 & 0.5382 & 0.4876 & 0.5395 & 0.5203 \\
SAPLMA & 0.4941 & 0.5191 & 0.5239 & 0.5390 & 0.5158 & 0.4901 & 0.5397 & 0.5174 & 0.5326 & 0.5311 & 0.5250 & 0.5013 & 0.5382 & 0.4818 & 0.5386 & 0.5212 \\
LapEigvals & 0.4817 & 0.5021 & 0.4740 & 0.4965 & 0.5264 & 0.4765 & 0.5295 & 0.4981 & 0.4981 & 0.4916 & 0.4712 & 0.4789 & 0.5083 & 0.4725 & 0.5225 & 0.4919 \\
\rowcolor{gray!18} \multicolumn{17}{l}{\textbf{Edge ablation variants}} \\
CHORD w/o Token & \underline{0.7433} & 0.7540 & \underline{0.7695} & 0.7344 & \underline{0.7837} & \underline{0.7057} & 0.7719 & \underline{0.7518} & \underline{0.7797} & 0.7652 & \underline{0.7881} & 0.7293 & 0.7925 & 0.6807 & 0.8010 & 0.7624 \\
CHORD w/o Layer & 0.7140 & \underline{0.7677} & 0.7437 & \underline{0.7603} & 0.7389 & 0.7053 & \underline{0.8048} & 0.7478 & 0.7507 & 0.7770 & 0.7851 & \underline{0.7522} & 0.7546 & \underline{0.6931} & \underline{0.8272} & \underline{0.7628} \\
CHORD w/o Intra & 0.7405 & 0.7518 & 0.7430 & 0.7573 & 0.7738 & 0.6957 & 0.7689 & 0.7473 & 0.7532 & \underline{0.7850} & 0.7636 & 0.7457 & \underline{0.7982} & 0.6779 & 0.7978 & 0.7602 \\
\rowcolor{gray!18} \multicolumn{17}{l}{\textbf{Proposed method}} \\
\rowcolor{SkyBlueRow} \textbf{CHORD} & \textbf{0.7766} & \textbf{0.7926} & \textbf{0.7991} & \textbf{0.7920} & \textbf{0.8193} & \textbf{0.7351} & \textbf{0.8437} & \textbf{0.7941} & \textbf{0.7989} & \textbf{0.8102} & \textbf{0.8156} & \textbf{0.7823} & \textbf{0.8362} & \textbf{0.7307} & \textbf{0.8650} & \textbf{0.8056} \\
\midrule[1.5pt]
\multicolumn{17}{c}{} \\[-2mm]
& \multicolumn{8}{c}{Llama3.3-70B} & \multicolumn{8}{c}{Qwen3-8B} \\
\cmidrule(lr){2-9}\cmidrule(lr){10-17}
\midrule
\rowcolor{gray!18} \multicolumn{17}{l}{\textbf{Training-free methods}} \\
Perplexity & 0.6575 & 0.6747 & 0.6717 & 0.6748 & 0.6765 & 0.6353 & 0.6931 & 0.6691 & 0.6345 & 0.6398 & 0.6781 & 0.6542 & 0.6653 & 0.6179 & 0.6726 & 0.6518 \\
Semantic Entropy & 0.7096 & 0.7366 & 0.7231 & 0.7272 & 0.7410 & 0.6697 & 0.7406 & 0.7211 & 0.6840 & 0.6800 & 0.7127 & 0.6994 & 0.7143 & 0.6502 & 0.7228 & 0.6948 \\
EigenScore & 0.7076 & 0.7087 & 0.6961 & 0.7095 & 0.7094 & 0.6703 & 0.7093 & 0.7016 & 0.6785 & 0.6817 & 0.6869 & 0.6873 & 0.6813 & 0.6683 & 0.7085 & 0.6846 \\
Lexical Similarity & 0.6844 & 0.6797 & 0.7040 & 0.7088 & 0.6854 & 0.6432 & 0.6858 & 0.6845 & 0.6538 & 0.6538 & 0.7089 & 0.6966 & 0.6441 & 0.6401 & 0.6688 & 0.6666 \\
Verbalize & 0.5472 & 0.5531 & 0.5472 & 0.5385 & 0.5592 & 0.5139 & 0.5584 & 0.5454 & 0.5193 & 0.5090 & 0.5200 & 0.5170 & 0.5339 & 0.4969 & 0.5463 & 0.5203 \\
InterrogateLLM & 0.7343 & 0.7454 & 0.7472 & 0.7219 & 0.7473 & 0.7181 & 0.7477 & 0.7374 & 0.6950 & 0.7077 & 0.7044 & 0.7173 & 0.7082 & 0.6834 & 0.7483 & 0.7092 \\
\rowcolor{gray!18} \multicolumn{17}{l}{\textbf{Training-based methods}} \\
MM & 0.5095 & 0.5244 & 0.5288 & 0.5195 & 0.5259 & 0.4814 & 0.5153 & 0.5150 & 0.4923 & 0.4849 & 0.5044 & 0.5095 & 0.5046 & 0.4837 & 0.5171 & 0.4995 \\
SEP & 0.5117 & 0.5352 & 0.5400 & 0.5243 & 0.5263 & 0.4905 & 0.5297 & 0.5225 & 0.5147 & 0.5030 & 0.4993 & 0.5062 & 0.5159 & 0.4823 & 0.5228 & 0.5063 \\
SAPLMA & 0.5250 & 0.5177 & 0.5231 & 0.5364 & 0.5375 & 0.4840 & 0.5386 & 0.5232 & 0.4822 & 0.5223 & 0.5171 & 0.5093 & 0.5167 & 0.4922 & 0.5384 & 0.5112 \\
LapEigvals & 0.5083 & 0.4992 & 0.5167 & 0.5204 & 0.5132 & 0.4718 & 0.5280 & 0.5082 & 0.4831 & 0.4872 & 0.5141 & 0.4873 & 0.5063 & 0.4716 & 0.5292 & 0.4970 \\
\rowcolor{gray!18} \multicolumn{17}{l}{\textbf{Edge ablation variants}} \\
CHORD w/o Token & 0.7958 & 0.7940 & \underline{0.7967} & 0.7780 & \underline{0.8352} & \underline{0.7443} & 0.8039 & \underline{0.7926} & 0.7322 & 0.7391 & 0.7893 & 0.7432 & \underline{0.7830} & 0.6897 & 0.7707 & 0.7496 \\
CHORD w/o Layer & 0.7770 & 0.8042 & 0.7920 & \underline{0.8008} & 0.7813 & 0.7168 & \underline{0.8252} & 0.7853 & 0.7313 & \underline{0.7518} & \underline{0.7941} & \underline{0.7709} & 0.7386 & \underline{0.7104} & \underline{0.8140} & \underline{0.7587} \\
CHORD w/o Intra & \underline{0.8136} & \underline{0.8078} & 0.7874 & 0.7908 & 0.8180 & 0.7121 & 0.8095 & 0.7913 & \underline{0.7599} & 0.7478 & 0.7649 & 0.7701 & 0.7626 & 0.6983 & 0.7720 & 0.7537 \\
\rowcolor{gray!18} \multicolumn{17}{l}{\textbf{Proposed method}} \\
\rowcolor{SkyBlueRow} \textbf{CHORD} & \textbf{0.8276} & \textbf{0.8374} & \textbf{0.8257} & \textbf{0.8302} & \textbf{0.8568} & \textbf{0.7605} & \textbf{0.8713} & \textbf{0.8299} & \textbf{0.7896} & \textbf{0.7941} & \textbf{0.8192} & \textbf{0.7991} & \textbf{0.8256} & \textbf{0.7481} & \textbf{0.8484} & \textbf{0.8034} \\
\midrule[1.5pt]
\multicolumn{17}{c}{} \\[-2mm]
& \multicolumn{8}{c}{Qwen3-14B} & \multicolumn{8}{c}{Qwen3.5-35B} \\
\cmidrule(lr){2-9}\cmidrule(lr){10-17}
\midrule
\rowcolor{gray!18} \multicolumn{17}{l}{\textbf{Training-free methods}} \\
Perplexity & 0.6408 & 0.6650 & 0.6442 & 0.6519 & 0.6363 & 0.6164 & 0.6851 & 0.6485 & 0.6460 & 0.6488 & 0.6751 & 0.6601 & 0.6614 & 0.6361 & 0.6732 & 0.6572 \\
Semantic Entropy & 0.6810 & 0.6968 & 0.6966 & 0.6958 & 0.7193 & 0.6438 & 0.7195 & 0.6933 & 0.7078 & 0.7110 & 0.7211 & 0.7052 & 0.7392 & 0.6738 & 0.7278 & 0.7123 \\
EigenScore & 0.6931 & 0.6891 & 0.6965 & 0.6748 & 0.7013 & 0.6556 & 0.7090 & 0.6885 & 0.7061 & 0.7018 & 0.7072 & 0.6854 & 0.7093 & 0.6648 & 0.7046 & 0.6970 \\
Lexical Similarity & 0.6498 & 0.6560 & 0.6779 & 0.6795 & 0.6529 & 0.6368 & 0.6834 & 0.6623 & 0.6752 & 0.6756 & 0.7050 & 0.7083 & 0.6583 & 0.6377 & 0.6730 & 0.6762 \\
Verbalize & 0.5349 & 0.5262 & 0.5274 & 0.5261 & 0.5329 & 0.5091 & 0.5578 & 0.5306 & 0.5286 & 0.5352 & 0.5511 & 0.5279 & 0.5460 & 0.5150 & 0.5550 & 0.5370 \\
InterrogateLLM & 0.7009 & 0.7127 & 0.7062 & 0.7014 & 0.7496 & 0.6904 & 0.7435 & 0.7150 & 0.6978 & 0.6997 & 0.7248 & 0.7273 & 0.7438 & 0.6902 & 0.7376 & 0.7173 \\
\rowcolor{gray!18} \multicolumn{17}{l}{\textbf{Training-based methods}} \\
MM & 0.5033 & 0.5228 & 0.5270 & 0.4910 & 0.5120 & 0.4966 & 0.5269 & 0.5114 & 0.4844 & 0.4817 & 0.5243 & 0.5240 & 0.5280 & 0.4848 & 0.5116 & 0.5055 \\
SEP & 0.5277 & 0.5040 & 0.5072 & 0.5170 & 0.5258 & 0.4980 & 0.5317 & 0.5159 & 0.5395 & 0.5139 & 0.5284 & 0.5222 & 0.5301 & 0.4874 & 0.5350 & 0.5224 \\
SAPLMA & 0.5109 & 0.5192 & 0.5226 & 0.5385 & 0.5287 & 0.4916 & 0.5370 & 0.5212 & 0.5198 & 0.5174 & 0.5268 & 0.5303 & 0.5240 & 0.4868 & 0.5392 & 0.5206 \\
LapEigvals & 0.4840 & 0.5049 & 0.5113 & 0.4969 & 0.5062 & 0.4822 & 0.5263 & 0.5017 & 0.4712 & 0.4891 & 0.5108 & 0.4829 & 0.5289 & 0.4727 & 0.5291 & 0.4978 \\
\rowcolor{gray!18} \multicolumn{17}{l}{\textbf{Edge ablation variants}} \\
CHORD w/o Token & 0.7342 & 0.7800 & \underline{0.7768} & 0.7525 & \underline{0.8129} & 0.6913 & 0.8012 & 0.7641 & 0.7558 & 0.7668 & \underline{0.8129} & 0.7515 & \underline{0.7993} & 0.6932 & 0.7758 & 0.7650 \\
CHORD w/o Layer & 0.7313 & \underline{0.7941} & 0.7710 & \underline{0.7785} & 0.7708 & \underline{0.7004} & \underline{0.8193} & \underline{0.7665} & 0.7525 & \underline{0.7814} & 0.8028 & \underline{0.7826} & 0.7717 & \underline{0.7032} & \underline{0.8182} & \underline{0.7732} \\
CHORD w/o Intra & \underline{0.7450} & 0.7863 & 0.7596 & 0.7693 & 0.7891 & 0.6732 & 0.8045 & 0.7610 & \underline{0.7833} & 0.7705 & 0.7913 & 0.7784 & 0.7949 & 0.6963 & 0.7728 & 0.7696 \\
\rowcolor{gray!18} \multicolumn{17}{l}{\textbf{Proposed method}} \\
\rowcolor{SkyBlueRow} \textbf{CHORD} & \textbf{0.7720} & \textbf{0.8237} & \textbf{0.8143} & \textbf{0.8070} & \textbf{0.8387} & \textbf{0.7248} & \textbf{0.8722} & \textbf{0.8075} & \textbf{0.8069} & \textbf{0.8104} & \textbf{0.8333} & \textbf{0.8027} & \textbf{0.8406} & \textbf{0.7207} & \textbf{0.8472} & \textbf{0.8088} \\
\midrule[1.5pt]
\multicolumn{17}{c}{} \\[-2mm]
& \multicolumn{8}{c}{Ministral3-8B} & \multicolumn{8}{c}{Ministral3-14B} \\
\cmidrule(lr){2-9}\cmidrule(lr){10-17}
\midrule
\rowcolor{gray!18} \multicolumn{17}{l}{\textbf{Training-free methods}} \\
Perplexity & 0.6443 & 0.6328 & 0.6576 & 0.6397 & 0.6528 & 0.6308 & 0.6785 & 0.6481 & 0.6335 & 0.6510 & 0.6471 & 0.6357 & 0.6526 & 0.6029 & 0.6635 & 0.6409 \\
Semantic Entropy & 0.6828 & 0.6800 & 0.7148 & 0.6867 & 0.7045 & 0.6424 & 0.7338 & 0.6921 & 0.6882 & 0.6925 & 0.7006 & 0.6930 & 0.7104 & 0.6534 & 0.7283 & 0.6952 \\
EigenScore & 0.6856 & 0.6729 & 0.6918 & 0.6788 & 0.6836 & 0.6519 & 0.7076 & 0.6817 & 0.6668 & 0.7015 & 0.6846 & 0.6746 & 0.6926 & 0.6509 & 0.7090 & 0.6829 \\
Lexical Similarity & 0.6530 & 0.6474 & 0.7075 & 0.6906 & 0.6479 & 0.6265 & 0.6698 & 0.6632 & 0.6632 & 0.6595 & 0.6945 & 0.6788 & 0.6610 & 0.6255 & 0.6682 & 0.6644 \\
Verbalize & 0.5019 & 0.5116 & 0.5440 & 0.5172 & 0.5230 & 0.5112 & 0.5528 & 0.5231 & 0.5143 & 0.5340 & 0.5158 & 0.5075 & 0.5231 & 0.5048 & 0.5506 & 0.5214 \\
InterrogateLLM & 0.6963 & 0.7002 & 0.7030 & 0.6969 & 0.7254 & 0.6961 & 0.7446 & 0.7089 & 0.6989 & 0.7295 & 0.7097 & 0.6862 & 0.7163 & 0.6957 & 0.7432 & 0.7114 \\
\rowcolor{gray!18} \multicolumn{17}{l}{\textbf{Training-based methods}} \\
MM & 0.4807 & 0.4840 & 0.5294 & 0.5123 & 0.5298 & 0.4853 & 0.5285 & 0.5071 & 0.4866 & 0.5095 & 0.5123 & 0.4854 & 0.5284 & 0.4924 & 0.5232 & 0.5054 \\
SEP & 0.5232 & 0.5188 & 0.5162 & 0.4990 & 0.5033 & 0.5030 & 0.5288 & 0.5132 & 0.4860 & 0.5076 & 0.5123 & 0.4999 & 0.5186 & 0.4895 & 0.5391 & 0.5076 \\
SAPLMA & 0.4816 & 0.5275 & 0.5396 & 0.5073 & 0.5335 & 0.4944 & 0.5392 & 0.5176 & 0.4865 & 0.5232 & 0.4952 & 0.5091 & 0.5266 & 0.4839 & 0.5386 & 0.5090 \\
LapEigvals & 0.4708 & 0.4737 & 0.5019 & 0.5031 & 0.4983 & 0.4733 & 0.5295 & 0.4929 & 0.4847 & 0.4766 & 0.5179 & 0.4800 & 0.5058 & 0.4702 & 0.5161 & 0.4930 \\
\rowcolor{gray!18} \multicolumn{17}{l}{\textbf{Edge ablation variants}} \\
CHORD w/o Token & 0.7183 & 0.7466 & \underline{0.7543} & 0.7247 & \underline{0.7622} & 0.6957 & 0.7697 & 0.7388 & 0.7232 & 0.7528 & \underline{0.7739} & 0.7326 & 0.7769 & 0.7104 & 0.7762 & 0.7494 \\
CHORD w/o Layer & 0.7137 & \underline{0.7813} & 0.7526 & \underline{0.7628} & 0.7548 & 0.7097 & \underline{0.8014} & \underline{0.7538} & 0.7194 & \underline{0.7704} & 0.7708 & \underline{0.7638} & 0.7457 & 0.7008 & \underline{0.8143} & 0.7550 \\
CHORD w/o Intra & \underline{0.7340} & 0.7570 & 0.7408 & 0.7396 & 0.7538 & \underline{0.7121} & 0.7800 & 0.7453 & \underline{0.7400} & 0.7630 & 0.7511 & 0.7478 & \underline{0.7881} & \underline{0.7159} & 0.7924 & \underline{0.7569} \\
\rowcolor{gray!18} \multicolumn{17}{l}{\textbf{Proposed method}} \\
\rowcolor{SkyBlueRow} \textbf{CHORD} & \textbf{0.7663} & \textbf{0.8005} & \textbf{0.7914} & \textbf{0.7836} & \textbf{0.8021} & \textbf{0.7263} & \textbf{0.8474} & \textbf{0.7882} & \textbf{0.7760} & \textbf{0.7988} & \textbf{0.8046} & \textbf{0.7933} & \textbf{0.8193} & \textbf{0.7237} & \textbf{0.8557} & \textbf{0.7959} \\
\bottomrule
\end{tabular}%
}
\vspace{-2mm}
\end{table*}

Table~\ref{tab:edge_ablation_full} reports the full per-dataset and per-model edge-ablation results. 
The table also includes training-free and training-based baselines, allowing the ablated variants to be compared both against the complete method and against the strongest non-\method{} alternatives.

The first observation is that removing any single edge channel consistently weakens \method{}. 
Full \method{} achieves the best performance across the table, while each edge-ablated variant loses performance on average. 
This supports the view that the three relation channels are complementary: none of them alone explains the entire gain, but each contributes useful information to the final detector. 
At the same time, the single-edge ablations remain much stronger than ordinary training-based baselines. 
This is expected because each ablated variant still retains most of \method{}, including the other relation channels and the meta-learning objective. 
Therefore, the edge ablation should be interpreted as measuring marginal channel contribution, not as a complete removal of relational modeling.

The second observation is that different edge channels matter for different benchmark types. 
Removing token-level edges tends to hurt settings where answer trajectory and token-level consistency are important, especially xLAM and dialogue-style or answer-development benchmarks such as CoQA and HaluEval. 
This pattern is natural because token-level relations capture how hidden states vary across generated answer positions. 
When this channel is removed, the detector has less access to the sequential structure of the response, which is particularly relevant for agent-oriented hallucinations and multi-step answer generation.

Removing layer-level edges has a different profile. 
Its effect is more visible on reasoning and semantic-consistency benchmarks such as GSM8K and TruthfulQA. 
Layer-level relations encode how representations evolve across Transformer depth, and this evolution can carry information about whether an answer is internally supported or whether the model is drifting toward an unsupported conclusion. 
The stronger degradation on reasoning-oriented settings suggests that layerwise representation dynamics provide useful signals beyond token-level variation alone.

Removing intra-block feature edges affects yet another part of the evaluation. 
The drop is more pronounced on MBPP and also appears in some agent and knowledge-intensive settings. 
Intra-block edges capture fine-grained relations among feature blocks inside the unified hidden representation. 
Such structure can be useful when hallucination signals are not expressed primarily as token-position changes or layerwise shifts, but instead appear in local activation patterns. 
This helps explain why the intra-block channel is especially relevant for code and other settings where correctness may depend on subtle internal distinctions.

The three ablations therefore do not form a single universal ordering. 
Instead, their relative importance changes with the benchmark and target model. 
This is important because it would be less informative if one edge type dominated all settings or if all ablations produced indistinguishable drops. 
The observed pattern suggests that token, layer, and intra-block edges capture different failure modes of generated answers. 
\method{} benefits from combining them because hallucinations are heterogeneous: some errors are reflected in token-level answer development, some in layerwise semantic evolution, and others in fine-grained feature-block structure.

Finally, the edge ablation results complement the larger module-level ablations. 
Removing one edge channel causes a smaller drop than removing relation-aware attention entirely, which is expected because the remaining channels can still carry relational information. 
However, the consistent degradation relative to full \method{} shows that these channels are not redundant. 
Together, Table~\ref{tab:edge_ablation_full} indicates that \method{}'s advantage comes from aggregating multiple complementary relation channels rather than relying on a single handcrafted relational view.

\subsection{Cross-Benchmark Transfer}
\label{app:exp_cross_dataset_transfer}

\begin{table*}[t]
\caption{Cross-dataset hallucination detection performance under leave-one-benchmark-out evaluation. Training-based methods are trained on six source benchmarks and evaluated on the held-out target benchmark for each target model, while training-free methods are directly applied to the same targets. The table is split into four blocks only due to space constraints. We report AUROC on each held-out benchmark and the average across benchmarks for each target model. Best results are \textbf{bold}, second-best are \underline{underlined}.}
\vspace{1mm}
\centering
\label{tab:cross_dataset_full}
\resizebox{\linewidth}{!}{%
\begin{tabular}{lcccccccccccccccc}
\toprule
\multirow{4}{*}{\textbf{Method}} & \multicolumn{16}{c}{\textbf{Target Model}} \\
& \multicolumn{8}{c}{Llama3.2-3B} & \multicolumn{8}{c}{Llama3.1-8B} \\
\cmidrule(lr){2-9}\cmidrule(lr){10-17}
& \multicolumn{8}{c}{Mean AUROC$\uparrow$} & \multicolumn{8}{c}{Mean AUROC$\uparrow$} \\
& TruthfulQA & HaluEval & TriviaQA & CoQA & GSM8K & MBPP & xLAM & \textbf{Average} & TruthfulQA & HaluEval & TriviaQA & CoQA & GSM8K & MBPP & xLAM & \textbf{Average} \\
\midrule
\rowcolor{gray!18} \multicolumn{17}{l}{\textbf{Training-free methods}} \\
Perplexity & 0.6329 & 0.6464 & 0.6448 & 0.6296 & 0.6536 & 0.6188 & 0.6621 & 0.6412 & 0.6546 & 0.6452 & 0.6545 & 0.6395 & 0.6523 & 0.6024 & 0.6857 & 0.6477 \\
Semantic Entropy & \underline{0.6862} & 0.7095 & 0.7006 & \underline{0.6934} & 0.7112 & 0.6571 & 0.7223 & 0.6972 & 0.6991 & 0.7113 & 0.7065 & \underline{0.6919} & 0.7258 & 0.6636 & 0.7438 & 0.7060 \\
EigenScore & 0.6636 & 0.6872 & 0.6767 & 0.6719 & 0.6772 & 0.6508 & 0.7045 & 0.6760 & 0.6985 & 0.6963 & 0.6803 & 0.6747 & 0.7068 & 0.6614 & 0.6957 & 0.6877 \\
Lexical Similarity & 0.6542 & 0.6747 & 0.6878 & 0.6786 & 0.6454 & 0.6225 & 0.6603 & 0.6605 & 0.6760 & 0.6681 & 0.6928 & 0.6628 & 0.6495 & 0.6236 & 0.6871 & 0.6657 \\
Verbalize & 0.5199 & 0.5306 & 0.5220 & 0.5290 & 0.5207 & 0.5069 & 0.5470 & 0.5252 & 0.5238 & 0.5398 & 0.5110 & 0.5050 & 0.5474 & 0.4884 & 0.5595 & 0.5250 \\
InterrogateLLM & 0.6845 & \underline{0.7353} & \underline{0.7012} & 0.6915 & \underline{0.7177} & \underline{0.6915} & \underline{0.7253} & \underline{0.7067} & \underline{0.7185} & \underline{0.7211} & \underline{0.7201} & 0.6881 & \underline{0.7293} & \underline{0.6865} & \underline{0.7469} & \underline{0.7158} \\
\rowcolor{gray!18} \multicolumn{17}{l}{\textbf{Training-based methods}} \\
MM & 0.6128 & 0.6659 & 0.6527 & 0.6291 & 0.6503 & 0.6089 & 0.6529 & 0.6389 & 0.6636 & 0.6423 & 0.6537 & 0.6280 & 0.6339 & 0.6024 & 0.6656 & 0.6414 \\
SEP & 0.6338 & 0.6718 & 0.6496 & 0.6345 & 0.6556 & 0.6181 & 0.6709 & 0.6478 & 0.6705 & 0.6723 & 0.6904 & 0.6495 & 0.6704 & 0.6289 & 0.6936 & 0.6679 \\
SAPLMA & 0.6542 & 0.6944 & 0.6592 & 0.6812 & 0.6720 & 0.6325 & 0.6793 & 0.6675 & 0.6822 & 0.6838 & 0.6929 & 0.6746 & 0.6998 & 0.6480 & 0.7057 & 0.6839 \\
LapEigvals & 0.5986 & 0.6534 & 0.6150 & 0.6036 & 0.6389 & 0.5993 & 0.6474 & 0.6223 & 0.6646 & 0.6276 & 0.6517 & 0.6292 & 0.6408 & 0.5847 & 0.6562 & 0.6364 \\
\rowcolor{gray!18} \multicolumn{17}{l}{\textbf{Proposed method}} \\
\rowcolor{SkyBlueRow} \textbf{CHORD} & \textbf{0.7586} & \textbf{0.7821} & \textbf{0.7749} & \textbf{0.7723} & \textbf{0.8027} & \textbf{0.7026} & \textbf{0.8060} & \textbf{0.7713} & \textbf{0.7767} & \textbf{0.7991} & \textbf{0.8013} & \textbf{0.7632} & \textbf{0.8072} & \textbf{0.7018} & \textbf{0.8450} & \textbf{0.7849} \\
\midrule[1.5pt]
\multicolumn{17}{c}{} \\[-2mm]
& \multicolumn{8}{c}{Llama3.3-70B} & \multicolumn{8}{c}{Qwen3-8B} \\
\cmidrule(lr){2-9}\cmidrule(lr){10-17}
\midrule
\rowcolor{gray!18} \multicolumn{17}{l}{\textbf{Training-free methods}} \\
Perplexity & 0.6575 & 0.6747 & 0.6717 & 0.6748 & 0.6765 & 0.6353 & 0.6931 & 0.6691 & 0.6345 & 0.6398 & 0.6781 & 0.6542 & 0.6653 & 0.6179 & 0.6726 & 0.6518 \\
Semantic Entropy & 0.7096 & 0.7366 & 0.7231 & \underline{0.7272} & 0.7410 & 0.6697 & 0.7406 & 0.7211 & 0.6840 & 0.6800 & \underline{0.7127} & 0.6994 & \underline{0.7143} & 0.6502 & 0.7228 & 0.6948 \\
EigenScore & 0.7076 & 0.7087 & 0.6961 & 0.7095 & 0.7094 & 0.6703 & 0.7093 & 0.7016 & 0.6785 & 0.6817 & 0.6869 & 0.6873 & 0.6813 & 0.6683 & 0.7085 & 0.6846 \\
Lexical Similarity & 0.6844 & 0.6797 & 0.7040 & 0.7088 & 0.6854 & 0.6432 & 0.6858 & 0.6845 & 0.6538 & 0.6538 & 0.7089 & 0.6966 & 0.6441 & 0.6401 & 0.6688 & 0.6666 \\
Verbalize & 0.5472 & 0.5531 & 0.5472 & 0.5385 & 0.5592 & 0.5139 & 0.5584 & 0.5454 & 0.5193 & 0.5090 & 0.5200 & 0.5170 & 0.5339 & 0.4969 & 0.5463 & 0.5203 \\
InterrogateLLM & \underline{0.7343} & \underline{0.7454} & \underline{0.7472} & 0.7219 & \underline{0.7473} & \underline{0.7181} & \underline{0.7477} & \underline{0.7374} & \underline{0.6950} & \underline{0.7077} & 0.7044 & \underline{0.7173} & 0.7082 & \underline{0.6834} & \underline{0.7483} & \underline{0.7092} \\
\rowcolor{gray!18} \multicolumn{17}{l}{\textbf{Training-based methods}} \\
MM & 0.6706 & 0.6847 & 0.7105 & 0.6692 & 0.7041 & 0.6350 & 0.6871 & 0.6802 & 0.6192 & 0.6436 & 0.6503 & 0.6425 & 0.6329 & 0.5942 & 0.6756 & 0.6369 \\
SEP & 0.6964 & 0.7065 & 0.7329 & 0.6931 & 0.7026 & 0.6523 & 0.6759 & 0.6942 & 0.6439 & 0.6477 & 0.6685 & 0.6460 & 0.6303 & 0.6077 & 0.6933 & 0.6482 \\
SAPLMA & 0.7005 & 0.7212 & 0.7161 & 0.7093 & 0.7049 & 0.6695 & 0.7169 & 0.7055 & 0.6642 & 0.6827 & 0.6705 & 0.7003 & 0.6639 & 0.6323 & 0.7035 & 0.6739 \\
LapEigvals & 0.6537 & 0.6799 & 0.6808 & 0.6741 & 0.6889 & 0.6470 & 0.6984 & 0.6747 & 0.6017 & 0.6283 & 0.6406 & 0.6409 & 0.6182 & 0.5876 & 0.6514 & 0.6241 \\
\rowcolor{gray!18} \multicolumn{17}{l}{\textbf{Proposed method}} \\
\rowcolor{SkyBlueRow} \textbf{CHORD} & \textbf{0.8044} & \textbf{0.8281} & \textbf{0.8120} & \textbf{0.8124} & \textbf{0.8304} & \textbf{0.7362} & \textbf{0.8594} & \textbf{0.8118} & \textbf{0.7697} & \textbf{0.7607} & \textbf{0.8005} & \textbf{0.7667} & \textbf{0.7990} & \textbf{0.7089} & \textbf{0.8173} & \textbf{0.7747} \\
\midrule[1.5pt]
\multicolumn{17}{c}{} \\[-2mm]
& \multicolumn{8}{c}{Qwen3-14B} & \multicolumn{8}{c}{Qwen3.5-35B} \\
\cmidrule(lr){2-9}\cmidrule(lr){10-17}
\midrule
\rowcolor{gray!18} \multicolumn{17}{l}{\textbf{Training-free methods}} \\
Perplexity & 0.6408 & 0.6650 & 0.6442 & 0.6519 & 0.6363 & 0.6164 & 0.6851 & 0.6485 & 0.6460 & 0.6488 & 0.6751 & 0.6601 & 0.6614 & 0.6361 & 0.6732 & 0.6572 \\
Semantic Entropy & 0.6810 & 0.6968 & 0.6966 & 0.6958 & 0.7193 & 0.6438 & 0.7195 & 0.6933 & \underline{0.7078} & \underline{0.7110} & 0.7211 & 0.7052 & 0.7392 & 0.6738 & 0.7278 & 0.7123 \\
EigenScore & 0.6931 & 0.6891 & 0.6965 & 0.6748 & 0.7013 & 0.6556 & 0.7090 & 0.6885 & 0.7061 & 0.7018 & 0.7072 & 0.6854 & 0.7093 & 0.6648 & 0.7046 & 0.6970 \\
Lexical Similarity & 0.6498 & 0.6560 & 0.6779 & 0.6795 & 0.6529 & 0.6368 & 0.6834 & 0.6623 & 0.6752 & 0.6756 & 0.7050 & 0.7083 & 0.6583 & 0.6377 & 0.6730 & 0.6762 \\
Verbalize & 0.5349 & 0.5262 & 0.5274 & 0.5261 & 0.5329 & 0.5091 & 0.5578 & 0.5306 & 0.5286 & 0.5352 & 0.5511 & 0.5279 & 0.5460 & 0.5150 & 0.5550 & 0.5370 \\
InterrogateLLM & \underline{0.7009} & \underline{0.7127} & \underline{0.7062} & \underline{0.7014} & \underline{0.7496} & \underline{0.6904} & \underline{0.7435} & \underline{0.7150} & 0.6978 & 0.6997 & \underline{0.7248} & \underline{0.7273} & \underline{0.7438} & \underline{0.6902} & \underline{0.7376} & \underline{0.7173} \\
\rowcolor{gray!18} \multicolumn{17}{l}{\textbf{Training-based methods}} \\
MM & 0.6327 & 0.6421 & 0.6423 & 0.6398 & 0.6688 & 0.6105 & 0.6749 & 0.6444 & 0.6213 & 0.6319 & 0.6573 & 0.7084 & 0.6799 & 0.6067 & 0.6972 & 0.6575 \\
SEP & 0.6528 & 0.6684 & 0.6703 & 0.6469 & 0.6739 & 0.6098 & 0.6725 & 0.6564 & 0.6515 & 0.6722 & 0.6779 & 0.6904 & 0.6927 & 0.6212 & 0.6719 & 0.6683 \\
SAPLMA & 0.6703 & 0.6888 & 0.6935 & 0.6585 & 0.7080 & 0.6343 & 0.7157 & 0.6813 & 0.6489 & 0.6593 & 0.7000 & 0.6993 & 0.7055 & 0.6561 & 0.7164 & 0.6836 \\
LapEigvals & 0.6249 & 0.6361 & 0.6366 & 0.6271 & 0.6798 & 0.5883 & 0.6601 & 0.6361 & 0.6297 & 0.6481 & 0.6629 & 0.6433 & 0.6646 & 0.6084 & 0.6712 & 0.6469 \\
\rowcolor{gray!18} \multicolumn{17}{l}{\textbf{Proposed method}} \\
\rowcolor{SkyBlueRow} \textbf{CHORD} & \textbf{0.7601} & \textbf{0.8037} & \textbf{0.7914} & \textbf{0.7923} & \textbf{0.8005} & \textbf{0.7046} & \textbf{0.8337} & \textbf{0.7838} & \textbf{0.7712} & \textbf{0.7842} & \textbf{0.8091} & \textbf{0.7789} & \textbf{0.8186} & \textbf{0.7099} & \textbf{0.8213} & \textbf{0.7847} \\
\midrule[1.5pt]
\multicolumn{17}{c}{} \\[-2mm]
& \multicolumn{8}{c}{Ministral3-8B} & \multicolumn{8}{c}{Ministral3-14B} \\
\cmidrule(lr){2-9}\cmidrule(lr){10-17}
\midrule
\rowcolor{gray!18} \multicolumn{17}{l}{\textbf{Training-free methods}} \\
Perplexity & 0.6443 & 0.6328 & 0.6576 & 0.6397 & 0.6528 & 0.6308 & 0.6785 & 0.6481 & 0.6335 & 0.6510 & 0.6471 & 0.6357 & 0.6526 & 0.6029 & 0.6635 & 0.6409 \\
Semantic Entropy & 0.6828 & 0.6800 & \underline{0.7148} & 0.6867 & 0.7045 & 0.6424 & 0.7338 & 0.6921 & 0.6882 & 0.6925 & 0.7006 & \underline{0.6930} & 0.7104 & 0.6534 & 0.7283 & 0.6952 \\
EigenScore & 0.6856 & 0.6729 & 0.6918 & 0.6788 & 0.6836 & 0.6519 & 0.7076 & 0.6817 & 0.6668 & 0.7015 & 0.6846 & 0.6746 & 0.6926 & 0.6509 & 0.7090 & 0.6829 \\
Lexical Similarity & 0.6530 & 0.6474 & 0.7075 & 0.6906 & 0.6479 & 0.6265 & 0.6698 & 0.6632 & 0.6632 & 0.6595 & 0.6945 & 0.6788 & 0.6610 & 0.6255 & 0.6682 & 0.6644 \\
Verbalize & 0.5019 & 0.5116 & 0.5440 & 0.5172 & 0.5230 & 0.5112 & 0.5528 & 0.5231 & 0.5143 & 0.5340 & 0.5158 & 0.5075 & 0.5231 & 0.5048 & 0.5506 & 0.5214 \\
InterrogateLLM & \underline{0.6963} & \underline{0.7002} & 0.7030 & \underline{0.6969} & \underline{0.7254} & \underline{0.6961} & \underline{0.7446} & \underline{0.7089} & \underline{0.6989} & \underline{0.7295} & \underline{0.7097} & 0.6862 & \underline{0.7163} & \underline{0.6957} & \underline{0.7432} & \underline{0.7114} \\
\rowcolor{gray!18} \multicolumn{17}{l}{\textbf{Training-based methods}} \\
MM & 0.6109 & 0.6494 & 0.6493 & 0.6284 & 0.6447 & 0.6029 & 0.6698 & 0.6365 & 0.6154 & 0.6598 & 0.6545 & 0.6308 & 0.6458 & 0.6033 & 0.6816 & 0.6416 \\
SEP & 0.6152 & 0.6499 & 0.6529 & 0.6253 & 0.6576 & 0.6295 & 0.6643 & 0.6421 & 0.6469 & 0.6750 & 0.6469 & 0.6401 & 0.6535 & 0.6178 & 0.6956 & 0.6537 \\
SAPLMA & 0.6508 & 0.6713 & 0.6737 & 0.6555 & 0.6753 & 0.6477 & 0.6850 & 0.6656 & 0.6551 & 0.7133 & 0.6705 & 0.6676 & 0.6669 & 0.6428 & 0.7154 & 0.6759 \\
LapEigvals & 0.6150 & 0.6231 & 0.6371 & 0.6093 & 0.6298 & 0.6112 & 0.6737 & 0.6285 & 0.5982 & 0.6484 & 0.6386 & 0.6004 & 0.6225 & 0.6083 & 0.6500 & 0.6238 \\
\rowcolor{gray!18} \multicolumn{17}{l}{\textbf{Proposed method}} \\
\rowcolor{SkyBlueRow} \textbf{CHORD} & \textbf{0.7365} & \textbf{0.7712} & \textbf{0.7661} & \textbf{0.7492} & \textbf{0.7624} & \textbf{0.7098} & \textbf{0.8136} & \textbf{0.7584} & \textbf{0.7539} & \textbf{0.7764} & \textbf{0.7821} & \textbf{0.7663} & \textbf{0.7791} & \textbf{0.7107} & \textbf{0.8228} & \textbf{0.7702} \\
\bottomrule
\end{tabular}%
}
\vspace{-2mm}
\end{table*}

The main experiments focus on cross-model generalization under a leave-one-family-out protocol. 
In that setting, a detector is trained on source LLMs and evaluated on unseen target LLMs, with the same benchmark used for both source and target evaluation. 
A natural question is whether the observed gains depend on this same-benchmark training setup. 
To address this, we further evaluate cross-benchmark transfer, where the training and test benchmarks are different while the target LLM is fixed.

\paragraph{Protocol.}
For each LLM $\mathbb{P}_{\boldsymbol{\theta}_i}$ and each held-out benchmark 
$d\in\{\text{TruthfulQA},\text{HaluEval},\text{TriviaQA},\text{CoQA},\text{GSM8K},\text{MBPP},\text{xLAM}\}$, 
we train each training-based detector on the other six benchmarks generated by the same LLM and evaluate it on the held-out benchmark $d$. 
Concretely, if $\mathcal{D}_{i,d}$ denotes the dataset constructed from model $\mathbb{P}_{\boldsymbol{\theta}_i}$ on benchmark $d$, then the cross-benchmark training set for held-out benchmark $d$ is
\[
    \mathcal{D}_{i,\neg d}
    =
    \bigcup_{d'\neq d}\mathcal{D}_{i,d'}.
\]
The detector is trained on $\mathcal{D}_{i,\neg d}$ and evaluated on $\mathcal{D}_{i,d}$. 
Thus, each column in Table~\ref{tab:cross_dataset_full} should be read as a \emph{held-out benchmark}: the corresponding benchmark is used only for evaluation, while the remaining six benchmarks are used for training.

This protocol isolates benchmark shift from model shift. 
It is therefore easier than the leave-one-family-out cross-model setting in the main experiment, because the source and test features come from the same LLM and share the same hidden-state coordinate system. 
At the same time, it is still a meaningful transfer test: the detector must generalize from one set of task distributions to a different held-out benchmark, rather than being trained and tested on the same benchmark.

\paragraph{Treatment of training-free methods.}
Training-free methods do not use the source benchmarks. 
For a fixed model and held-out benchmark, their scores are therefore identical to the corresponding model--benchmark evaluation in the main table. 
We include them in Table~\ref{tab:cross_dataset_full} as source-independent reference baselines. 
Their repeated consistency with the main results is expected and should not be interpreted as an additional training condition. 
This also provides a useful anchor for comparing training-based detectors under benchmark shift.

\paragraph{Results.}
Table~\ref{tab:cross_dataset_full} reports the full cross-benchmark results. 
\method{} achieves the strongest performance across target models and held-out benchmarks, showing that its advantage is not restricted to training and testing on the same benchmark. 
The result is particularly important because cross-benchmark transfer tests a different form of robustness from the main cross-model experiment. 
The main experiment asks whether a detector trained on some LLMs can generalize to unseen LLMs; this experiment asks whether a detector trained on some benchmarks can generalize to unseen benchmarks for the same LLM. 
\method{} remains effective in both settings.

The behavior of ordinary training-based baselines is also informative. 
Unlike in the leave-one-family-out cross-model setting, methods such as MM, SEP, SAPLMA, and LapEigvals do not collapse to chance here. 
Their AUROC is often substantially above $0.5$ and can approach strong training-free baselines in favorable cases. 
This is expected: when training and evaluation are performed on the same LLM, the representation space is shared, so a supervised detector can reuse model-specific coordinates and decision directions across benchmarks. 
The result therefore suggests that these baselines are not intrinsically ineffective or improperly tuned; rather, their weakness in the main experiment is largely caused by cross-model representation mismatch.

Training-free baselines remain competitive, especially Semantic Entropy and InterrogateLLM. 
Their strength is natural because they do not rely on a learned source-benchmark representation and therefore avoid overfitting to a particular training benchmark. 
However, they do not exploit cross-benchmark supervision, and \method{} consistently improves over them. 
This indicates that useful hallucination signals can be learned across benchmarks when the detector is built on transferable relational internal features.

\paragraph{Benchmark-level trends.}
The cross-benchmark results follow several task-level patterns. 
QA-style benchmarks such as TruthfulQA, HaluEval, TriviaQA, and CoQA transfer relatively smoothly because they share similar input-output formats and factuality-oriented labels. 
GSM8K also shows strong performance, suggesting that reasoning errors induce internal relational patterns that can be learned from other benchmarks and reused on mathematical reasoning. 
MBPP remains the most fragile benchmark: the gap between \method{} and the strongest baselines is smaller, reflecting the fact that code correctness is more execution-dependent and differs more sharply from natural-language QA. 
xLAM remains comparatively favorable for \method{}, consistent with the main results that agent-oriented hallucinations benefit from relational internal signals.

\paragraph{Implication.}
This experiment clarifies the relationship between benchmark shift and model shift. 
Cross-benchmark transfer within a fixed LLM is less severe than cross-model transfer, and ordinary training-based detectors can retain meaningful performance in this setting. 
However, \method{} remains consistently stronger, indicating that its relational formulation is useful beyond the specific benchmark used for training. 
Together with the leave-one-family-out results, Table~\ref{tab:cross_dataset_full} shows that \method{} addresses both sources of generalization difficulty: it transfers across benchmarks when the LLM is fixed, and it transfers across LLMs when the benchmark is fixed.

\newpage
\input{checklist.tex}

\end{document}

%% file: checklist.tex
\section*{NeurIPS Paper Checklist}

\begin{enumerate}

\item {\bf Claims}
    \item[] Question: Do the main claims made in the abstract and introduction accurately reflect the paper's contributions and scope?
    \item[] Answer: \answerYes{}
    \item[] Justification: The claims made in the abstract and Section 1 are explicitly supported by our methodology in Section \ref{sec:method} and experimental results in Section \ref{sec:exp}.

\item {\bf Limitations}
    \item[] Question: Does the paper discuss the limitations of the work performed by the authors?
    \item[] Answer: \answerYes{}
    \item[] Justification: Please see Appendix~\ref{app:limitations}.

\item {\bf Theory assumptions and proofs}
    \item[] Question: For each theoretical result, does the paper provide the full set of assumptions and a complete (and correct) proof?
    \item[] Answer: \answerNA{}
    \item[] Justification: This paper focuses exclusively on empirical evaluations and architectural design, and thus does not include theoretical results or mathematical proofs.

    \item {\bf Experimental result reproducibility}
    \item[] Question: Does the paper fully disclose all the information needed to reproduce the main experimental results of the paper to the extent that it affects the main claims and/or conclusions of the paper (regardless of whether the code and data are provided or not)?
    \item[] Answer: \answerYes{}
    \item[] Justification: All necessary code, datasets, and detailed instructions required to reproduce the main experimental results are publicly available in our repository: \url{https://github.com/MiracleworkerD/CHORD}.

\item {\bf Open access to data and code}
    \item[] Question: Does the paper provide open access to the data and code, with sufficient instructions to faithfully reproduce the main experimental results, as described in supplemental material?
    \item[] Answer: \answerYes{}
    \item[] Justification: All necessary code, datasets, and detailed instructions required to reproduce the main experimental results are publicly available in our repository: \url{https://github.com/MiracleworkerD/CHORD}.

\item {\bf Experimental setting/details}
    \item[] Question: Does the paper specify all the training and test details (e.g., data splits, hyperparameters, how they were chosen, type of optimizer) necessary to understand the results?
    \item[] Answer: \answerYes{}
    \item[] Justification: All experimental settings, including data splits, hyperparameters, and optimization details, are extensively detailed in Appendix~\ref{app:exp_details}.

\item {\bf Experiment statistical significance}
    \item[] Question: Does the paper report error bars suitably and correctly defined or other appropriate information about the statistical significance of the experiments?
    \item[] Answer: \answerYes{}
    \item[] Justification: We report suitable error bars and provide appropriate information regarding statistical significance in Figure~\ref{fig:obs_alignment_detection}, with comprehensive details discussed in Appendix~\ref{app:obs_alignment_transfer_details}.

\item {\bf Experiments compute resources}
    \item[] Question: For each experiment, does the paper provide sufficient information on the computer resources (type of compute workers, memory, time of execution) needed to reproduce the experiments?
    \item[] Answer: \answerYes{}
    \item[] Justification: Detailed information regarding the required compute resources, including hardware specifications, is provided in the instructions of our code repository.
    
\item {\bf Code of ethics}
    \item[] Question: Does the research conducted in the paper conform, in every respect, with the NeurIPS Code of Ethics \url{https://neurips.cc/public/EthicsGuidelines}?
    \item[] Answer: \answerYes{}
    \item[] Justification: We have reviewed the NeurIPS Code of Ethics and confirm that the research conducted in this paper conforms to it in every respect.

\item {\bf Broader impacts}
    \item[] Question: Does the paper discuss both potential positive societal impacts and negative societal impacts of the work performed?
    \item[] Answer: \answerNA{}
    \item[] Justification: This paper focuses on foundational machine learning methodology and does not present a direct path to specific applications or negative societal harms.
    
\item {\bf Safeguards}
    \item[] Question: Does the paper describe safeguards that have been put in place for responsible release of data or models that have a high risk for misuse (e.g., pre-trained language models, image generators, or scraped datasets)?
    \item[] Answer: \answerNA{}
    \item[] Justification: The methods and data presented in this paper do not pose a high risk for malicious misuse, thus no specific safeguards for release are required.

\item {\bf Licenses for existing assets}
    \item[] Question: Are the creators or original owners of assets (e.g., code, data, models), used in the paper, properly credited and are the license and terms of use explicitly mentioned and properly respected?
    \item[] Answer: \answerYes{}
    \item[] Justification: All creators of the existing datasets and codebases used in this work are properly cited, and their respective licenses are detailed in Appendix~\ref{app:licenses}.

\item {\bf New assets}
    \item[] Question: Are new assets introduced in the paper well documented and is the documentation provided alongside the assets?
    \item[] Answer: \answerNA{}
    \item[] Justification: This paper focuses on a new algorithmic method and evaluating existing models/datasets; it does not release new foundational models or large-scale datasets.

\item {\bf Crowdsourcing and research with human subjects}
    \item[] Question: For crowdsourcing experiments and research with human subjects, does the paper include the full text of instructions given to participants and screenshots, if applicable, as well as details about compensation (if any)? 
    \item[] Answer: \answerNA{}
    \item[] Justification: This paper is strictly computational and does not involve any crowdsourcing or research with human subjects.

\item {\bf Institutional review board (IRB) approvals or equivalent for research with human subjects}
    \item[] Question: Does the paper describe potential risks incurred by study participants, whether such risks were disclosed to the subjects, and whether Institutional Review Board (IRB) approvals (or an equivalent approval/review based on the requirements of your country or institution) were obtained?
    \item[] Answer: \answerNA{}
    \item[] Justification: Not applicable, as this research does not involve human subjects.

\item {\bf Declaration of LLM usage}
    \item[] Question: Does the paper describe the usage of LLMs if it is an important, original, or non-standard component of the core methods in this research? Note that if the LLM is used only for writing, editing, or formatting purposes and does \emph{not} impact the core methodology, scientific rigor, or originality of the research, declaration is not required.
    \item[] Answer: \answerYes{}
    \item[] Justification: We describe the usage of LLMs explicitly in Section~\ref{sec:method}, as utilizing them for generating responses and evaluation is central to our cross-model transfer methodology.

\end{enumerate}